\pdfoutput=1

\def\MODE{1} 
\if\MODE1

\else

\fi

\documentclass[10pt]{article}

\usepackage{palatcm}
\usepackage{graphicx}
\usepackage{subfigure} 
\usepackage{soul}
\usepackage{color, xcolor}
\usepackage[thinlines]{easytable}
\usepackage{relsize}
\usepackage{xfrac}
\usepackage{verbatim}
\usepackage{algorithm}
\usepackage[noend]{algorithmic}
\usepackage{amsmath}
\usepackage{amssymb}
\usepackage{amsthm}
\usepackage{epsfig}
\usepackage{wrapfig}
\usepackage{url}
\usepackage[colorlinks=true,citecolor=blue,linkcolor=blue]{hyperref}
\usepackage{multirow}
\usepackage{fullpage}

\definecolor{ed}{RGB}{225,0,0}

\newcommand{\bA}{\mathbf{A}}

\usepackage{wrapfig}
\usepackage{bbm}
\usepackage{hyperref}       
\usepackage{url}            
\usepackage{booktabs}       
\usepackage{amsfonts}       
\usepackage{microtype}      
\usepackage{microtype}
\usepackage{graphicx}
\usepackage{booktabs} 
\usepackage{amsthm,amsmath,amssymb}
\usepackage{float,url,amsfonts,alltt}
\usepackage{mathtools,rotating}
\usepackage{ifpdf,fancyvrb}
\usepackage{enumitem}
\usepackage{subfigure}
\usepackage{microtype}
\usepackage{graphicx}
\usepackage{booktabs} 
\usepackage{hyperref}

\usepackage{graphicx,xspace,verbatim,comment}
\usepackage{hyperref,array,color,balance,multirow}
\usepackage{balance,float,url,amsfonts,alltt}
\usepackage{mathtools,rotating,amsmath,amssymb}
\usepackage{color,ifpdf,fancyvrb,array}
\usepackage{etoolbox,listings}
\usepackage{bigstrut,morefloats}
\usepackage{makecell}
\usepackage{pbox}
\usepackage[numbers]{natbib}
\usepackage{algorithm, algorithmic}
\usepackage[boxruled,algo2e,linesnumbered]{algorithm2e}
\DeclarePairedDelimiterX{\inp}[2]{\langle}{\rangle}{#1, #2}

\newtheorem{theorem}{Theorem}

\newtheorem{corollary}[theorem]{Corollary}
\newtheorem{lemma}[theorem]{Lemma}
\newtheorem{definition}{Definition}
\newcommand{\eat}[1]{}

\newcommand{\systemnameSolon}{\textsc{Solon}}

\newcommand{\R}{\mathbb{R}}

\newcommand{\draco}{\textsc{Draco}}
\newcommand{\solon}{\textsc{Solon}}
\newcommand{\bulyan}{\textsc{Bulyan}}

\newcommand{\signum}{\textsc{Signum}}

\newcommand{\ie}{{\emph i.e.,} }
\newcommand{\eg}{{\it e.g.}, }

\DeclarePairedDelimiter{\ceil}{\lceil}{\rceil}

\newcommand{\floor}[1]{\lfloor#1\rfloor}

\numberwithin{equation}{section}

\usepackage{amsmath}
\usepackage{accents}
\newlength{\dhatheight}

\newcommand{\supp}{\text{supp}}

\usepackage[T1]{fontenc}
\usepackage{upgreek}

\usepackage[greek,english]{babel}
\usepackage{pifont}
\usepackage{float}
\usepackage{autobreak}

\newcommand{\WH}[1]{}
\newcommand{\ST}[1]{}
\newcommand{\XP}[1]{}

\newtoggle{tr}
\toggletrue{tr}
\iftoggle{tr}{
	\makeatother
}{}

\usepackage{subfigure}

\title{\solon{}: Communication-efficient Byzantine-resilient Distributed Training via Redundant Gradients}

\author{
Lingjiao Chen$^1$, Leshang Chen$^2$, Hongyi Wang$^3$\footnote{The first three authors contributed equally to this paper.}, Susan Davidson$^2$, Edgar Dobriban$^2$\\\\
Stanford University$^1$, University of Pennsylvania$^2$, University of Wisconsin$^3$
} 

\date{}

\begin{document}

\maketitle

\begin{abstract}
There has been a growing need to provide Byzantine-resilience in distributed  model training.
Existing robust distributed learning algorithms focus on developing sophisticated robust aggregators at the parameter servers, but pay less attention to balancing the communication cost and robustness. 
In this paper, we propose \systemnameSolon{}, an algorithmic framework that exploits gradient redundancy to provide communication efficiency and Byzantine robustness simultaneously.
Our theoretical analysis shows a fundamental trade-off among 
computational load, communication cost, and Byzantine robustness. 
We also develop a concrete algorithm to achieve the optimal trade-off, borrowing ideas from coding theory and sparse recovery.
Empirical experiments on various datasets demonstrate that \solon{} provides significant speedups over existing methods to achieve the same accuracy, \eg over  10$\times$ faster than \bulyan{} and 80\% faster than \draco{}.
We also show that carefully designed Byzantine attacks break \signum{} and \bulyan{}, but do not affect the successful convergence of \solon{}.
\end{abstract}

\section{Introduction}\label{Sec:SOLON:Intro}
The growing size of datasets and machine learning models has led to many developments in distributed training using stochastic optimization~\cite{devlin2018bert,deng2009imagenet,sergeev2018horovod,dean2012large,dobriban2020wonder}.
One of the most widely used 
settings is the {\em parameter server (PS) model}~\cite{li2013parameter,li2014scaling,jiang2020unified}, 
where the gradient computation is partitioned among all compute nodes, typically using stochastic gradient descent (SGD) or its variants. A central parameter server then aggregates the calculated gradients from all compute nodes to update the global model.

However, scaling PS models to large clusters introduces two challenges: guarding against \textit{Byzantine attacks} and managing  \textit{communication overhead}.
Byzantine attacks include erroneous gradients sent from unreliable compute nodes due to power outages, hardware or software errors, as well as malicious attacks. 
The communication overhead of sending gradients to the PS in large clusters can also be extremely high, potentially dominating the training time, ~\cite{dean2012large,alistarh2017qsgd,wen2017terngrad,Wang18Atomo, DRACO}, since the number of gradients sent is linear in the number of compute nodes. 

Although recent work has studied the problem of
Byzantine attacks under the PS model~\cite{ALIE2019,guerraoui2018hidden,xie2020fall}, the communication overhead remains prohibitive. For example, ~\cite{blanchard2017machine,chen2017distributed,yin2018byzantine,AggregathorDamaskinosEGGR19} use {\em robust aggregators} at the PS to mitigate  unreliable gradients, 
while ~\cite{DRACO} introduces {\em algorithmic redundancy} to detect and remove Byzantine nodes. 
Robust aggregators are computationally
expensive due to their super-linear  (often quadratic) dependence on the number
of nodes. They also often have limited convergence guarantees under
Byzantine attacks, e.g., only establishing convergence in the limit, or only guaranteeing that the output of the aggregator has a positive inner product with the true gradient. 
They often require strong bounds on the dimension of the model.
Although algorithmic redundancy or coding-theoretic~\cite{Gradient_Coding, DRACO}
approaches offer strong convergence guarantees, these approaches have high communication overhead.

Thus, it remains an open question to simultaneously provide \textbf{\emph{Byzantine-resilience with strong guarantees and communication efficiency}} for distributed learning.

\begin{figure*}[t] 
\centering
\subfigure[Example of a \solon{} instance]{\includegraphics[width=0.6\textwidth]{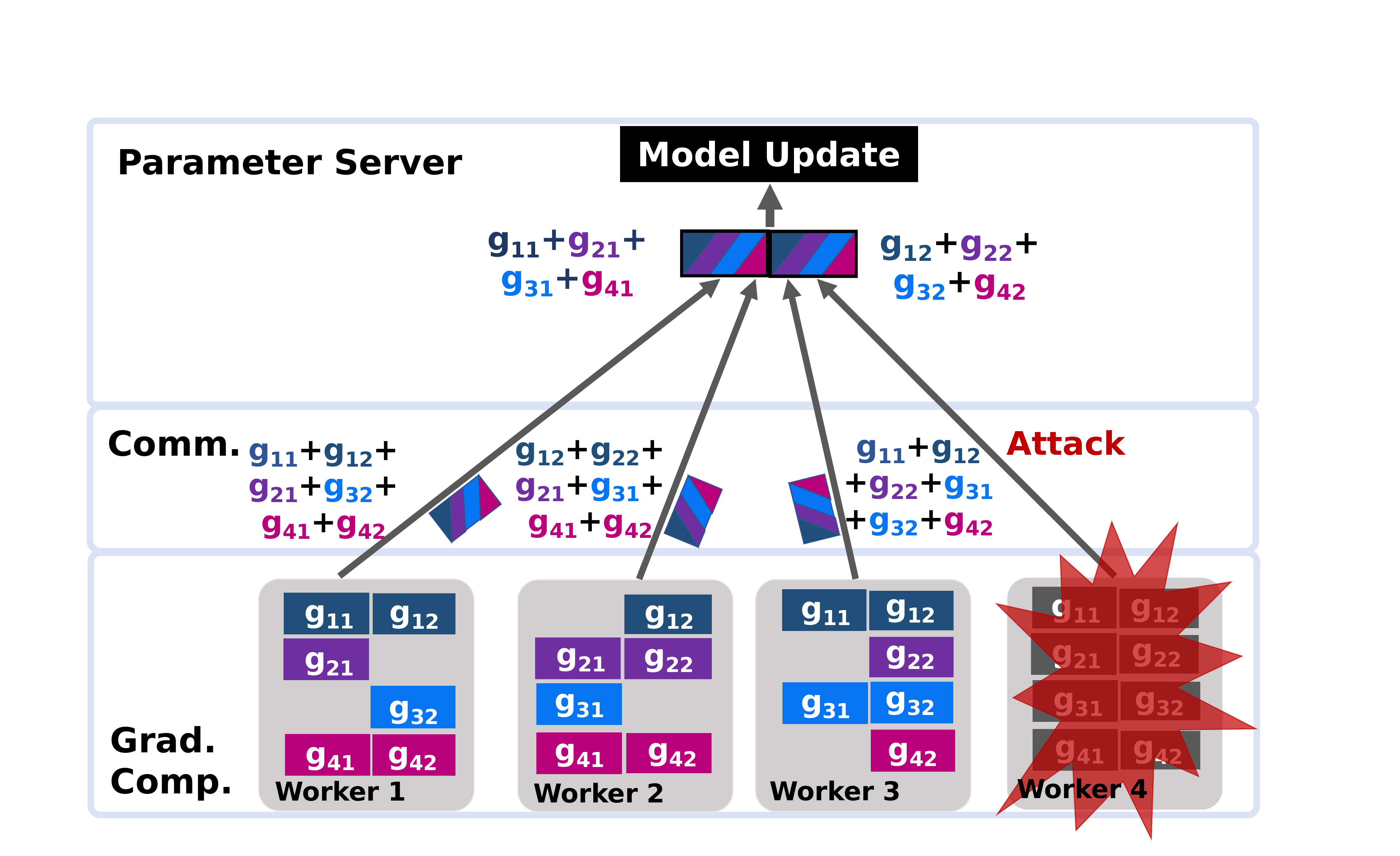}}
\hskip 5pt
\subfigure[Performance of \solon{}]{\includegraphics[width=0.37\textwidth]{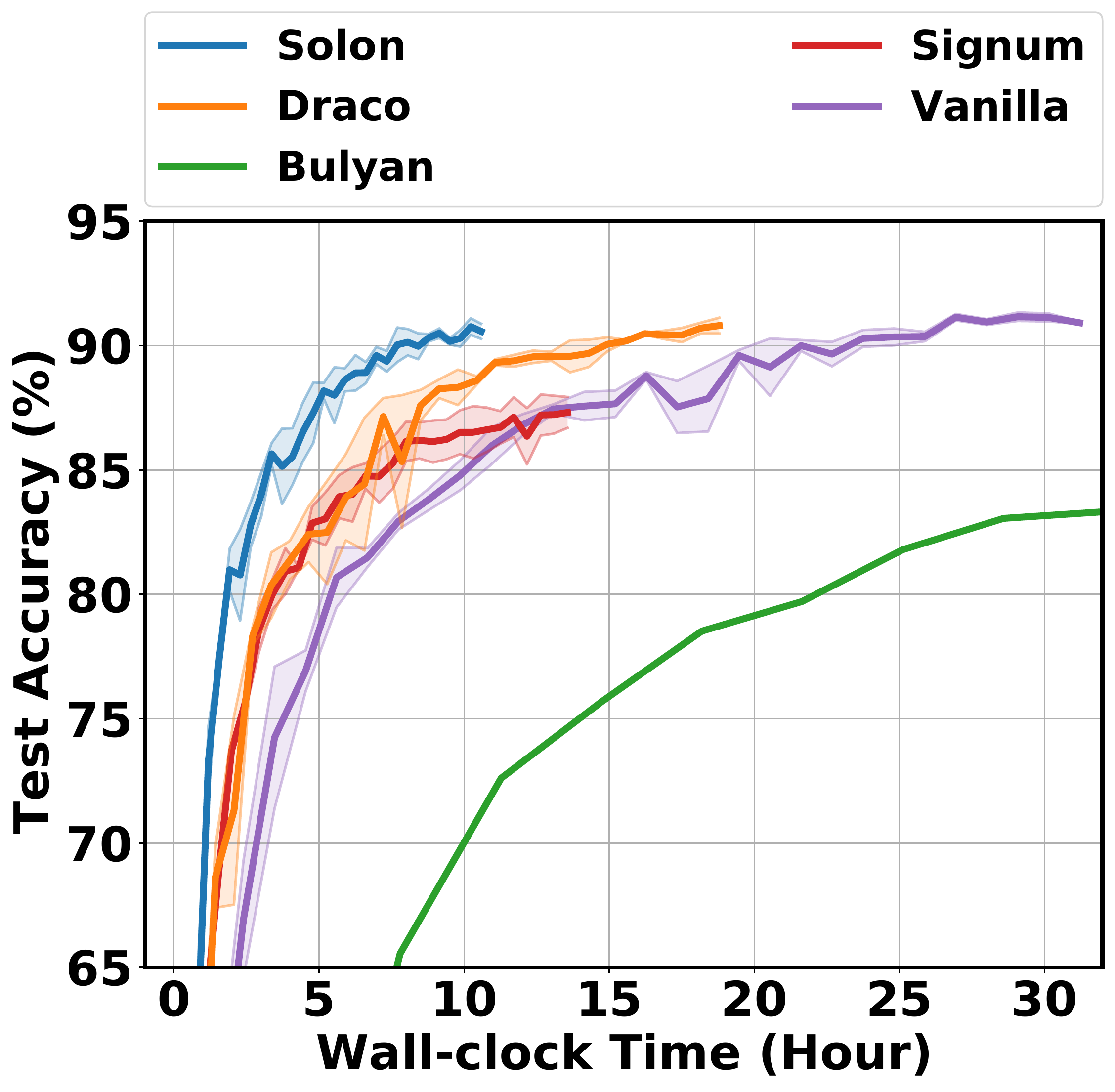}}
\vspace{-2mm}
\caption{\footnotesize{Demonstration of \solon{}. (a) gives an example of \solon{}, where the cluster consists of  four compute nodes.  (b) shows the runtime performance of \solon{} $r_c=10$ when training ResNet-18 on CIFAR-10 under rev-grad attack. Compared to various baselines, \solon{} converges faster and provides higher final accuracy.}}
\label{fig:RunningExample}
\end{figure*}

To address this problem, we propose \solon{}, a distributed training framework leveraging algorithmic redundancy to protect against Byzantine attacks, and ideas from sparse recovery~\cite{Prony} and coding theory~\cite{Gradient_Coding} to reduce communication overhead. 
The approach is as follows: 
In a Byzantine-free PS model with $P$ compute nodes and $B$ gradients to be computed, at each iteration
each of the $P$ nodes computes $B/P$ gradients, and sends them to the PS.  In \solon{}, gradients are computed redundantly to tolerate Byzantine failures; each node computes $rB/P$ gradients, incurring a {\em computational redundancy ratio} of $r$.  To reduce communication overhead, compressed gradients are sent to the PS.  For $d$-dimensional gradients, each node only sends a $d/r_c$-dimensional vector to the PS, where  $r_c$ is the {\em compression ratio}. 

We show that under worst-case adversarial conditions where the adversarial nodes have access to the complete data and gradients, and can send arbitrary results to the PS, there is a fundamental trade-off among Byzantine-resilience, communication overhead and computation cost. To tolerate $s$ Byzantine nodes, the redundancy ratio must satisfy $r\geq 2s+r_c$.
We provide a concrete encoding and decoding technique for \solon{} based on Vandermonde matrices and building on Prony's method from signal processing ~\cite{Prony,communciation_gradcoding_min_ye} that matches the optimal condition  $r= 2s+r_c$.

\paragraph{Example.}
Figure \ref{fig:RunningExample}(a) shows
four two-dimensional gradient vectors, $g_1, g_2,g_3, g_4$, where gradients $g_i=(g_{i1},g_{i2})$ are in different colors.
To tolerate Byzantine workers, 
each node computes four gradients. 
However, instead of sending a two-dimensional vector to the PS, each node only sends a one-dimensional linear combination of the elements in the local gradients. For instance, worker 1 computes and sends the scalar $g_{11}+g_{12}+g_{21}+g_{32}+g_{41}+g_{42}$. Here, the redundancy ratio $r=4$ and the compression ratio $r_c=2$. The PS uses a decoding scheme presented in Section \ref{sec:solon}. Therefore \solon{} can recover the correct gradient with one Byzantine worker ($4 \geq 2\times 1 + 2$). 
\qed

We implemented \solon{} in Pytorch and conducted extensive experiments on a large cluster.  Our results show that
\solon{} can provide significant speedups across various ML models and datasets over existing methods, such as \bulyan{} \cite{guerraoui2018hidden} and \draco{}~\cite{DRACO}, shown in Figure \ref{fig:RunningExample}(b) using ResNet-18~\cite{ResNet} on CIFAR-10~\cite{Cifar10} under a reverse gradient attack. 
In addition, 
\solon{} successfully defends against strong Byzantine attacks such as ``A little is enough'' (ALIE)~\cite{ALIE2019}, on which some methods such as \bulyan{} and \signum{} \cite{bernstein2018signsgd} fail to converge or result in significant accuracy loss, see Section \ref{Sec:SOLON:Experiment}. 

\vspace{-0.1in}
\paragraph{Contributions.}   Our contributions include:
\vspace{-0.1in}
\begin{enumerate}
    \item \solon{}, a distributed training framework that exploits algorithmic redundancy to simultaneously provide Byzantine-resilience and communication efficiency. 
    \vspace{-0.07in}
    \item A concrete encoding and decoding mechanism, which is provably efficient and  achieves the optimal trade-off between Byzantine resilience, communication overhead, and computational cost.  
    \item Extensive experiments which show that \solon{} exhibits significant speedups as well as strong Byzantine-resilience over previous approaches.   
\end{enumerate}

The \solon{} framework can be used for any distributed algorithm which requires the sum of multiple functions, including gradient descent, SVRG \cite{johnson2013accelerating}, coordinate descent, and projected or accelerated versions of these algorithms. However, in this paper, we focus on mini-batch SGD.
The rest of the paper is organized as follows: Section \ref{Sec:SOLON:relatedwork} discusses related work. 
Section \ref{sec:solon} presents the \solon{} framework and theoretical guarantees. Experimental results are given in Section \ref{Sec:SOLON:Experiment}. 

\section{Related work}
\label{Sec:SOLON:relatedwork}
Byzantine fault tolerance against worst-case and/or adversarial failures such as system crashes, power outages, software bugs, and adversarial agents that exploit security flaws has been extensively studied since the 1980s~\cite{lamport2019byzantine}. In distributed machine learning, these failures may appear when a subset of compute nodes returns to the PS erroneous updates. It is well understood that first-order methods, such
as gradient descent or mini-batch SGD, are not robust to Byzantine errors; even a single erroneous update can introduce arbitrary errors to the optimization variables~\cite{blanchard2017machine,chen2017distributed}. At the same time, distributed model training suffers from communication overhead due to frequent gradient updates transmitted between compute nodes~\cite{LargeScaleDeepNet_Dean12,mcmahan2017communication,li2014communication,konevcny2016federated,dobriban2020wonder,dobriban2021distributed}. \solon{} aims at improving both Byzantine-resilience and communication-efficiency in distributed model training.

More recently, attention has turned to Byzantine-resilient distributed machine learning techniques.  Results show that while average-based gradient methods are susceptible to adversarial nodes, robust gradient aggregation methods can, in some cases, achieve better convergence while being robust to some attacks~\cite{chen2017distributed,el2019sgd,blanchard2017machine,xie2018zeno,xie2020fall,blanchard2017machine,yin2018byzantine,guerraoui2018hidden,jaggilearningfromhistory}. Despite
theoretical guarantees, the proposed algorithms often only ensure
a weak form of resilience against Byzantine failures, and can fail against strong Byzantine attacks~\cite{guerraoui2018hidden,xie2020fall,ALIE2019}. Another line of work proposes to use algorithmic redundancy to attain black-box Byzantine-resilience guarantees. However, many of these techniques require redundant computation from compute nodes (such as \draco{}~\cite{DRACO}) or place a heavy computation overhead on the PS (such as
\bulyan{}~\cite{guerraoui2018hidden}). Furthermore, they introduce a heavy communication overhead. \cite{detox2019} interpolates between \draco{} and robust aggregation methods for faster computation on both compute nodes and the PS. However, it does not mitigate the communication bottleneck, whereas \solon{} introduces both black-box Byzantine-resilience guarantee and communication-efficiency.  

Communication-efficient distributed machine learning has gained a lot of attention. Various methods propose to use gradient compression, \eg via quantization~\cite{seide20141,alistarh2016qsgd,wen2017terngrad,suresh2017distributed} or sparsification~\cite{stich2018sparsified,Wang18Atomo,lin2017deep,vogels2019powersgd,agarwal2020accordion} to enhance the communication efficiency. These methods massively compress the gradients, however, their Byzantine-resilience is not clear. 

The methods that are the most similar to \solon{} are \signum{}~\cite{bernstein2018signsgd} and the one proposed in~\cite{ghosh2020communication}. These methods introduce both communication-efficiency and Byzantine-resilience, however their Byzantine-resilience guarantees are not as strong as for \solon{} in theory, i.e. they typically use lossy compression for coding and decoding schemes, which only achieve approximated recovery under certain attacks. 
\cite{communciation_gradcoding_min_ye} considers the trade-offs between communication efficiency and straggler tolerance. 
On the other hand, our work focuses on improving communication efficiency in a Byzantine-aware distributed system.

\section{\solon{}}
\label{sec:solon}

In this section we give an overview of the \solon{} framework, discuss constraints on the encoding and decoding functions, and define optimal coding schemes.

The proofs are left to the appendix.
\subsection{Preliminaries}	
\paragraph{Basic notations.}
For a matrix $\mathbf{A}$, let $\mathbf{A}_{i,j}$, $\mathbf{A}_{i,\cdot}$, and $\mathbf{A}_{\cdot, j}$ denote entries, rows, and columns, respectively. More generally, $\mathbf{A}_{S,T}$ is the submatrix of $\mathbf{A}$ with rows indexed by  $S$ and columns indexed by $T$. The Hadamard, or elementwise, product $\mathbf{A} \odot \mathbf{B}$ of two matrices of the same size has entries $(\mathbf{A}\odot \mathbf{B})_{i,j} = \mathbf{A}_{i,j}\mathbf{B}_{i,j}$. Let $m$ be the dimension of the data, $n$ be the size of the training set, and $\mathbf{x}_i\in\mathbb{R}^m$, $i=1,\ldots,m$ be the data points. Let $\ell(\cdot;\cdot)$ be the loss function, $d$ be the model dimension, and ${\bf w}\in\mathbb{R}^d$ be the model parameters.  Let $\mathbf{1}_m$ and $\mathbf{1}_{n\times m}$ be the $m\times 1$ vector, and $n\times m$ matrix, of all ones, respectively. Similarly, let $\mathbf{0}_{m}, \mathbf{0}_{n\times m}$ contain zeros.  The empirical risk minimization (ERM) \cite{NIPS1991ERM,vapnik2013nature} objective is: 
$\min_{{\bf w}} {n}^{-1}\sum_{i=1}^n \ell({\bf w};\mathbf{x}_i).$
The most common current approach is to use first-order stochastic optimization to solve this, in particular mini-batch stochastic gradient descent (SGD). Starting at an initial point ${\bf w}_0$, we iterate 
${\bf w}_k = {\bf w}_{k-1} - \gamma/|S_k| \sum_{i \in S_k}\nabla\ell_{\bf w}({\bf w}_{k-1}; \mathbf{x}_{i}),$
where $S_k \subseteq \{1,\ldots, n\}$ is a random subset of size $B$ and $\gamma>0$ is the learning rate.

We relabel $S_k$ to $\{1,\ldots, B\}$ and denote $\nabla \ell({\bf w}_{k-1};\mathbf{x}_i)$ by $\mathbf{g}_i$. 

\paragraph{Distributed learning.}
We aim to compute $\mathbf{g}=\sum_{i = 1}^B \mathbf{g}_i$ in a distributed, {\it adversary-resistant}, and {\it communication efficient} manner. We consider a distributed training model where gradient computations are partitioned across $P$ compute nodes at each iteration. These operate on a potentially reduced dimension $d_c$ for communication efficiency, and we let the gradient \emph{compression ratio} be $\smash{r_c \triangleq d/d_c}$. After computing and summing up their assigned gradients, each node sends their answer back to the parameter server (PS). This sums them and updates the model. By applying \solon{}, we reduce the communication complexity for sending gradients to server from $\mathcal{O}(Pd)$ to $\mathcal{O}(Pd_c)$. The broadcast phase of sending aggregated gradients from server to compute nodes takes $\mathcal{O}(\log{(P)}d)$, which is not the major overhead. 

We assume that at most $s$ compute nodes are unreliable, Byzantine, or adversarial, and can send to the PS an arbitrary update. 
We consider the strongest possible adversaries: with infinite computational power, knowing the entire data set, the training algorithm, any defenses present in the system, and able to collaborate.

\subsection{Framework}	\vspace{-0.1in}
\solon{} is defined by the tuple, or \emph{mechanism}, $(\mathbf{A}, E, D)$, where $\mathbf{A}$ is an allocation matrix specifying how to assign gradients to nodes, $E$ are encoding functions determining how each compute node should locally encode its gradients, and $D$ is a decoding function determining how the PS should decode the output of the nodes. As an example, in Figure \ref{fig:RunningExample}(a), $A$ corresponds to the gradient computation assignment of the compute nodes, $E$ corresponds to the summation of the gradients by each node, and $D$ refers to the decoding phase at the PS. 
We generalize the scheme in Figure \ref{fig:RunningExample} to $P$ compute nodes and $B$ gradients. 

{\bf Allocation matrix, $\mathbf{A}$.} 
At each iteration of the training process, we assign the $B$ gradients to the $P$ compute nodes using a $P\times B$ {\it allocation matrix} $\mathbf{A}$, where $\mathbf{A}_{j,k}$ is equal to unity ("1") if node $j$ is assigned to the $k$th gradient ${\bf g}_k$, and zero ("0") otherwise. The support of $\mathbf{A}_{j,\cdot}$, denoted $\supp\left(\mathbf{A}_{j,\cdot}\right)$, is the set of indices of gradients evaluated by the $j$th node. For simplicity, we will assume $B=P$. 
Let $\|\mathbf{A}\|_0$ be to the $L_0$ norm of a matrix, i.e., the number of nonzero entries.
Following \cite{DRACO}, we define the \emph{redundancy ratio} of an allocation as the average number of gradients assigned to each compute node, or equivalently $r\triangleq  \|\mathbf{A}\|_0/P$.

We define the $d \times P$ matrix $\mathbf{G}$ with gradients as its columns: $\mathbf{G} \triangleq [\mathbf{g}_1,\mathbf{g}_2,\cdots, \mathbf{g}_P]$. The $j$th node first picks out its assigned gradients using the allocation matrix $\mathbf{A}$, computing a $d\times P$ gradient matrix $\smash{\mathbf{Y}_j \triangleq \left( \mathbf{1}_d \mathbf{A}_{j,\cdot} \right)\odot \mathbf{G}}$. 
The columns of this matrix are $\mathbf{g}_k$ if the $k$th gradient $\mathbf{g}_k$ is allocated to the $j$th compute node, \ie $\mathbf{A}_{j,k} \not = 0$, and zero otherwise.

{\bf Encoding Functions, $E$.} The $j$th compute node is equipped with an encoding function $E_j$ that maps the $d\times P$ matrix $\mathbf{Y}_j$ of its assigned gradients to a $d_c$-dimensional vector.
The $j$th compute node computes and sends $\smash{\mathbf{z}_j \triangleq E_j(\mathbf{Y}_j)}$ to the PS.
If the $j$th node is adversarial, then it instead sends $\smash{\mathbf{z}_j+\mathbf n_{j}}$ to the PS, where $\mathbf n_{j}$ is an arbitrary $d_c$-dimensional \emph{Byzantine} vector.
We let  $E = \{ E_1, E_2, \cdots, E_P\}$ be the set of local encoding functions.

{\bf Decoding Function, $D$}. The $d_c \times P$ matrix $\smash{\mathbf{Z}=\mathbf{Z}^{\mathbf{A},E,\mathbf{G}} \triangleq [\mathbf{z}_1, \mathbf{z}_2,\cdots, \mathbf{z}_P]}$ contains all outputs of the nodes. The $d_c \times P$ matrix $\smash{\mathbf{N} \triangleq [\mathbf{n}_1, \mathbf{n}_2,\cdots, \mathbf{n}_P]}$ contains all Byzantine vectors, with at most $s$ non-zero columns.
Then, the PS receives a $d_c\times P$ matrix $\smash{\mathbf{R} \triangleq \mathbf{Z} +\mathbf{N}}$, and computes a $d$-dimensional vector $\smash{\mathbf{u} \triangleq D(\mathbf{R})}$ using a decoding function $D$.

We require that the algorithm at the PS recovers the $d$-dimensional sum of gradients, $\mathbf{G} \mathbf{1}_P$:

\begin{definition}
	\solon{} with $(\mathbf{A}, E, D)$ can tolerate $s$ adversarial nodes, if for any $\mathbf{N} = [\mathbf{n}_1,\mathbf{n}_2,\cdots, \mathbf{n}_P]$ such that $ \left\vert{ \{j:\mathbf{n}_{j}\not=0\} }\right\vert \leq s$, we have $D(\mathbf{Z}+\mathbf{N}) = \mathbf{G}\mathbf{1}_P$.
\end{definition}

If we defend against the Byzantine attack, then the model update at each iteration is identical to the adversary-free setting. 
This implies that  convergence guarantees for the adversary-free case transfer to the adversarial case.

\subsection{Encoding and decoding functions}	\vspace{-0.1in}
What are the fundamental limits of the above allocation, encoding, and decoding schemes, in particular of the redundancy ratio used in allocation and the compression ratio used in encoding?  Perhaps surprisingly, \emph{the redundancy ratio does not depend on the compression ratio}.
The encoded gradients at each compute node can be arbitrarily compressed without affecting the our ability to tolerate Byzantine attacks.
The reason is that any $d$-dimensional real vector can be mapped one-to-one to a real number. This is stated in the following theorem.

\begin{theorem}\label{Thm:GenericBound}
	If there is a mechanism $(\mathbf{A}, E, D)$ of gradient allocation, encoding, and decoding with redundancy ratio $r$ tolerating $s$ adversarial nodes with a compression ratio $r_c$ of unity, then there is a mechanism $(\mathbf{A}', E', D')$ with redundancy ratio $r$ tolerating $s$ adversarial nodes for any compression ratio $r_c>0$.
\end{theorem}

However $(\mathbf{A}', E', D')$ is in a sense pathological and it is unclear if it can reduce the the number of bits communicated.
Therefore, we seek classes of \textit{regular} encoder and decoder functions $E,D$, to reduce communication cost.

\begin{definition}
	 A set of encoding functions $E$ is called {\em regular} if each output element of each function $E_j$ is a function of linear combinations of columns of the input. Formally, if $E_{j,v}$ is the $v$th element of $E_j$, then
	 $(\mathbf{A}, E, D)$ is regular if there exists a $d\times P$ matrix $\mathbf{U}_{j,v}$ and functions $\smash{\hat{E}_{j,v}}$, such that $\smash{E_{j,v}(\mathbf{Y}_j) = \hat{E}_{j,v}\left(\mathbf{1}_d^T\left(\mathbf{U}_{j,v} \odot \mathbf{Y}_{j}\right)\right)}$.
\end{definition}

When $d=1$, $\mathbf{Y}_j$ has only one row and $\mathbf{1}_{d}=1$ is a scalar.
Thus, $E_{j,v}(\mathbf{Y}_j) = \hat{E}_{j,v}(\mathbf{U}_{j,v} \odot \mathbf{Y}_j)$ implies that $E$ is an arbitrary function of $\mathbf{Y}_j$.
When $d>1$, each output coordinate only depends on linear combinations of input columns.
Since linear combinations do not introduce extra bits, this allows practical communication compression. We will study regular encoders $E$.

\paragraph{Redundancy Bound.}
We first study redundancy requirements for exact recovery of the sum of gradients with $s$ adversaries and compression ratio $r_c$. 
\begin{theorem}\label{Thm:RedundancyRatioBound}
A mechanism $(\mathbf{A}, E, D)$ of gradient allocation, regular encoding, and decoding with compression ratio $r_c$ tolerating $s$ adversarial nodes must have a redundancy ratio $r \geq 2s+r_c$.
\end{theorem}

Thus, for any regular encoder, each gradient has to be replicated on average at least $2s+r_c$ times to defend against $s$ adversarial nodes with a communication compression ratio of $r_c$. If a mechanism tolerates $s$ adversarial nodes with a communication compression ratio of $r_c$, by Theorem \ref{Thm:RedundancyRatioBound}, each compute node encodes at least $(2s+r_c)$ $d$-dimensional vectors on average. If the encoding has linear time complexity, then each encoder requires $O((2s+r_c)d)$ operations in the worst case. If the decoder $D$ has linear time complexity, then it requires at most $O(Pd_c)$ operations in the worst case, as it needs to use the $d$-dimensional input from all $P$ compute nodes. This gives a computational cost of $O(Pd_c)$, which is less than the bound $\mathcal{O}(Pd)$  for the repetition code in \cite{DRACO}.

To better understand the lower bound, we give an equivalent formulation it in the language of linear algebra. Let $g$ be one of the $d\times 1$ dimensional gradients. Suppose it gets sent to $r$ nodes. Then the potentially  corrupted output of the linear encoder at each node $j$  can be represented by $R_j = Z_j g + n_j$, and $n_j$ is the noise here for $j=1,\ldots, r$, where each $Z_j$ is a $d_c \times d$ dimensional matrix. If the number of adversaries is at most $s$, then at most $s$ vectors $n_j$ are nonzero. We call this set of perturbation $B_{s,r}$. Thus, the goal is to design the $r$ matrices $Z_j$, such that for any collection of vectors $n_j$ at most $s$ of which are nonzero, it is possible to recover $g$ from the observations $(Z_j,r_j)$, $j=1,\ldots, r$.
Let $R:=(R_1,\ldots, R_r)$ and $n:=(n_1,\ldots,n_r)$, which we view as a concatenation of vectors belonging to the allowed set $B_{s,r}$. For the recovery of $g$ to be possible, we need that if $R(g,n) = R(g',n')$ for $g,g'\in \R^d$ and $n,n'\in B_{s,r}$, then $g=g'$. 

We can write $R$ as a linear function $R = Zg+n$, where $Z$ is an $rd_c\times d$ concatenation of the matrices $Z_i$. Thus, we can write $R(g,n) = R(g',n')$ as
$Zg+n=Zg'+n' \iff Z(g-g')=n-n'.$
Moreover, we have $n''  = n-n' \in B_{2s,r}$, because at  most $2s$ of its $P$ sub-vectors of size $d_c$ are nonzero. Clearly, all vectors in $B_{2s,r}$ can be written in this form.  Denoting $x = g-g'$, the problem is to understand when there exists a matrix $Z$ such that for all $x \in \R^d$, we have $Zx \notin B_{2s,r}$. Now, $B_{2s,r}$ is a union of several $2sd_c$-dimensional linear subspaces in $rd_c$ dimensions. Moreover, $Zx$ belongs to $span(Z)$, which is an at most $d$-dimensional subspace in $rd_c$ dimensions. Thus, for this to be possible, by counting dimensions we obtain that we need $rd_c\ge 2sd_c +d$. Since $r_c=d/d_c$, this is equivalent to $r\ge 2s +r_c$. This finishes the proof of the fundamental lower bound on the redundancy.

\subsection{Optimal Coding Schemes}
\label{ocs}
	\vspace{-0.1in}
Can we achieve the optimal redundancy bound with linear-time encoding and decoding? 
More formally, can we design a tuple $(\mathbf{A},E,D)$ with redundancy ratio $r=2s+r_c$ and computation complexity $\mathcal{O}((2s+r_c)d)$ at the compute nodes and $\mathcal{O}(Pd_c)$ at the PS?
We give a positive answer by designing certain {\em linear block codes} that match the above bounds. 

This is a challenging problem. In fact, we can show (see Appendix \ref{sp}) that this question is \emph{exactly equivalent to a sparse recovery problem}, where we wish to recover an unknown sparse vector from linear combinations of a fixed set of vectors \cite{chen1994basis,chen2001atomic,elad2010sparse}. 
With this lens, our algorithms are related to the classical Prony's method in signal processing \cite{hurst1987scattering,Prony}. However, a key difference is that in our case, we have a \emph{structured set of perturbations}, where entire sub-vectors corresponding to gradients are perturbed at the same time. We leverage this to develop algorithms faster than Prony's method.
\vspace{-0.1in}
\paragraph{Linear Block Code.} 
We focus on the case when $2s+r_c$ divides $P$; otherwise we can change $P$ or $r_c$ until $2s+r_c$ divides $P$. 
Divide the compute nodes into $q:=P/(2s+r_c) = P/r$ ``blocks'' or groups.
We assign each node in the same block to compute the same gradients. 
Each node sends some linear combination of coordinates of the assigned gradients to the PS (``linear'').
The PS solves systems of linear equations to get the desired gradient sum.  
Following the convention of coding theory \cite{abclbc}, we call our approach \textit{linear block codes}.

The linear block code $(\mathbf{A}, E, D)=(\mathbf{A}^{LBC}, E^{LBC}, D^{LBC})$ is defined as follows. 
The assignment matrix is $\mathbf{A} = \mathbf{I}_q \otimes \mathbf{1}_{r \times r}$.

The $j$th compute node first selects its allocated gradients $\mathbf{Y}_j=\left( \mathbf{1}_d \mathbf{A}_{j,\cdot} \right)\odot \mathbf{G}$.
Its encoder function sums up the allocated gradients into $\mathbf{Y}_j \mathbf{1}_P$, then computes and sends $\mathbf{z}_j = E_j(\mathbf{Y}_j) = \mathbf{W}_j \mathbf{Y}_j\mathbf{1}_P$  to the PS, where $\mathbf{W}_j$ is a $d_c \times d$ matrix. 

The decoder function, summarized in Algorithm \ref{Alg:DecoderFunction_LinearBlock}, partitions the received updates into the $q$ blocks computing identical gradients.
For each, a block decoder $\psi(\cdot,\cdot)$ is called to recover the sum of all gradients in this group.
\begin{algorithm}[htbp]
	\SetKwInOut{Input}{Input}
	\SetKwInOut{Output}{Output}
	\Input{Received $d_c \times P$ matrix $\mathbf{R}$}
	\Output{Desired gradient summation $\mathbf{u}$}
	Let $\mathbf{R} = \left[\mathbf{R}_1, \mathbf{R}_2, \cdots, \mathbf{R}_{q}\right]$, where each $\mathbf{R}_j$ is a $d_c\times r $ matrix
	
    \For{$j=1$ \KwTo $q$ }
    {
         \textrm{ } $ V = \phi(\mathbf{R}_j,j)$ // Locate the adversarial nodes\\
		$U = \{1,2,\cdots,r\} \setminus V$ // Non-adversarial nodes\\
    $\mathbf{u}_j = \psi(\mathbf{R}_j,j, U)$ // Decode each block using the non-adversarial nodes
    }
	Compute and return $\mathbf{u} = \sum_{i=1}^{q}\mathbf{u}_i$
	\caption{Decoder Function $D$.}
	\label{Alg:DecoderFunction_LinearBlock}
\end{algorithm}

There are three questions left: (i) how $\mathbf{W}_j$ is constructed, (ii) how the adversarial node index location function $\phi(\cdot)$ works, and (iii) how the block decoder function $\psi(\cdot,\cdot)$ works.

Given any distinct nonzero scalars $w_1, w_2, \cdots, w_P$, we propose to construct $\mathbf{W}_j$ as $\mathbf{W}_j \triangleq \mathbf{I}_{d_c}\allowbreak$ $\otimes [\allowbreak1,\allowbreak w_j,\allowbreak w_j^2,\allowbreak \cdots,\allowbreak w_j^{r_c-1}]$. The adversarial node index locating function $\phi(\cdot)$ works as follows.
Given the $d_c \times r$ matrix $\mathbf{R}_j$ received from the compute nodes, we first generate a $1\times d_c$ random vector $ \mathbf{f} \sim \mathcal{N}(\mathbf{1}_{1\times d_c},\,\mathbf{I}_{d})$,
and then compute $\smash{\mathbf{r}_{j,c} \triangleq \mathbf{f} \mathbf{R}_j}$.
Next, we obtain an $r$-dimensional vector $\mathbf{a} = [a_1, a_2,\cdots, a_{r}]$ by  solving the linear system
$\begin{bmatrix}
     \hat{\mathbf{W}}_{j,r_c+s-1},\allowbreak  & -\hat{\mathbf{W}}_{j,s-1} \odot\allowbreak \left( \mathbf{r}_{j,c}^T \mathbf{1}_{s}^T\right)
\end{bmatrix}
    \mathbf{a}= \mathbf{r}_{j,c} \odot \left[\hat{\mathbf{W}}_{j,s}\right]_{\cdot,s}$,
 where 
	\[
\hat{\mathbf{W}}_{\mathit{j},v} \triangleq
\begin{bmatrix}
1 & w_{(j-1)r+1} &  w_{(j-1)r+1}^2 &\cdots &  w_{(j-1)r+1}^{v} \\
1 & w_{(j-1)r+2} &  w_{(j-1)r+2}^2 &\cdots &  w_{(j-1)r+2}^{v} \\
\vdots & \vdots & \vdots & \vdots & \vdots\\
1 & w_{jr} &  w_{jr}^2 &\cdots &  w_{jr}^{v}
 \\
\end{bmatrix}.
\\
\]
Finally compute $P_j(w) \triangleq (\sum_{i=0}^{r_c+s-1} a_{i+1} w^i)/(w^s+\sum_{i=0}^{s-1} a_{i+r_c+s+1} w^i) $,
and return $V=\{i|P_j(w_{(r-1)j+i}) \not =\left[ \mathbf{r}_{j,c}\right]_i\}$.
The decoding function $\psi(\cdot, \cdot)$ computes and returns $\textit{vec}\left(\left[\mathbf{R}_{j}\right]_{\cdot, U} \left[\hat{W}_{j,r_c-1}\right]_{U,\cdot}^{-1,T}\right)$ given the non-adversarial node indices $U$.

The following lemma ensures that the Byzantine nodes are correctly found.
\begin{lemma}\label{Thm:LBCDetection}
	Suppose $\left\vert{\{j:\|\mathbf{n}_{j}\|_0 \not=0\}}\right\vert$ $\leq s$ and $r\geq r_c+2s$. 
	Then $\phi(\mathbf{R}_j,j) =  \{i:\|\mathbf{n}_{j(r-1)+i}\|_0 \not=0\}$ with probability equal to unity. 
\end{lemma}
The next lemma demonstrates that within each group the gradient is correctly recovered.
\begin{lemma}\label{thm:LBClocalcorrect}
	If $U$ consists solely of at least $r-s$ non-adversarial nodes, then  with probability equal to unity, $\mathbf{u}_j = \sum_{k=(j-1)r+1}^{jr}\mathbf{y}_{k}$.
\end{lemma}

Combing the above two results,  we show that the linear block code tolerates any $s$ adversaries, achieving the optimal redundancy and compression ratio with linear-time encoding and decoding.

\begin{theorem}\label{Thm:LinearBlockCodeOptimality}
	The linear block code  $(\mathbf{A}, E, D)$ tolerates any $s$ adversaries with probability equal to unity, and achieves the redundancy ratio  bound. For $d\gg P$, its encoding and decoding achieve linear-time computational complexity. 
\end{theorem} 

Theorem \ref{Thm:LinearBlockCodeOptimality} shows that the linear block code is information theoretically tight and enjoys a small computational overhead even for large ML models.

\vspace{-0.1in}
\section{Experiments}\label{Sec:SOLON:Experiment}
\vspace{-0.1in}
Now we present an empirical study on \solon{} compared with several existing methods including \draco{}~\cite{DRACO}, \bulyan{}~\cite{guerraoui2018hidden}, and \signum{}~\cite{bernstein2018signsgd}. 
Across diverse ML models trained on real world datasets, we have found 1) that \solon{} results in significant speedups over existing methods, including $10\times$ faster than \bulyan{} and $80\%$ faster than \draco{} while reaching the same accuracy, and  2) that \solon{} consistently leads to successful convergence for all Byzantine attacks considered, while previous approaches may fail on different attacks (\eg \signum{} on constant attack, and \bulyan{} on "A little is enough" (ALIE) attack~\cite{ALIE2019}). 

\begin{table}[b]
\centering
\begin{minipage}{.58\linewidth}
    \medskip
	\centering
	\caption{Summary of the datasets, models, and hyper-parameters used in our experiments.}
	\scalebox{0.95}{
	\begin{tabular}{cccc}
		\toprule Dataset
		& CIFAR-10 & SVHN & \makecell{WikiText-2}\bigstrut\\
		\midrule
		\midrule
		\# data points & 60,000 & 600,000 & 2,551,843 \bigstrut\\
		Model & ResNet-18 & VGG13-BN  & LSTM \bigstrut\\
		\# Parameters & 11,173k & 9,923k  & 7,332k \bigstrut\\
		Optimizer & SGD & SGD &SGD \bigstrut\\
		Batch Size & 120 & 120 & 60 \bigstrut\\
		\bottomrule
    	\end{tabular}
    	}%
		\label{Tab:solon:DataStat}%
\end{minipage}
\hskip 10pt
\begin{minipage}{.37\linewidth}
    \medskip
	\centering
    \caption{The size of gradients to transmit per worker before and after compression (MB, $10^6$ bytes). }
    \scalebox{0.86}{
	
	\begin{tabular}{cccc}
		\toprule Model & 
		  ResNet18 & VGG13 & LSTM \bigstrut\\
		\midrule
		\midrule
		Size & 89.6 & 79.4 & 58.8   \bigstrut\\
		$r_c=6$ & 14.9 & 13.2 & 9.78 \bigstrut\\
		$r_c=8$ & 11.2 & 9.92 & 7.35  \bigstrut\\
		$r_c=10$ & 8.96  & 7.94  & 5.88  \bigstrut\\
		
		\bottomrule
	\end{tabular}
	}%
	\label{Tab:compressionsize}
	\end{minipage}
	\vspace{-3mm}
\end{table}

\begin{figure}[t] 
\centering
\includegraphics[width=0.94\textwidth]{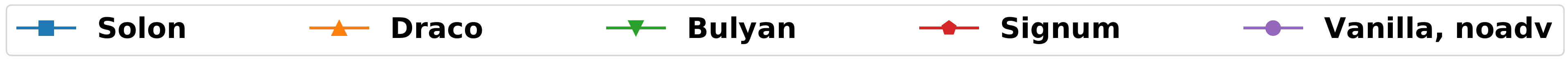}\\
\vspace{-2mm}
\subfigure[Accuracy vs steps, rev-grad]{\includegraphics[width=0.325\textwidth]{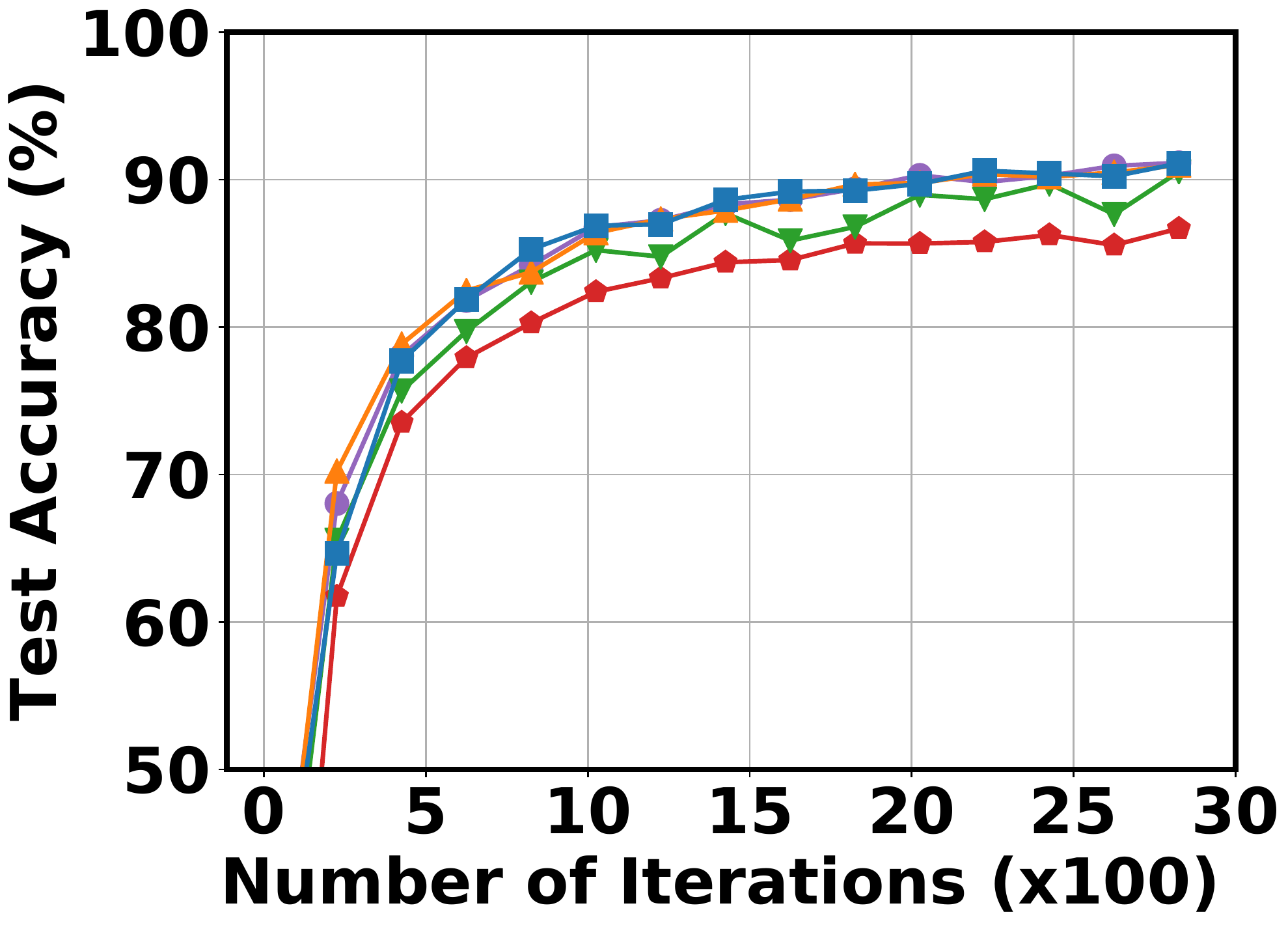}}
\subfigure[Accuracy vs steps,  constant]{\includegraphics[width=0.31\textwidth]{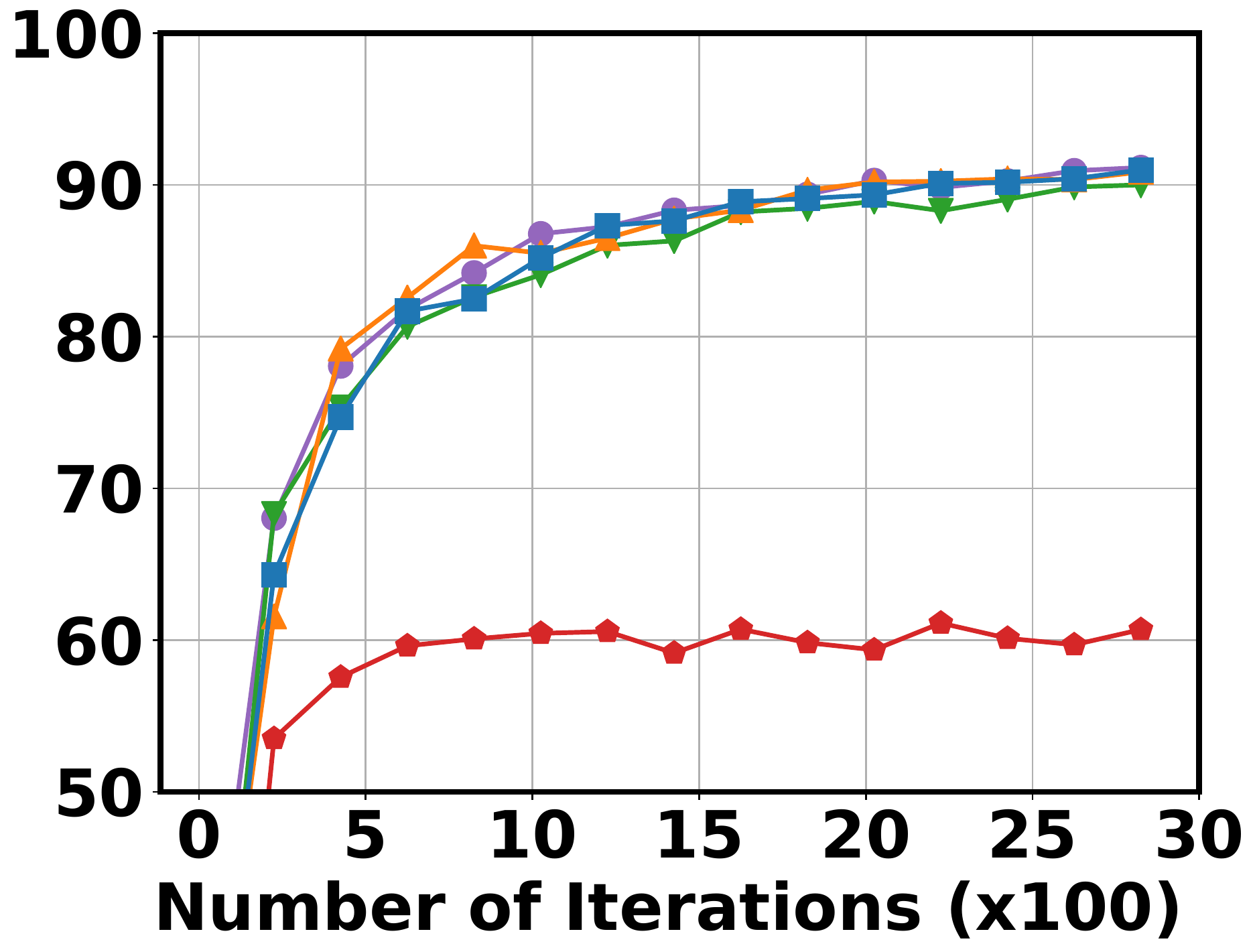}}
\subfigure[Accuracy vs steps, ALIE]{\includegraphics[width=0.31\textwidth]{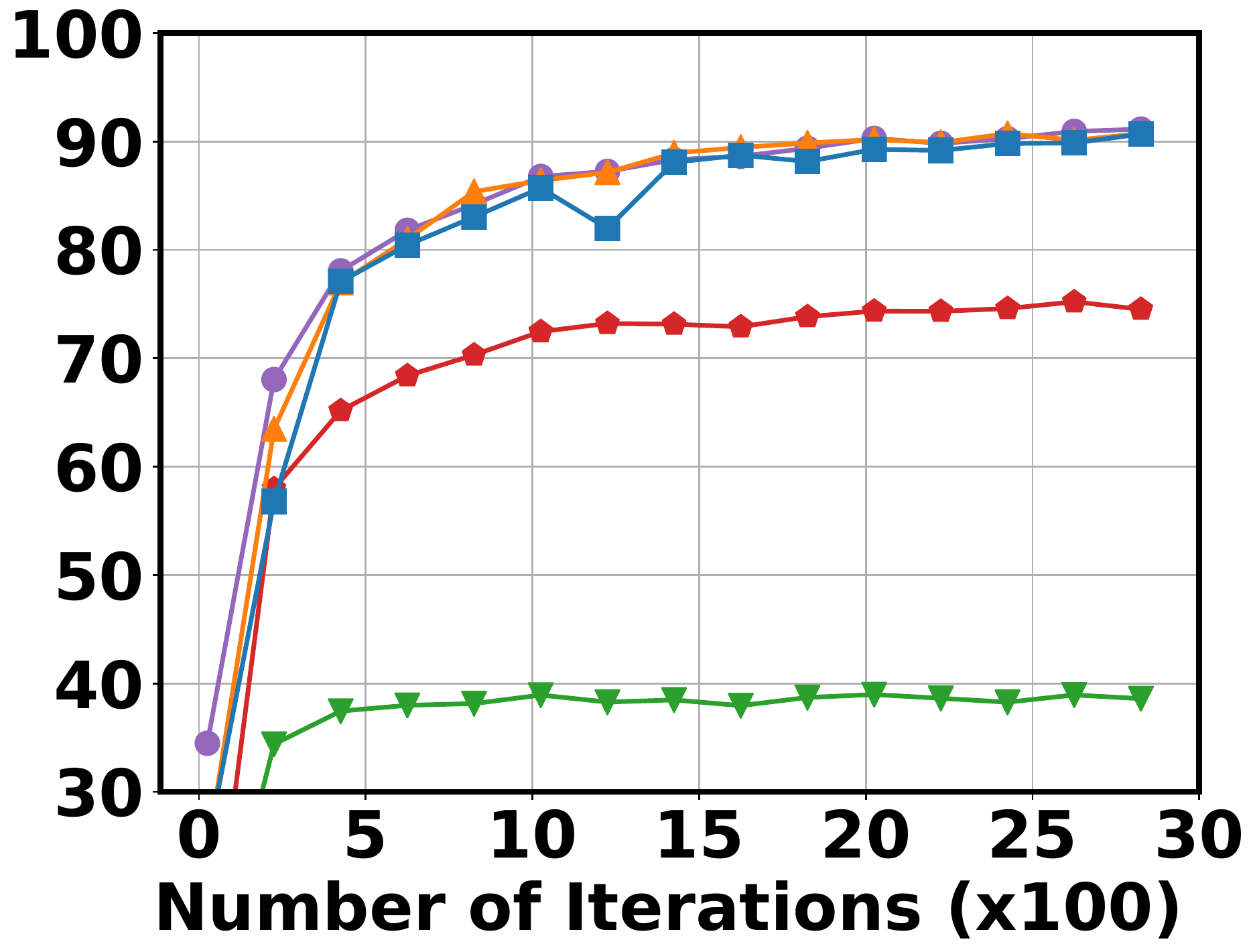}}\\
\vspace{-3mm}
\subfigure[Accuracy vs time, rev-grad]{\includegraphics[width=0.325\textwidth]{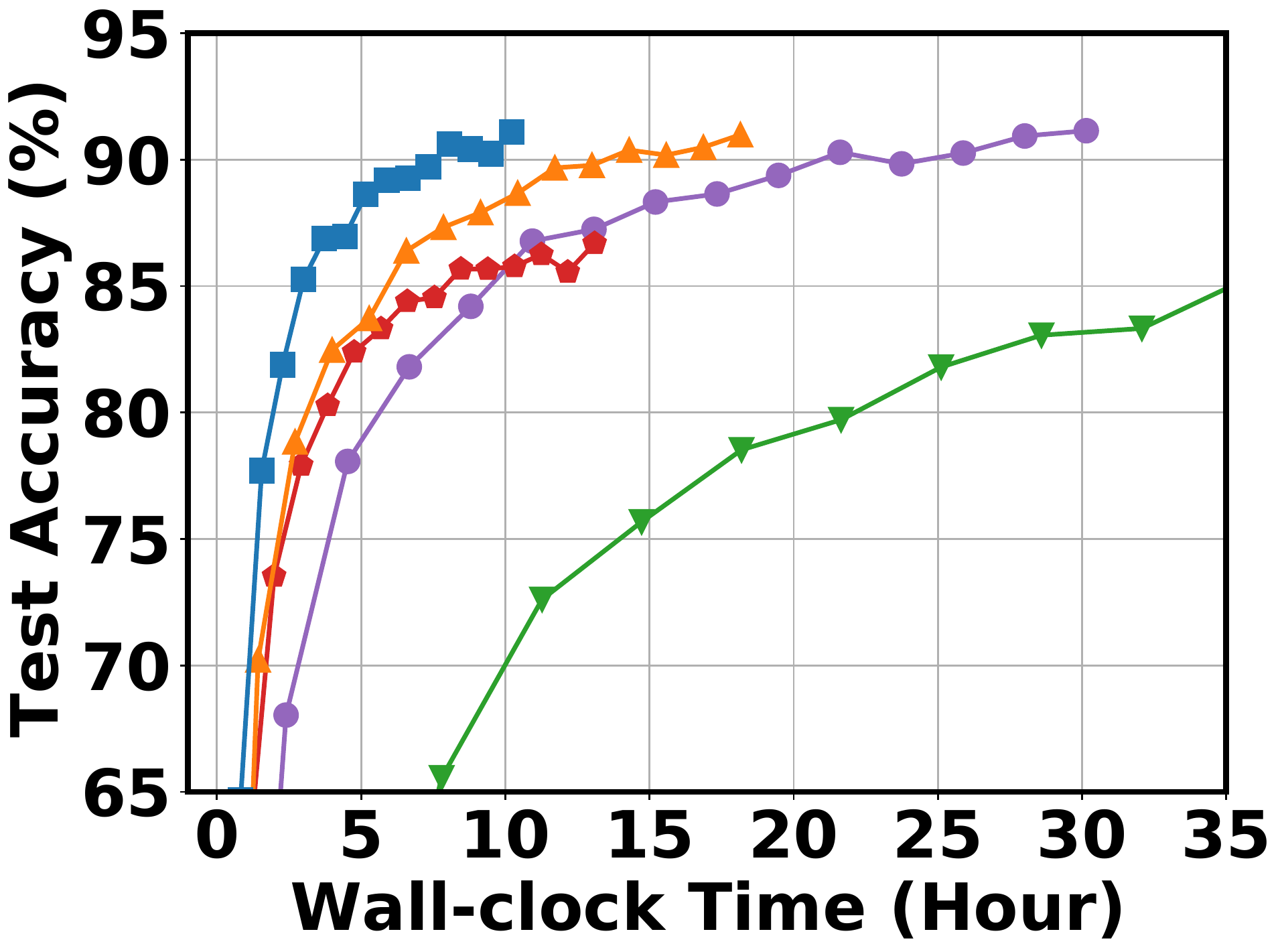}}
\subfigure[Accuracy vs time, constant]{\includegraphics[width=0.31\textwidth]{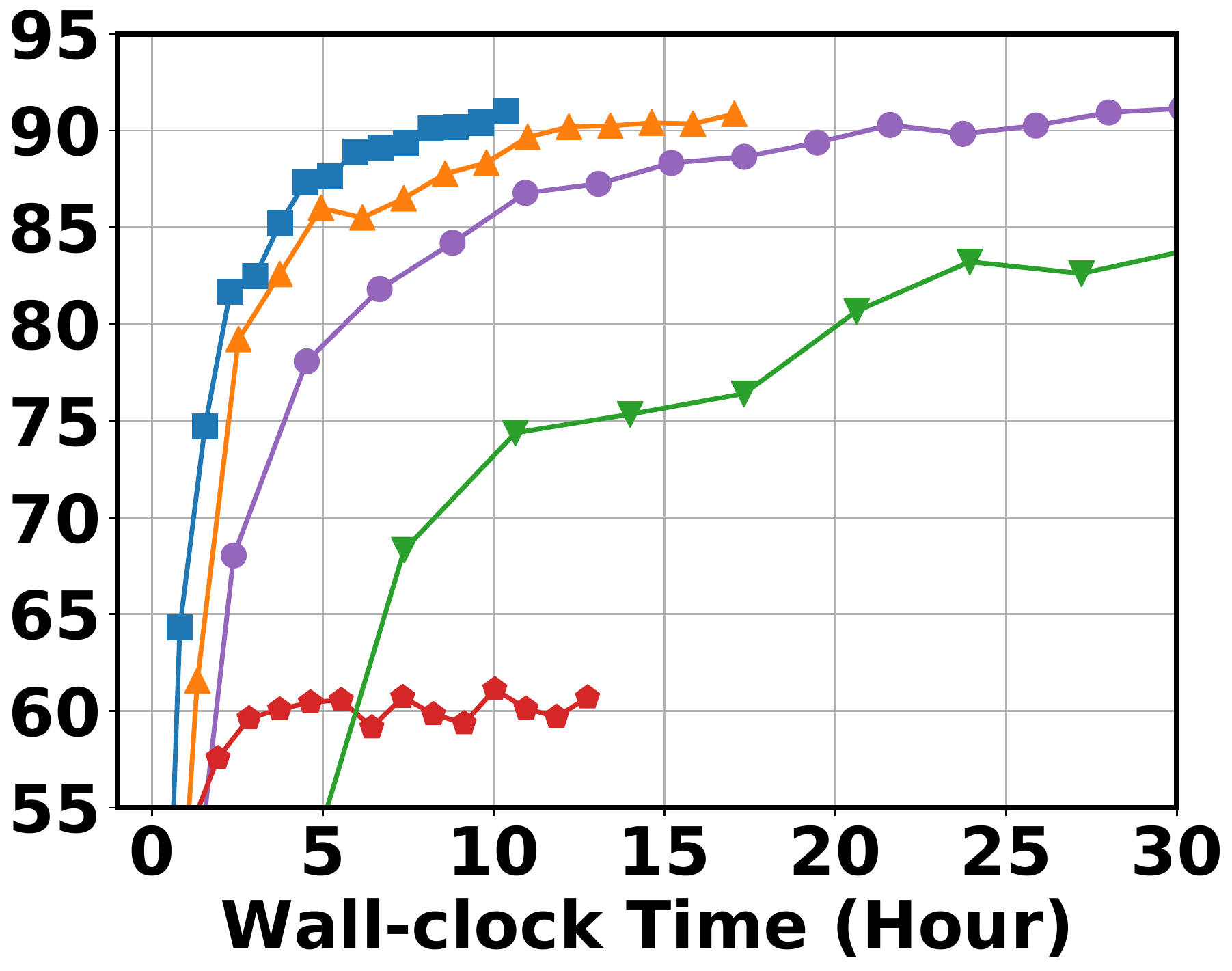}}
\subfigure[Accuracy vs time, ALIE]{\includegraphics[width=0.31\textwidth]{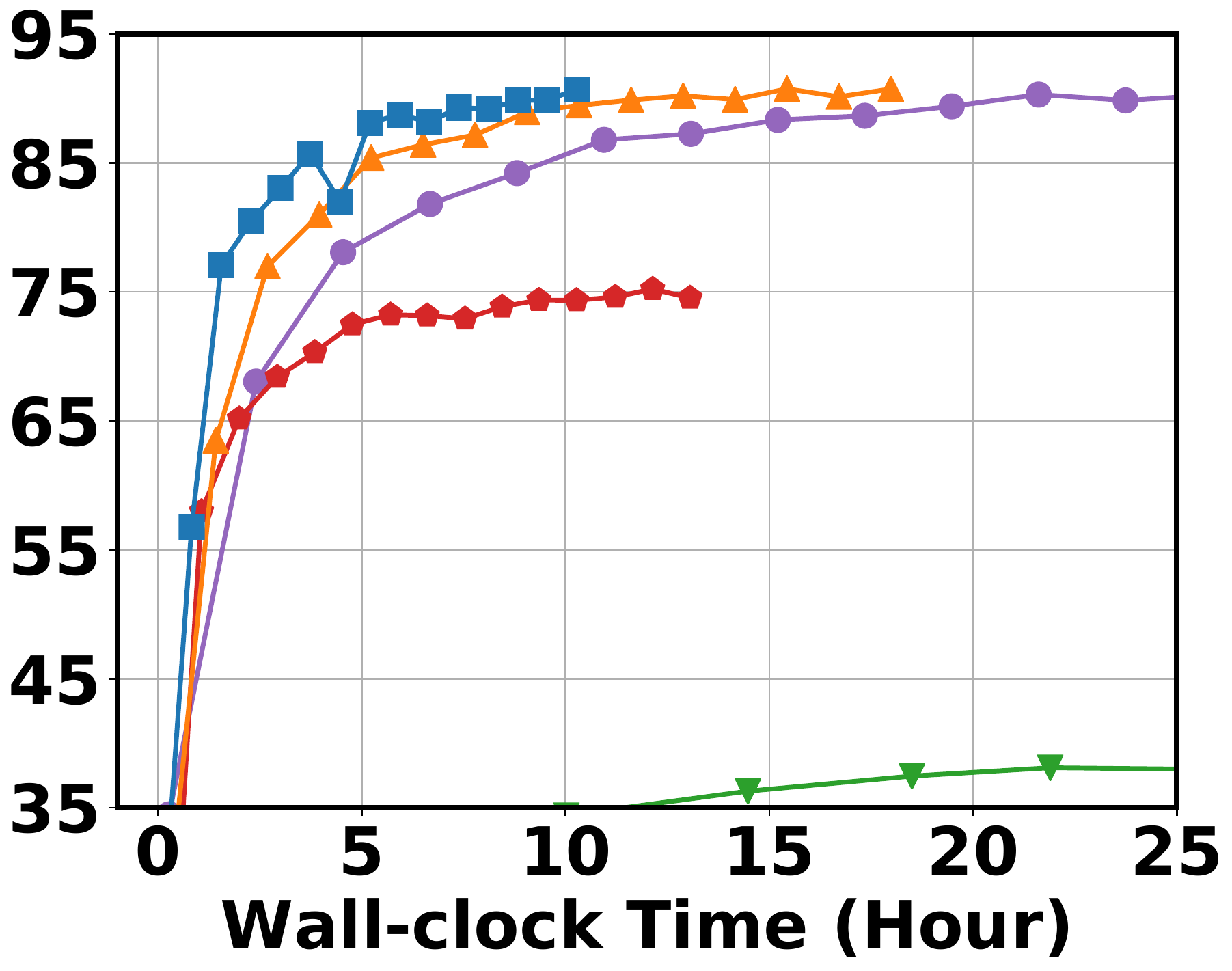}}
\vspace{-3mm}
\caption{\footnotesize{End to end convergence performance of \solon{} and other baselines on ResNet-18 and CIFAR-10. (a)-(c): Comparison of test accuracy vs. the number of iterations between 
\solon{} $r_c=10$ and other methods under different attacks. (d)-(f): Test accuracy vs. running time of \solon{} and other methods. Vanilla SGD simply averages gradients received on PS and is tested without adversary. Accuracy may fluctuate occasionally due to randomness and lr adjustments.} }
\vspace{-5mm}
\label{fig:against-defense}
\end{figure}

\paragraph{Experimental setup.} 
We implement \solon{} in PyTorch~\cite{paszke2019pytorch} with MPI~\cite{DALCIN20051108}.
The experiments were conducted on a cluster of 50 real machines from Cloudlab  \cite{CloudlabDuplyakin} with 1 Gbps network speed and 100 virtual compute nodes. We trained three large scale models, namely, ResNet-18~\cite{ResNet} on CIFAR-10~\cite{Cifar10}, VGG13-BN~\cite{VGGverydeepconv} on SVHN~\cite{svhn}, and a two-layer stacked LSTM \cite{LSTMnips96} (nhid=200) on WikiText-2~\cite{wikitext-2}, respectively.  
The details are summarized in Table \ref{Tab:solon:DataStat}.
For comparison with \solon{}, we also evaluate two robust aggregator-based approches, \bulyan{}, \signum{}, and an algorithmic redundancy-based approach, \draco{}.  \solon{} splits the virtual machines evenly into 5 groups, each with 20 redundant machines. To compare with the best possible performance of \draco{}, we set the $r=11$ and only use 5 groups for \draco{} to reduce its communication overhead. 
More details are in the appendix.

\vspace{-2mm}
\paragraph{Attacks.} We use three different attacks:  reverse gradient,  constant, and ALIE (``A little is enough''~\cite{ALIE2019}). 
In the reverse gradient attack (rev-grad), Byzantine nodes always send $\kappa$ times the true gradient to the PS. 
In the constant attack, Byzantine nodes always send a constant multiple $c$ of the all-ones vector.
In the experiments shown, $\kappa = -100$ and $c=-100$.

In ALIE, Byzantine nodes use local information to estimate the mean and variance of the gradients computed at the other nodes, and then manipulate the gradient as $ \hat{\mu} + z \cdot \hat{\sigma}$ where $\hat{\mu}$ and $\hat{\sigma}$ are the mean and standard deviation of the gradients estimated by Byzantine nodes and $z$ is an adjustable hyper-parameter that adds an unnoticeable perturbation to disrupt the aggregation. In experiments, we set $z=1$.   
At each iteration, we randomly select $s = 5$ compute nodes as adversaries. 

\vspace{-2mm}
\paragraph{End to end performance.} We start by evaluating \solon{}'s end to end performance along with the baseline methods under different attacks, which is shown in Figure~\ref{fig:against-defense}.
We first note that previous approaches may result in significant accuracy loss under certain attacks. For example, \signum{}'s accuracy is 30\% worse than   the Byzantine-free vanilla SGD under constant attack (Figure~\ref{fig:against-defense}(b)), and ALIE attack leads a 50\% accuracy drops for \bulyan{} (Figure~\ref{fig:against-defense}(c)).
Nevertheless, across different attacks, \solon{} consistently converges and matches the accuracy performance of the vanilla SGD in a Byzantine-free environment. 
This is primarily due to \solon{}'s black box performance guarantee.

\begin{figure}[t]
\vspace{-4mm}
\centering
\subfigure[ResNet18, CIFAR-10]{\includegraphics[width=0.33\textwidth]{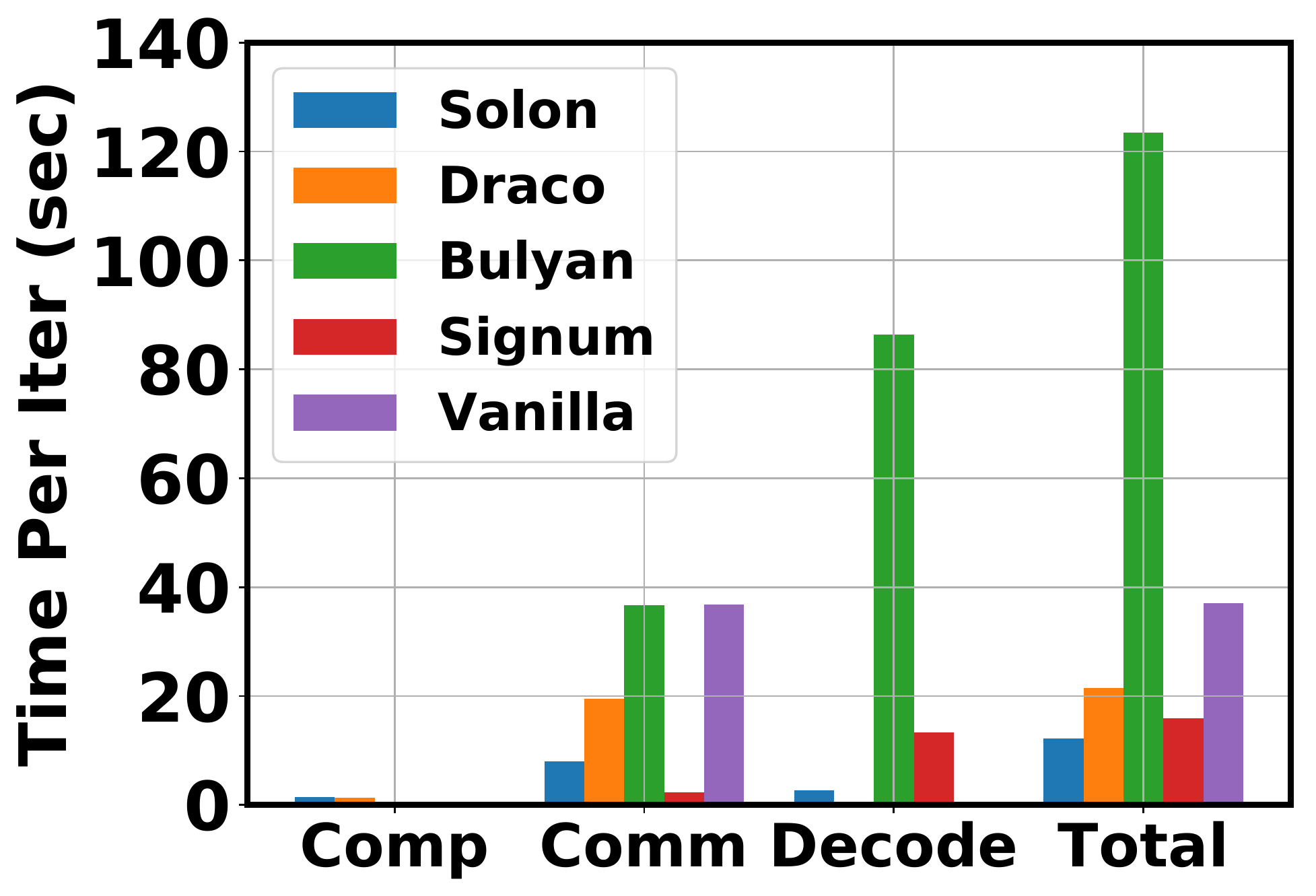}}
\subfigure[VGG13-BN, SVHN ]{\includegraphics[width=0.31\textwidth]{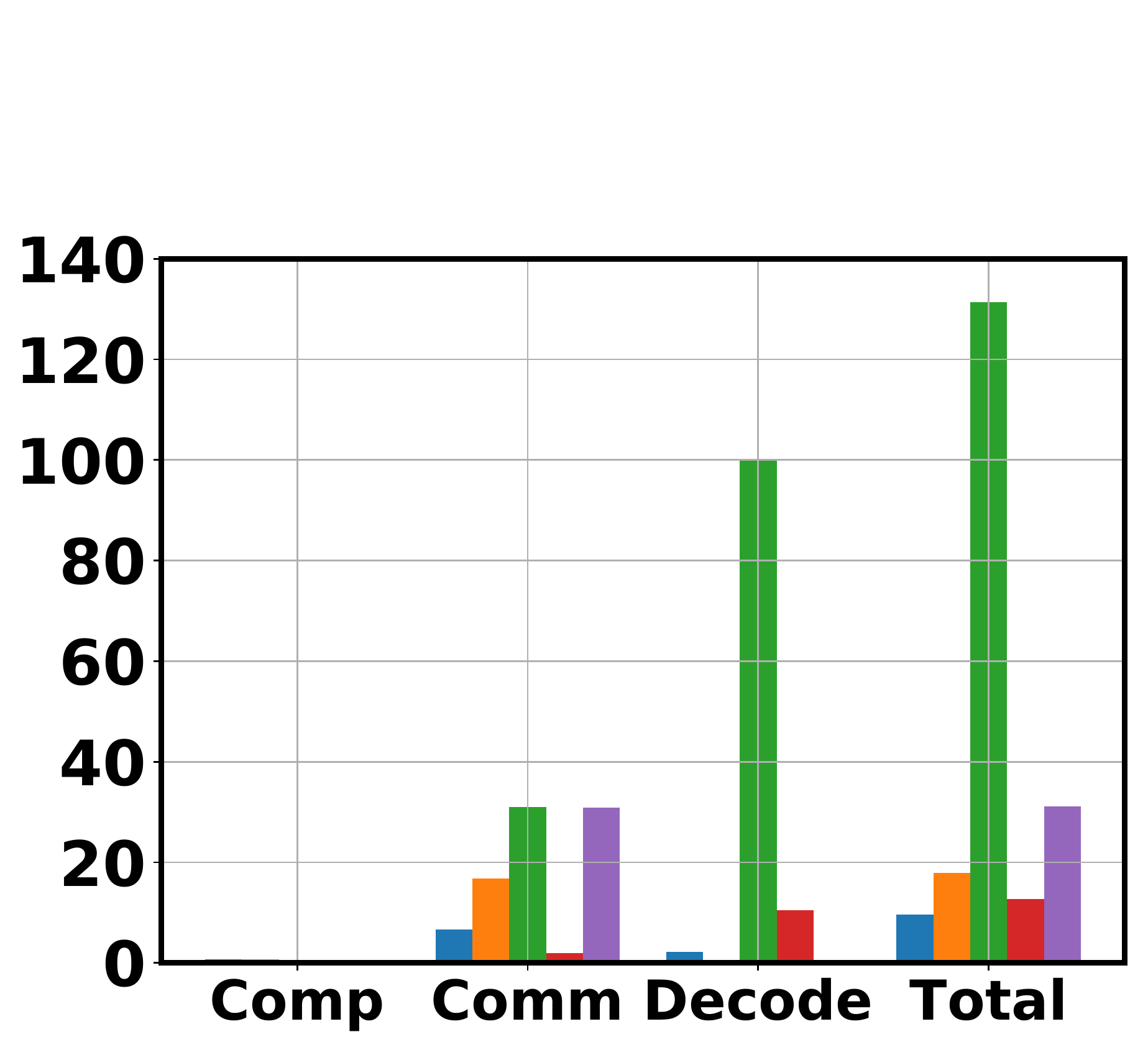}}
\subfigure[LSTM, WikiText-2]{\includegraphics[width=0.30\textwidth]{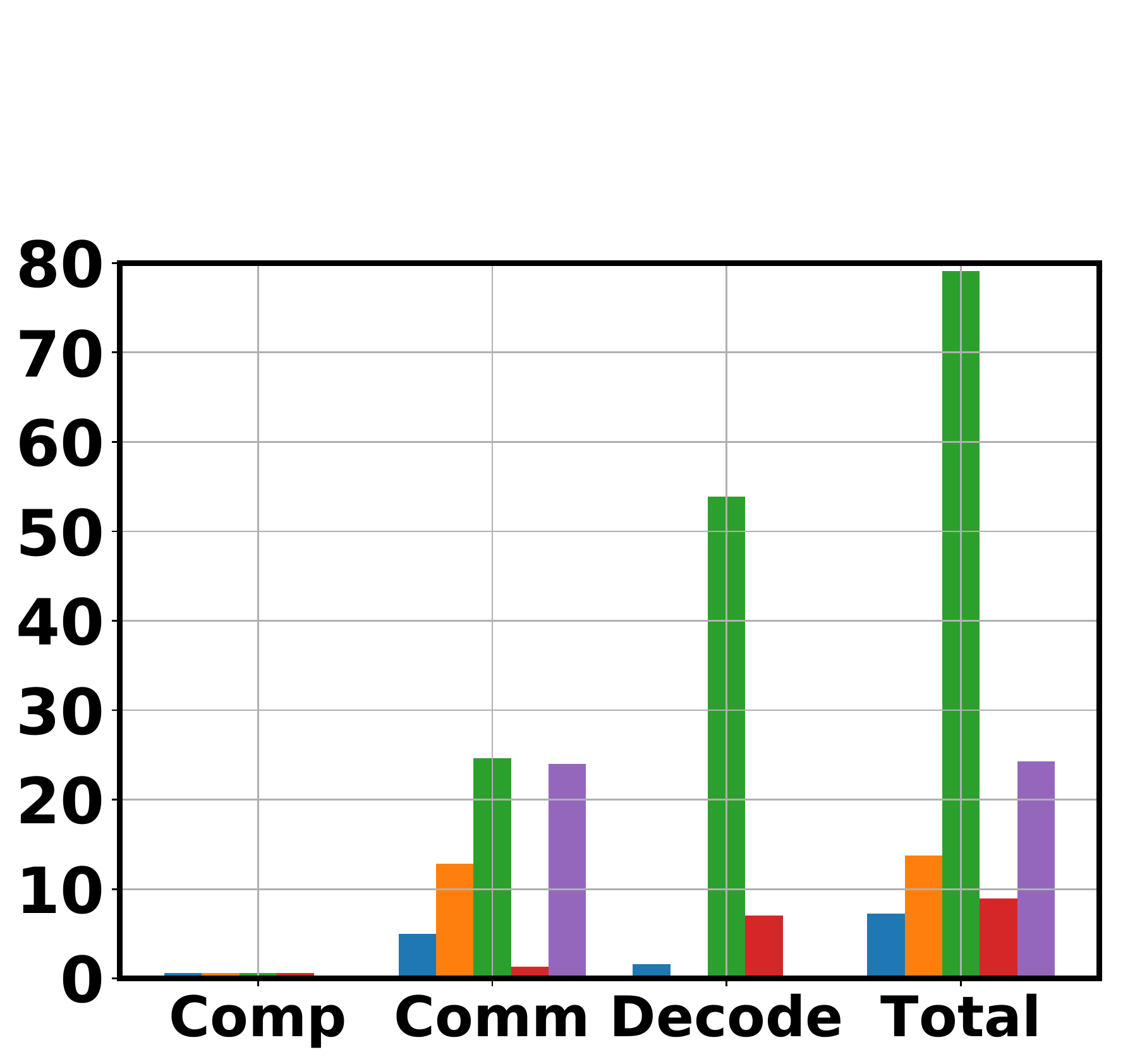}}
\vspace{-3mm}
\caption{\footnotesize{Time breakdown per iteration of \solon{} $r_c=10$ and all baseline methods over three models+datasets under rev-grad. The types of attack will not affect most of the times. Notice that even if \signum{} is the fastest in time per iteration among those methods, and even faster than vanilla SGD, its accuracy is sacrificed and thus \solon{} still has the best convergence performance.}} 
\label{fig:timebreakdown}
\vspace{-5mm}
\end{figure}

\begin{figure}[t]
  \centering
  \subfigure[Speed up ($\times$) of \solon{}]{\includegraphics[width=0.42\textwidth]{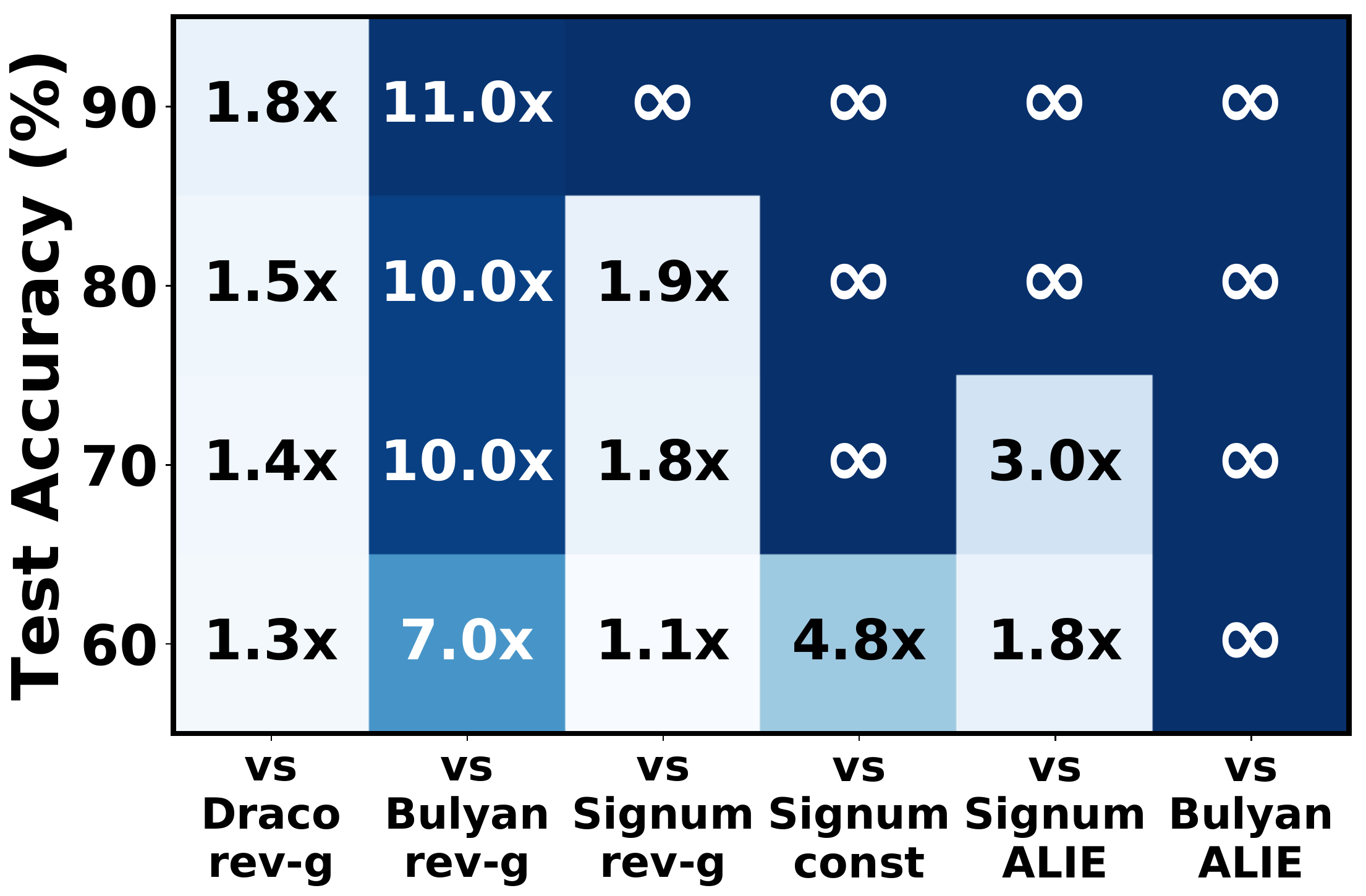}}
  \hskip 20pt
  \subfigure[Time breakdown of \solon{}]{\includegraphics[width=0.40\textwidth]{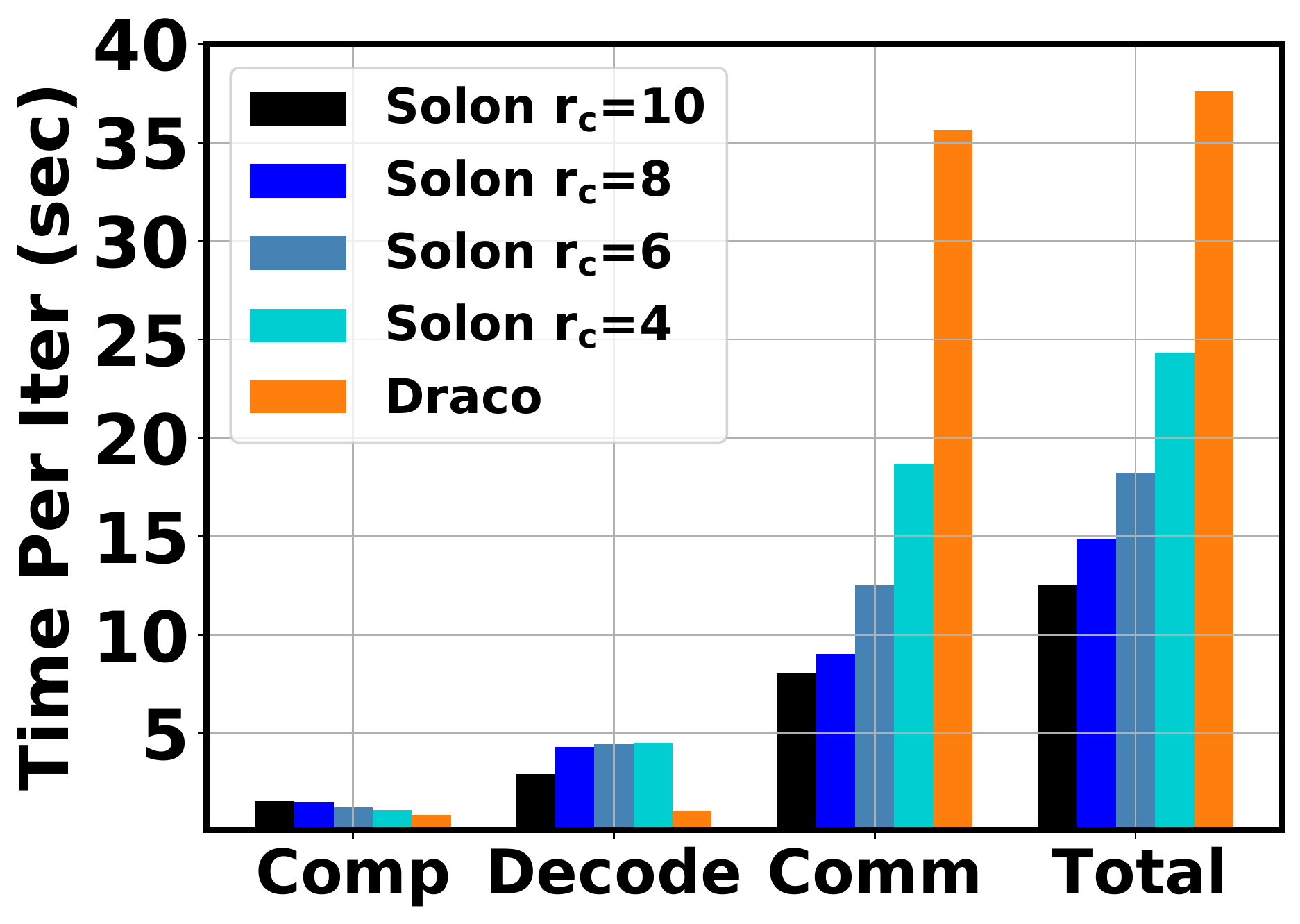}}
  \vspace{-3mm}
  \caption{\footnotesize{(a) The speed up ($\times$) of \solon{}, $r_c=10$ vs other baselines on converging to certain test accuracy level on ResNet-18+CIFAR-10 under selected attacks. Values are approximation. $\infty$ means the method does not converge to the accuracy in the experiment. (b) Time breakdown of \solon{} by varying $r_c$, fixing $s$ with similar \# of compute nodes on ResNet-18 trained over the CIFAR-10 dataset (regardless of attacks). } }
  \vspace{-3mm}
  \label{fig:solonspeedup}
\end{figure}

\begin{figure}[ht] 
\centering
\subfigure[Accuracy vs steps, rev-grad]{\includegraphics[width=0.4\textwidth]{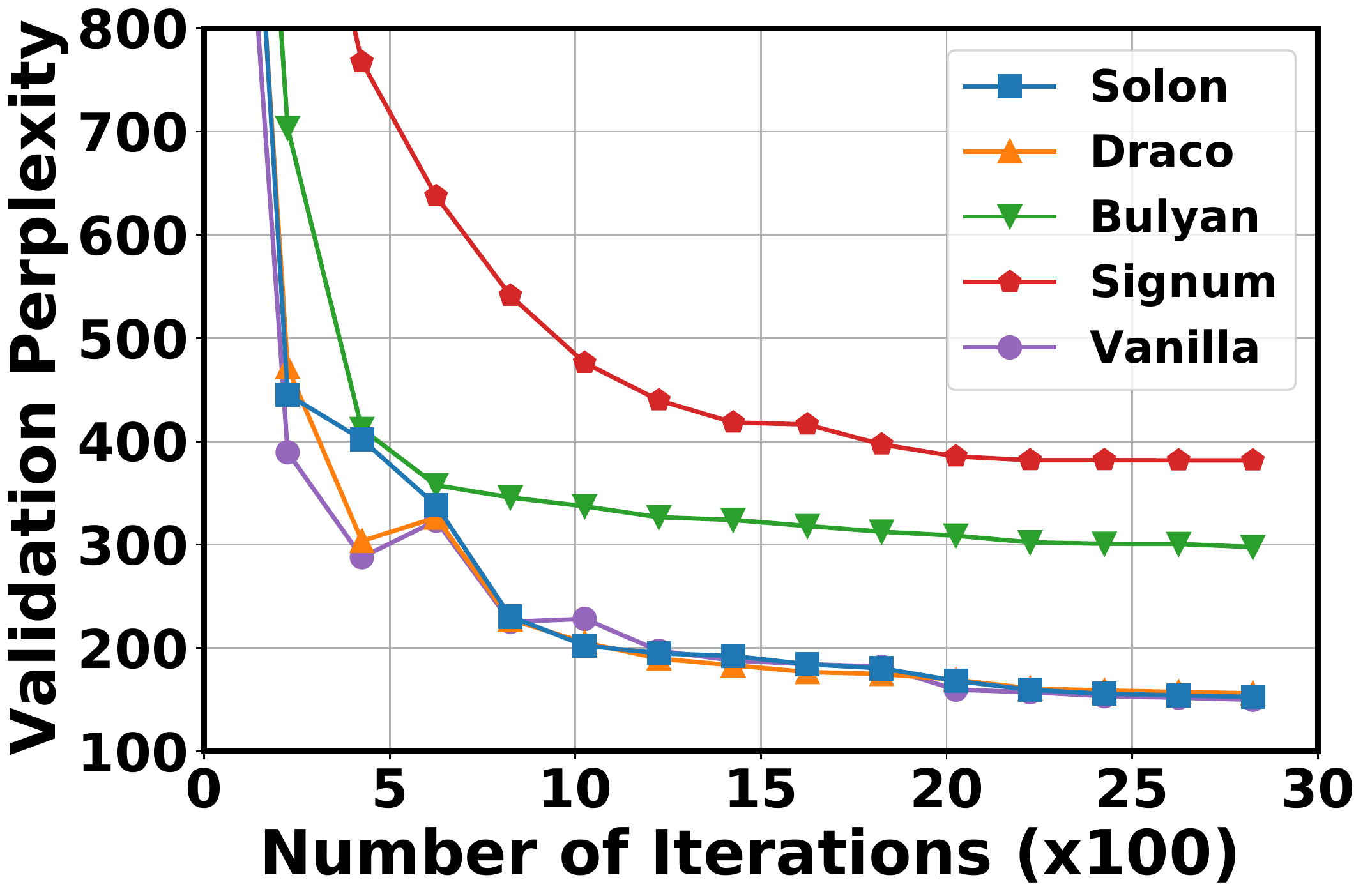}}
\hskip 20pt
\subfigure[Accuracy vs time, rev-grad]{\includegraphics[width=0.4\textwidth]{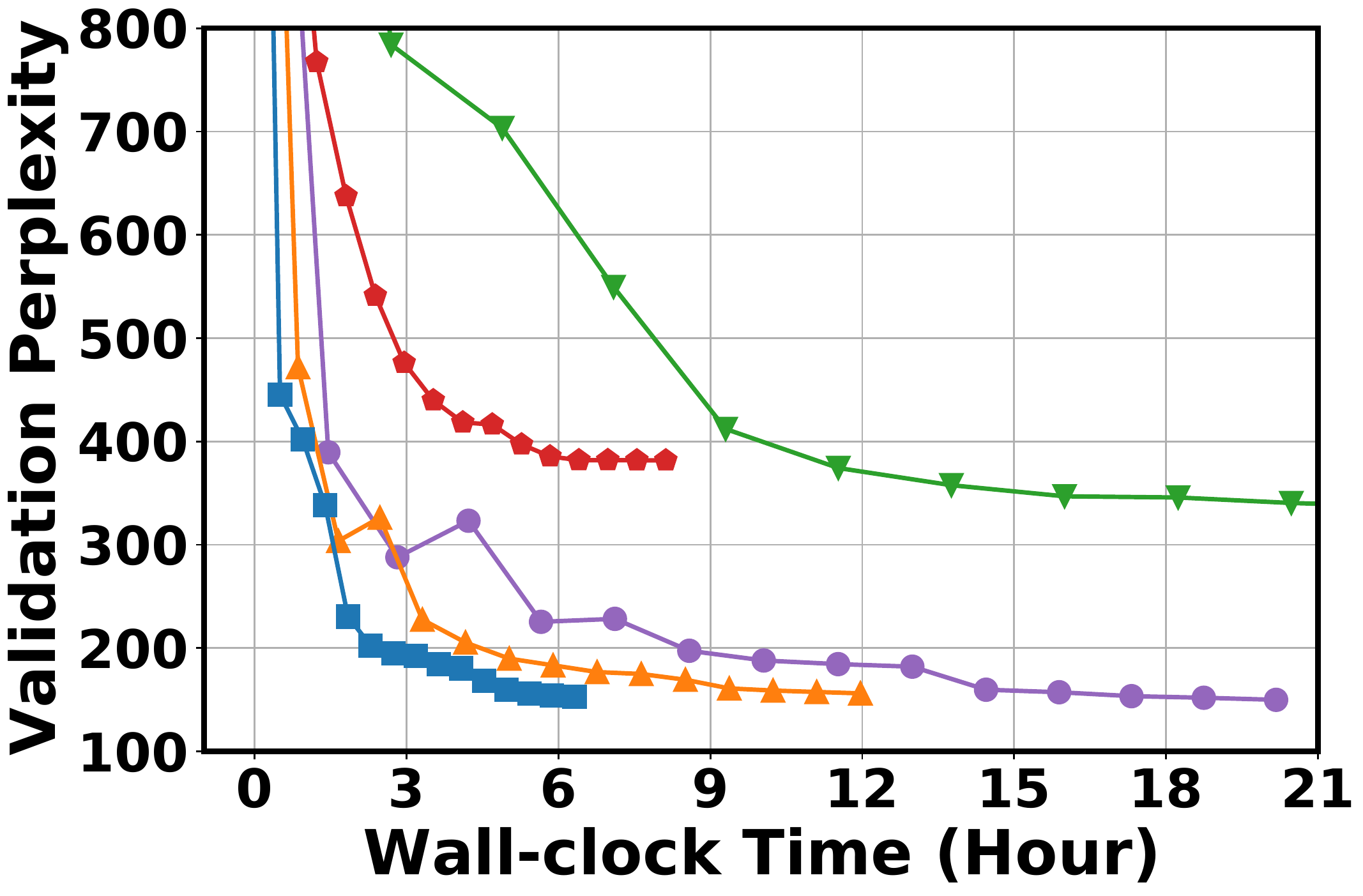}}
\vspace{-3mm}
\caption{\footnotesize{End to end convergence performance of \solon{} $r_c=10$ and other baselines on LSTM model trained over WikiText-2 dataset under reverse gradient attack, $s=5$. (a) Validation perplexity vs. \# iterations. (b) Validation perplexity vs. running time. }}
\label{fig:against-defense:appendix:LSTM}
\vspace{-3mm}
\end{figure}

Furthermore, \solon{} provides significant runtime speedups over existing methods.
For example, as shown in Figure \ref{fig:against-defense}(d), \solon{} converges faster than all the other Byzantine-resilient approaches. 
Its runtime performance even outperforms the vanilla SGD in a Byzantine-free environment.
This is primarily due to the communication efficiency of \solon{}. We observe a similar trend for the other two models (\eg LSTM under rev-grad attack shown in Figure \ref{fig:against-defense:appendix:LSTM}). 
Figure \ref{fig:solonspeedup}(a) gives a quantitative result of the speedups achieved by \solon{}.
To achieve a 90\% test accuracy, \solon{} obtains a speedup of 1.8$\times$ over \draco{} and 11$\times$ over \bulyan{} under the reverse gradient attack, while  \signum{} cannot reach 90\% accuracy. Other details and results can be found in the Appendix. 

\vspace{-2mm}
\paragraph{Per iteration cost.} 
Next, we dive into the per iteration cost of each approach.
As shown in Figure \ref{fig:timebreakdown}.
We note that \bulyan{} requires a significantly higher decoding time than all the other methods. 
This is probably because \bulyan{} deploys a computational expensive robust aggregator. 
Note that \solon{} reduces the communication cost by slightly increasing the computation and decoding complexity compared to \draco{}. Nevertheless, across all datasets and models considered in our experiments, \solon{} attains the fastest per iteration runtime.
This is because \solon{} largely reduces the communication cost, which is the bottleneck in a large cluster, and the extra computation and decoding cost is relatively small. 

\vspace{-2mm}
\paragraph{Effects of compression ratio.}
Finally we evaluate the effects of the compression ratio $r_c$ on \solon{}'s performance, as shown in Figure \ref{fig:solonspeedup}(b).
Here we vary compression rate $r_c$, fix attacks $s$, and change redundancy ratio $r$ accordingly. We keep the entire number of machines roughly at the same level by varying group numbers. 
Overall, as the compression ratio decreases, the communication cost increases almost linearly, and thus the total runtime.
This shows that \solon{} can be applied for different communication requirements with small extra overhead. 
In addition, we observe that the computation cost increases slightly when $r_c$ goes up, since the increase of $r_c$ will require an increase in the group batch size. The decode time also changes slightly. However, these are not major factors compared to communication. 
Other details can be found in the appendix.

In addition, we also evaluate the size of the gradients before and after compression, summarized in Table \ref{Tab:compressionsize}. 
Overall, \solon{} largely reduces the gradient size up to 10$\times$, depending on the specified compression ratio $r_c$.
In fact, \solon{} allows users to specify the compression ratio explicitly to satisfy different clusters' requirements.

\vspace{-0.1in}
\section{Conclusion}\label{Sec:SOLON:Conclusion}
\vspace{-0.1in}

In this paper, we propose \systemnameSolon{}, 
a distributed training framework that simultaneously resists Byzantine attack and reduces communication overhead via algorithmic redundancy. 
We show that there is a fundamental trade-off between Byzantine-resilience, communication cost, and computational cost.
Extensive experiments show that \solon{} provides significant speedups over existing methods, and consistently leads to successful convergence under different attacks. 
\newpage
{
\bibliography{MLService}
\bibliographystyle{plain}
}
\newpage
\appendix
\appendix

\paragraph{Outline} The supplement materials are organized as follows. All proofs are first presented in Section \ref{Sec:Solon:Proof}.
In addition, we provide a short discussion on how the Byzantine recovery problem is related to sparse recovery in Section \ref{sp}.
Section 
\ref{Sec:SOLON:Appendix:setup} and \ref{Sec:SOLON:Appendix:experiments} give the details of experimental setups and additional empirical findings, respectively.  Finally, we discuss the limitation and potential societal impact in more detail in Section \ref{Sec:SOLON:socialimpact}. 

\section{Proofs}\label{Sec:Solon:Proof}

\subsection{Proof of Theorem \ref{Thm:GenericBound} }
\begin{proof}
For $k=1,2,\ldots, d/r_c$, $j=\ldots, 1,0,-1,\ldots$, and $\ell = 1,2,\ldots, r_c$, define two functions $f_{1,r_c}(\cdot):\mathbb{R}^{d}\rightarrow \mathbb{R}^{d/r_c}$ and $f_{-1,r_c}(\cdot):\mathbb{R}^{d/r_c}\rightarrow \mathbb{R}^{d}$ as 
\begin{equation*}
\begin{split}
\lceil f_{1,r_c}(x_1,x_2,\ldots, x_d) _k/10^{r_cj+\ell}\rceil \mod 10 \equiv \lceil x_{(k-1)r_c+\ell}/10^j\rceil \mod 10
\end{split}
\end{equation*}
\begin{equation*}
\begin{split}
\lceil f_{-1,r_c}(y_1,y_2,\ldots, y_{d/r_c})_{r_c (k-1)+\ell }/10^{j}\rceil \mod 10 \equiv \lceil y_{k}/10^{r_c j +\ell } \rceil \mod 10.
\end{split}
\end{equation*}
The second subscript of the functions, $k$, and $r_c(k-1)+\ell $, respectively,  denotes the $k$th and $r_c(k-1)+\ell $th output of the functions.

The function $f_{1,r_c}(\cdot)$ compresses a $d$-dimensional vector by cascading all digits of each element of the input into a long vector, and $f_{-1,r_c}(\cdot)$  reverses the process.
For example, if $x_1 = 123, x_2 = 456, x_3 = 789$, and $d=r_c=3$, then $f_{1,r_c}(x_1,x_2,x_3) = 9638529630$.
Given $y_1 = 9638529630$ and $d=r_c = 3$, we have $f_{-1,r_c}(y_1)=(123,456,789).$
In general, we have the following lemma.
\begin{lemma}[The two functions are inverses.]
$f_{-1,r_c}(f_{1,r_c}(x_1,x_2,\ldots, x_d)) = (x_1, x_2,\ldots, x_d)$. 	
\end{lemma}
\begin{proof}
	Let $z_k = f_{1,r_c}(x_1,x_2,\ldots, x_d)_k$.
	We have 
	\begin{equation*}
	\begin{split}
	\lceil f_{-1,r_c}(z_1,z_2,\ldots, z_{d/r_c})_{r_c (k-1)+\ell }/10^{j}\rceil \mod 10 \equiv \lceil z_{k}/10^{r_c j +\ell } \rceil \mod 10
	\end{split}
	\end{equation*}
	and by definition of $z_k$, we have
\begin{equation*}
\begin{split}
\lceil z_k/10^{r_cj+\ell}\rceil \mod 10 \equiv \lceil x_{(k-1)r_c+\ell}/10^j\rceil \mod 10.
\end{split}
\end{equation*}
Thus, we have for all $j$
\begin{align*}
\lceil f_{-1,r_c}(z_1,z_2,\ldots, z_{d/r_c})_{r_c (k-1)+\ell }/10^{j}\rceil \mod 10 &\equiv \lceil z_{k}/10^{r_c j +\ell } \rceil \mod 10 
\\
&\equiv \lceil x_{(k-1)r_c+\ell}/10^j\rceil \mod 10.
\end{align*}
Therefore,  each digit of $f_{-1,r_c}(z_1,z_2,\ldots, z_{d/r_c})_{r_c (k-1)+\ell}$ is the same as that of $x_{r_c (k-1)+\ell}$. Hence, we must have $f_{-1,r_c}(z_1,z_2,\ldots, z_{d/r_c})_{r_c (k-1)+\ell} = x_{r_c (k-1)+\ell}$, which holds for $k=1,2,\ldots, d/r_c, \ell = 1,2,\ldots, r_c$.
Thus, we have  $f_{-1,r_c}(f_{1,r_c}(x_1,x_2,\ldots, x_d)) = (x_1, x_2,\ldots, x_d)$, which finishes the proof.
\end{proof}

Now we are ready to construct $(A',E',D')$. Given $(A,E,D)$, we let $A' \triangleq A$,  $E'_j(\mathbf{Y}_j) \triangleq f_{1,r_c}(E_j(\mathbf{Y}_j))$, and $D'_j(\mathbf{R}) \triangleq D(f_{-1,r_c}(\mathbf{R}_1), f_{-1,r_c}(\mathbf{R}_2), \ldots, f_{-1,r_c}(\mathbf{R}_P))$. 
By definition, it is clear that the constructed $(A',E',D')$ compresses the size of vectors (or also the communication cost) by a factor of $r_c$. 
The remaining part is to prove that $(A', E', D')$ can resist $s$ adversarial nodes.
W.l.o.g., assume the first $P-2s$ nodes are not Byzantine. Then we have
$\mathbf{R}_j = f_{1,r_c}(E(\mathbf{Y}_j))$, and thus $f_{-1,r_c}(\mathbf{R}_j)  = f_{-1,r_c}(f_{1,r_c}(E(\mathbf{Y}_j)))  = E_j(\mathbf{Y}_j), j= 1,2,\ldots, P-2s$.
Hence, we have $D'(\mathbf{R}) = D(E_1(\mathbf{Y}_1), E_2(\mathbf{Y}_2), \ldots, E_{P-2s}(\mathbf{Y}_{P-2s}), \mathbf{R}_{P-2s+1},\ldots, \mathbf{R}_P)$. Since $(A,E,D)$ can resist $s$ Byzantine nodes, given $P-2s$ correctly received parts $E_j(\mathbf{Y}_j), j=1,2,\ldots, P-2s$ and $2s$ arbitrary parts, the decoder should return the correct gradient sum. In other words, we have  $D(E_1(\mathbf{Y}_1), E_2(\mathbf{Y}_2), \ldots, E_{P-2s}(\mathbf{Y}_{P-2s}), \mathbf{R}_{P-2s+1},\ldots, \mathbf{R}_P) = \sum_{i=1}^{P}\mathbf{g}_i$ for any $\mathbf{R}_{P-2s+1}, \mathbf{R}_{P-2s+2},\ldots, \mathbf{R}_P$. Thus, $D'(\mathbf{R}) = \sum_{i=1}^{P}\mathbf{g}_i$, which demonstrates that $(A',E',D')$ can resist $s$ Byzantine nodes.
\end{proof}

\subsection{Proof of Theorem \ref{Thm:RedundancyRatioBound} }
\begin{proof}
	
	We define a valid $s$-attack first.
	\begin{definition} The matrix
		$\mathbf{N} =[\mathbf{n}_1,\mathbf{n}_2,\ldots, \mathbf{n}_P]$ is a valid $s$-attack if and only if $\left\vert{\{j: \|\mathbf{n}_{j}\|_0\not= 0 \}}\right\vert \leq s$. 
	\end{definition}
	Suppose $(\mathbf{A}, E, D)$ can resist $s$ adversaries.
	The goal is to prove 
	$\|\bA\|_0 \geq P(2s+r_c)$.
	In fact we can prove a slightly stronger claim: $\|\mathbf{A}_{\cdot,i}\|_0 \geq  \left(2s+r_c\right), i = 1,2,\ldots, B$.
	Suppose for some $i$,  $\|\mathbf{A}_{\cdot,i}\|_0 = \tau <  \left(2s+r_c\right)$.
	Without loss of generality, assume that $\mathbf{A}_{1, i}, \mathbf{A}_{2, i},\ldots, \mathbf{A}_{\tau,i} $ are nonzero. Let $\mathbf{G}_{-i} = [\mathbf{g}_1,\mathbf{g}_2,\ldots,\mathbf{g}_{i-1},\mathbf{g}_{i+1},\ldots, \mathbf{g}_P]$. 
	Since $(\mathbf{A},E,D)$ can protect against $s$ adversaries,  we have for any $\mathbf{G}$, 	\begin{align*}
	D(\mathbf{Z}^{\mathbf{A},E,\mathbf{G}}+\mathbf{N} ) = \mathbf{G} \mathbf{1}_P=  \mathbf{G}_{-i} \mathbf{1}_{P-1} + \mathbf{g}_i,
	\end{align*}
	for any valid $s$-attack $\mathbf{N}$.
	Our goal is to show a contradiction based on the above equation.
	
	Recall that for regular encoders, $\smash{E_{j,v}(\mathbf{Y}_j) = \hat{E}_{j,v}\left(\mathbf{1}_d^T\left(\mathbf{U}_{j,v} \odot \mathbf{Y}_{j}\right)\right)}$. Let $\mathbf{u}_{j,v}$ be the $i$th column of $\mathbf{U}_{j,v}$ and
	$\hat{\mathbf{U}} \triangleq [\mathbf{u}_{1,1}, \mathbf{u}_{1,2}, \ldots, \mathbf{u}_{1,d_c}, \mathbf{u}_{2,1}, \ldots, \mathbf{u}_{2,d_c},\ldots, \mathbf{u}_{\tau-2s,d_c}]$.\footnote{Since $i$ is fixed, we omit it from $\mathbf{u}_{j,v}$, $\hat{\mathbf{U}}$, and other quantities depending implicitly on it.}
	Note that 
	$$\tau<2s+r_c \Leftrightarrow	 \tau-2s <r_c \Leftrightarrow	 (\tau-2s) d_c < r_c d_c = d.$$
	Since $\hat{\mathbf{U}}$ is a $d \times (\tau-2s) d_c$ matrix, $\hat{\mathbf{U}}$ is not of full row rank.
	Therefore, there exists a $d$-dimensional vector $\hat{\mathbf{x}}\not=\mathbf{0}$ such that $\hat{\mathbf{U}}^T \hat{\mathbf{x}} =0$. 
	
	Let $\mathbf{g}^1_i = \mathbf{1}_{d}$,  $\mathbf{g}^2_i = \mathbf{1}_{d}+\hat{\mathbf{x}}$, and for $a=1,2$
	$$\mathbf{G}^a = [\mathbf{g}_1,\mathbf{g}_2,\ldots,\mathbf{g}_{i-1},\mathbf{g}_{i}^a,\mathbf{g}_{i+1},\ldots, \mathbf{g}_P].$$
	Then for any valid $s$-attacks $\mathbf{N}^1, \mathbf{N}^2$,
	\begin{align*}
	D(\mathbf{Z}^{\mathbf{A},E,\mathbf{G}^1}+\mathbf{N}^1 ) =  \mathbf{G}_{-i} \mathbf{1}_{P-1} + \mathbf{1}_d.
	\end{align*}	
	and
	\begin{align*}
	D(\mathbf{Z}^{\mathbf{A},E,\mathbf{G}^2}+\mathbf{N}^2 ) =  \mathbf{G}_{-i} \mathbf{1}_{P-1} + \mathbf{1}_d + \hat{\mathbf{x}}.
	\end{align*}	
	Now we find $\mathbf{N}^1, \mathbf{N}^2$ such that $D(\mathbf{Z}^{\mathbf{A},E,\mathbf{G}^1}+\mathbf{N}^1 ) = D(\mathbf{Z}^{\mathbf{A},E,\mathbf{G}^2}+\mathbf{N}^2 )$ which then  leads to a contradiction.
	Construct $\mathbf{N}^1$ and $\mathbf{N}^2$ by
	\begin{equation*}
	\mathbf{N}^1_{\ell,j}= 
	\begin{cases}
	\left[\mathbf{Z}^{\mathbf{A},E, \mathbf{G}^2}\right]_{\ell,j} - 	\left[\mathbf{Z}^{\mathbf{A},E, \mathbf{G}^1}\right]_{\ell,j} ,&j=\tau -2s+1, \tau-2s+2,\ldots, \tau-s\\
	0,              & \text{otherwise}
	\end{cases}
	\end{equation*}
	and
	\begin{equation*}
	\mathbf{N}^2_{\ell,j}= 
	\begin{cases}
	\left[\mathbf{Z}^{\mathbf{A},E, \mathbf{G}^1}\right]_{\ell,j} - 	\left[\mathbf{Z}^{\mathbf{A},E, \mathbf{G}^2}\right]_{\ell,j}, & j= \tau-s+1, \tau-s+2, \ldots, \tau\\
	0,              & \text{otherwise.}
	\end{cases}
	\end{equation*}
	One can readily verify that $\mathbf{N}^1, \mathbf{N}^2$ are both valid $s$-attacks. In addition,
	\begin{align*}
	\left[\mathbf{Z}^{\mathbf{A},E,\mathbf{G}^1}\right]_{\ell,j}+\mathbf{N}^1_{{\ell,j}} = \left[\mathbf{Z}^{\mathbf{A},E,\mathbf{G}^2}\right]_{\ell,j}+\mathbf{N}^2_{{\ell,j}},\,\,\, j=\tau-2s+1,\tau-2s+2,\ldots, \tau.
	\end{align*}	 
	Since $\mathbf{A}_{j,i}=0$ for all $j>\tau$, the encoder functions of compute nodes with index $j>\tau$  do not depend on the $i$th gradient. Since $\mathbf{G}^1$ and $\mathbf{G}^2$ only differ in the $i$th gradient, the encoder function of any compute node with index $j>\tau$ has the same output.
	Thus, we have  
	\begin{align*}
	\left[\mathbf{Z}^{\mathbf{A},E,\mathbf{G}^1}\right]_{\ell,j}+\mathbf{N}^1_{{\ell,j}} = \left[\mathbf{Z}^{\mathbf{A},E,\mathbf{G}^1}\right]_{\ell,j} = \left[\mathbf{Z}^{\mathbf{A},E,\mathbf{G}^2}\right]_{\ell,j} = \left[\mathbf{Z}^{\mathbf{A},E,\mathbf{G}^2}\right]_{\ell,j}+\mathbf{N}^2_{{\ell,j}}, \qquad j> \tau.
	\end{align*}
	Now consider $1\leq j\leq \tau-2s$. 
	By construction of $\hat{\mathbf{x}}$, we have $\hat{\mathbf{U}}^T \hat{\mathbf{x}} = \mathbf{0}$.
	That is, for $1\leq j \leq \tau-2s, 1\leq u\leq d_c$, $\mathbf{u}_{j,v}^T \hat{\mathbf{x}} = 0$, which implies 
	$$\mathbf{u}_{j,v}^T \mathbf{g}_i^1 = \mathbf{u}_{j,v}^T \mathbf{g}_i^1 + \mathbf{u}_{j,v}^T \hat{\mathbf{x}}  = \mathbf{u}_{j,v}^T \mathbf{g}_i^2.$$
	Let $\mathbf{Y}_j^1$ and $\mathbf{Y}_j^2$ be the gradients computed at the $j$th node for $\mathbf{G}^1$ and $\mathbf{G}^2$, respectively.
	In other words, $\mathbf{Y}_j^1 = \left( \mathbf{1}_d \mathbf{A}_{j,\cdot} \right)\odot \mathbf{G}^1$ and $\mathbf{Y}_j^2 = \left( \mathbf{1}_d \mathbf{A}_{j,\cdot} \right)\odot \mathbf{G}^2$.
	Since $\mathbf{G}^1$ and $\mathbf{G}^2$ only differ in the $i$th column,  $\mathbf{Y}_j^1$ and $\mathbf{Y}_j^2$ can also only differ in the $i$th column.
	Therefore, for $1\leq j\leq \tau-2s, 1\leq v\leq d_c$, the vectors $\mathbf{u}_{j,v}^T \mathbf{Y}_j^1$ and $\mathbf{u}_{j,v}^T \mathbf{Y}_j^2$ can only differ in the $i$th column (in this case, the $i$th entry).
	But the $i$th column of $\mathbf{u}_{j,v}^T \mathbf{Y}_j^1$ is $\mathbf{u}_{j,v}^T \mathbf{g}_i^1$, the $i$th column of $\mathbf{u}_{j,v}^T \mathbf{Y}_j^2$ is $\mathbf{u}_{j,v}^T \mathbf{g}_i^2$, and we have just shown that  $\mathbf{u}_{j,v}^T \mathbf{g}_i^1 = \mathbf{u}_{j,v}^T \mathbf{g}_i^2$.
	Thus, we must have $\mathbf{u}_{j,v}^T \mathbf{Y}_j^1 =  \mathbf{u}_{j,v}^T \mathbf{Y}_j^2$ for $1\leq j\leq \tau, 1\leq v\leq d_c$.
	This implies 
	$$\mathbf{1}_P^T(\mathbf{U}_{j,v}\odot \mathbf{Y}_{j}^1) = \mathbf{1}_P^T(\mathbf{U}_{j,v}\odot \mathbf{Y}_{j}^2),$$
	which in turn implies 
	$$E_{j,v}(\mathbf{Y}_j^1) = \hat{E}_{j,v}(\mathbf{1}_P^T(\mathbf{U}_{j,v}\odot \mathbf{Y}_{j}^1)) = \hat{E}_{j,v}(\mathbf{1}_P^T(\mathbf{U}_{j,v}\odot \mathbf{Y}_{j}^1)) = E_{j,v}(\mathbf{Y}_j^2).$$
	That is, $\left[\mathbf{Z}^{\mathbf{A},E,\mathbf{G}^1}\right]_{v,j} = \left[\mathbf{Z}^{\mathbf{A},E,\mathbf{G}^2}\right]_{v,j}$.
	Further noticing that $\mathbf{N}^1_{v,j} = \mathbf{N}^2_{v,j} = 0$ when $1\leq j\leq \tau-2s$, we have
	\begin{align*}
	\left[\mathbf{Z}^{\mathbf{A},E,\mathbf{G}^1}\right]_{\ell,j}+\mathbf{N}^1_{{\ell,j}} =  \left[\mathbf{Z}^{\mathbf{A},E,\mathbf{G}^2}\right]_{\ell,j}+\mathbf{N}^2_{{\ell,j}}, 1\leq j\leq \tau -2s.
	\end{align*}
	Hence, we have 		
	\begin{align*}
	\left[\mathbf{Z}^{\mathbf{A},E,\mathbf{G}^1}\right]_{\ell,j}+\mathbf{N}^1_{{\ell,j}} =  \left[\mathbf{Z}^{\mathbf{A},E,\mathbf{G}^2}\right]_{\ell,j}+\mathbf{N}^2_{{\ell,j}}, \qquad\forall j,
	\end{align*}
	which shows
	\begin{align*}
	\mathbf{Z}^{\mathbf{A},E,\mathbf{G}^1}+\mathbf{N}^1 = \mathbf{Z}^{\mathbf{A},E,\mathbf{G}^2}+\mathbf{N}^2.
	\end{align*}
	Therefore, we have 
	\begin{align*}
	D(\mathbf{Z}^{\mathbf{A},E,\mathbf{G}^1}+\mathbf{N}^1 ) = D(\mathbf{Z}^{\mathbf{A},E,\mathbf{G}^2}+\mathbf{N}^2 )
	\end{align*}
	and thus
	\begin{align*}
	\mathbf{G}_{-1} \mathbf{1}_{P-1} + \mathbf{1}_d = D(\mathbf{Z}^{\mathbf{A},E,\mathbf{G}^1}+\mathbf{N}^1 ) = D(\mathbf{Z}^{\mathbf{A},E,\mathbf{G}^2}+\mathbf{N}^2 ) = \mathbf{G}_{-1} \mathbf{1}_{P-1} + \mathbf{1}_d + \hat{\mathbf{x}}. 
	\end{align*}
	This gives us a contradiction. Hence, the assumption is not correct and we must have $\|\mathbf{A}_{\cdot,i}\|_0 \geq  \left(2s+r_c\right)$, for $i = 1,2,\ldots, P$.	
	Thus, we must have $\|A\|_0 \geq (2s+r_c)P$. 
	
	A direct but important corollary of this theorem is a bound on the number of adversaries \solon{} can resist.
	
	\begin{corollary}\label{Cor:AdversarialBound}
		$(\mathbf{A}, E, D)$ can resist at most $\frac{P-r_c}{2}$ adversarial nodes.
	\end{corollary}
	\begin{proof}
		According to Theorem \ref{Thm:RedundancyRatioBound}, the redundancy ratio is at least $2s+r_c$, meaning that every data point must be replicated at least $2s+r_c$ times. 
		Since there are  $P$ compute nodes in total, we must have $2s+r_c\leq P$, which implies $s\leq \frac{P-r_c}{2}$. 
		Thus, $(\mathbf{A},E,D)$ can resist at most $\frac{P-r_c}{2}$ adversaries.
	\end{proof}
	This corollary implies that the communication compression ratio cannot exceed the total number of compute nodes.
\end{proof}

\subsection{Proof of Lemma \ref{Thm:LBCDetection}}
\begin{proof}
    We will omit the superscript $LBC$ indicating the linear block code in this and the following proofs.
	We need a few lemmas first.
	\begin{lemma}\label{Lemma:CycProb}
		Define the $P$-dimensional vector $\mathbf{\gamma} \triangleq \left[\gamma_1, \gamma_2,\ldots, \gamma_P \right]^T = \left(\mathbf{f} \mathbf{N}\right)^T$. Then we have 
		\begin{align*}
		\textit{Pr}( \{ j : \gamma_j \not= 0 \} = \{j: \| \{\mathbf{N}_{\cdot,j} \|_0 \not= 0 \}) = 1.
		\end{align*}
	\end{lemma}
	\begin{proof}
		Let us prove that 
		\begin{align*}
		\textit{Pr}( \mathbf{N}_{\cdot,j} \not= 0  | \gamma_j \not = 0 ) = 1.
		\end{align*}
		and 
		\begin{align*}
		\textit{Pr}(\gamma_j \not = 0 | \mathbf{N}_{\cdot,j} \not= 0 ) = 1.
		\end{align*}
		for any $j$.	
		Combining those two equations we prove the lemma.
		
		The first equation is readily verified, because $\mathbf{N}_{\cdot,j} = 0$ implies $\gamma_j = \mathbf{f} \mathbf{N}_{\cdot,j} = 0$.
		For the second one, note that $\mathbf{f}$ has entries drawn independently from the standard normal distribution. Therefore we have that $\gamma_j = \mathbf{f} \mathbf{N}_{\cdot,j} \sim \mathcal{N}(\mathbf{1}^T \mathbf{N}_{\cdot,j}, \|\mathbf{N}_{\cdot,j}\|_2^2 )$.
		Since $\gamma_j$ is a random variable with a non-degenerate normal distribution when $\|\mathbf{N}_{\cdot,j}\|_2^2\neq 0$, which has an absolutely continuous density with respect to the Lebesgue measure, the probability of taking any particular value is $0$.
		In particular, 
		\begin{align*}
		\textit{Pr}( \gamma_j = 0 | \mathbf{N}_{\cdot,j} \not= 0) = 0,
		\end{align*}	
		and thus 
		$\textit{Pr}( \gamma_j \not= 0 | \mathbf{N}_{\cdot,j} \not= 0 ) = 1$.
		This proves the second equation and finishes the proof.
	\end{proof}

	\begin{lemma}\label{l9}
		The function $P_j(w) \triangleq (\sum_{i=0}^{r_c+s-1} a_{i+1} w^i)/(w^s+\sum_{i=0}^{s-1} a_{i+r_c+s+1} w^i) $ is a well-defined polynomial. In fact, we have $P_j(w) = \sum_{k=0}^{r_c-1} u_{k+1} w^k $, where $u_k  \triangleq \sum_{\ell=1}^{d_c} \mathbf{f}_\ell [\mathbf{Y}_{rj} \mathbf{1}_P]_{k+(\ell-1)r_c}$.
	\end{lemma}
	\begin{proof}
		Let $\bar{\mathbf{y}}_j \triangleq \mathbf{Y}_{rj} \mathbf{1}_P$. 
		Let  $i_1, i_2,\ldots, i_s$ denote be the indices of adversarial/Byzantine nodes in the $j$th group.
		Construct polynomials 
		$$P(w)  = \prod_{k=1}^{s}(w-w_{i_k})\triangleq  w^s+\sum_{k=0}^{s-1} \theta_k w^k $$
		and for $u_k$  defined in the statement of the lemma,
		$$Q(w) = \sum_{k=0}^{r_c-1} u_{k+1} w^k.$$ 
		Let 
		$$R(w)  = P(w) Q(w)  \triangleq \sum_{k=0}^{r_c+s-1} \beta_k w^k.$$
		Recall that $\mathbf{r}_{j,c} = \mathbf{f} \mathbf{R}_j$.
		When  $(j-1)r+k = i_\ell$ for some $\ell$, by definition, we have $P(w_{(j-1)r+k}) = 0$ and thus 
		$$		P(w_{(j-1)r+k}) \left[\mathbf{r}_{j,c}\right]_k =0= P(w_{(j-1)r+k}) Q(w_{(j-1)r+k}).$$
		When  $(j-1)r+k \not= i_\ell$ for any $\ell$, by definition, the $(j-1)r+k$th node, or the $k$th node in the $j$th group, is not adversarial/Byzantine. By definition, 
		\begin{equation*}
		\begin{split}
		[\mathbf{r}_{j,c}]_k 
		=& [\mathbf{f} \mathbf{R}_j]_k = \mathbf{f} [\mathbf{R}_j]_{\cdot,k}  = \mathbf{f} \mathbf{z}_{(j-1)r+k}\\
		=& \mathbf{f} \mathbf{W}_{(j-1)r+k} \mathbf{Y}_{(j-1)r+k}\mathbf{1}_P = \mathbf{f} \mathbf{W}_{(j-1)r+k} \mathbf{Y}_{rj}\mathbf{1}_P.
					\end{split}
		\end{equation*}
	This can be written as
	    \begin{equation*}
		\begin{split}
		 \mathbf{f} \mathbf{W}_{(j-1)r+k} \mathbf{\bar{y}}_j
		=& \mathbf{f} \mathbf{I}_{d_c} \otimes [1, w_{(j-1)r+k}, w_{(j-1)r+k}^2, \ldots, w_{(j-1)r+k}^{r_c-1}] \mathbf{\bar{y}}_j.
			\end{split}
		\end{equation*}
	This in turn equals	
	    \begin{equation*}
		\begin{split}
		  \sum_{p=1}^{d_c} \mathbf{f}_p \sum_{\ell = 0}^{r_c-1} w_{(j-1)r+k}^\ell [\mathbf{\bar{y}}_j]_{\ell+1+(p-1)r_c} 
		= &  \sum_{\ell = 0}^{r_c-1}  \sum_{p=1}^{d_c} \mathbf{f}_p [\mathbf{\bar{y}}_j]_{\ell+1+(p-1)r_c}  w_{(j-1)r+k}^\ell  \\
		= & \sum_{\ell = 0}^{r_c-1} u_{\ell+1}  w_{(j-1)r+k}^\ell = Q(w_{(j-1)r+k}).
		\end{split}
		\end{equation*}
		That is, $ \left[\mathbf{r}_{j,c}\right]_k =  Q(w_{(j-1)r+k})$, which implies 
		$$P(w_{(j-1)r+k}) \left[\mathbf{r}_{j,c}\right]_k = P(w_{(j-1)r+k}) Q(w_{(j-1)r+k}).$$
		Noting that $P(w) Q(w) = R(w)$,  we have
		\begin{equation}\label{PQR}
		P(w_{(j-1)r+k}) \left[\mathbf{r}_{j,c}\right]_k  = R(w_{(j-1)r+k})
		\end{equation}
		for any $k$.
		
		This is a linear system for the vector $\mathbf{b} \triangleq [\beta_0, \beta_1,\ldots, \beta_{r_c+s-1}, \theta_0,\theta_1,\ldots, \theta_{s-1} ]^T$. 
		Rewriting the system in a compact way, we have 
		$$\begin{bmatrix}
		\hat{\mathbf{W}}_{j,r_c+s-1},  & -\hat{\mathbf{W}}_{j,s-1} \odot \left( \mathbf{y} \mathbf{1}_{s}^T\right)
		\end{bmatrix}
		\mathbf{b}
		= \mathbf{r}_{j,c} \odot \left[\hat{\mathbf{W}}_{j,s}\right]_{\cdot,s}.$$
		There are now two cases to consider.
		
		(i) Solving the linear system leads to $\mathbf{a}= \mathbf{b}$.
		Then by construction of $P_j(w)$, we have
		\begin{equation*}
		\begin{split}
		P_j(w) = & (\sum_{i=0}^{r_c+s-1} a_{i+1} w^i)/(w^s+\sum_{i=0}^{s-1} a_{i+r_c+s+1} w^i)  
		= R(w) / P(w) = Q(w) = \sum_{k=0}^{r_c-1} u_{k+1} w^k.
		\end{split}
		\end{equation*}
		The second equality follows by plugging in the solution $\mathbf{a}= \mathbf{b}$ into the definition of both the numerator and the denominator of $P_j$. Then we see that the numerator reduces to $\sum_{k=0}^{r_c+s-1} \beta_k w^k = R(w)$, while the denominator reduces to $ w^s+\sum_{k=0}^{s-1} \theta_k w^k  = P(w)$.
		The third equality follows due to the definition of $R = PQ$, while the last equality holds due to the definition of $Q$.
		
		(ii) Solving the linear system gives another solution 
		$\tilde{\mathbf{b}} =  [\tilde\beta_0, \tilde\beta_1,\ldots, \tilde\beta_{r_c+s-1}, \tilde\theta_0,\tilde\theta_1,\ldots, \tilde\theta_{s-1}]^T.$ 	Construct polynomials $\tilde P(w) = w^s+\sum_{k=0}^{s-1} \tilde\theta_k w^k $  and  
		$\tilde R(w)  = \sum_{k=0}^{r_c+s-1} \beta_k w^k $. Since $\tilde{\mathbf{b}}$ is a solution to the original linear system, we must have
		\begin{equation}\label{PQR2}
		\tilde P(w_{(r-1)j+k}) \left[\mathbf{r}_{j,c}\right]_k  = \tilde R(w_{(r-1)j+k}).
		\end{equation}
		We can combine this with \eqref{PQR}.
		If $P(w_{(r-1)j+k})$ and $\tilde P(w_{(r-1)j+k})$ are not zero, then we have 
		\begin{equation*}
		\begin{split}
		& \tilde R(w_{(r-1)j+k}) / \tilde P(w_{(r-1)j+k})
		=  \left[\mathbf{r}_{j,c}\right]_k = R(w_{(r-1)j+k}) / P(w_{(r-1)j+k}) 
		\end{split}
		\end{equation*}
		which implies 
		\begin{equation}\label{pqr3}
		\begin{split}
		& \tilde R(w_{(r-1)j+k})  P(w_{(r-1)j+k})
		= R(w_{(r-1)j+k})  \tilde P(w_{(r-1)j+k}).
		\end{split}
		\end{equation}
		If $P(w_{(r-1)j+k}) = 0$, then by \eqref{PQR}, $R(w_{(r-1)j+k}) = 0$, and hence \eqref{pqr3} still holds.
		If $\tilde P(w_{(r-1)j+k}) = 0$, then similarly by \eqref{PQR2}, $\tilde R(w_{(r-1)j+k}) = 0$, and hence \eqref{pqr3} also holds. Therefore, \eqref{pqr3} holds 
		for any $k$. For fixed $j$, the index $k$ can take in total $r=r_c+2s$ values, and the degrees of both $P(w)\tilde R(w)$ and $\tilde P(w) R(w)$ are $r_c+2s-1$.
		Thus, we must have for all $w$
		\begin{equation*}
		\begin{split}
		& \tilde R(w)  P(w)
		=R(w)  \tilde P(w)  \end{split}
		\end{equation*}
		and hence
		\begin{equation*}
		\begin{split} 
		P_j(w) =& \tilde R(w) / \tilde P(w)
		= R(w)/P(w) = Q(w) = \sum_{k=0}^{r_c-1} u_{k+1} w^k.
		\end{split}
		\end{equation*}
		Thus $P_j(w) = \sum_{k=0}^{r_c-1} u_{k+1} w^k$ is a well-defined polynomial, finishing the proof.
	\end{proof}
	Now we are ready to prove the lemma, by a calculation very similar to the one we have done earlier in the proof of Lemma \ref{l9}.
	By definition, 
	\begin{equation*}
	\begin{split}
	[\mathbf{r}_{j,c}]_k 
	=& [\mathbf{f} \mathbf{R}_j]_k = \mathbf{f} [\mathbf{R}_j]_{\cdot,k}  = \mathbf{f}\left( \mathbf{z}_{(j-1)r+k}+\mathbf{n}_j\right)\\
	=& \mathbf{f} \mathbf{W}_{(j-1)r+k} \mathbf{Y}_{(j-1)r+k}\mathbf{1}_P +\mathbf{f} \mathbf{n}_j
	=  \mathbf{f} \mathbf{W}_{(j-1)r+k} \mathbf{Y}_{rj}\mathbf{1}_P +\mathbf{f} \mathbf{n}_j.
	\end{split}
	\end{equation*}
	This can be written as
	\begin{equation*}
		\begin{split}  \mathbf{f} \mathbf{W}_{(j-1)r+k} \mathbf{\bar{y}}_j+\mathbf{f} \mathbf{n}_j
	=& \mathbf{f} \mathbf{I}_{d_c} \otimes [1, w_{(j-1)r+k}, w_{(j-1)r+k}^2, \ldots, w_{(j-1)r+k}^{r_c-1}] \mathbf{\bar{y}}_j +\mathbf{f} \mathbf{n}_j.
	\end{split}
	\end{equation*}
	This equals	
	\begin{equation*}
		\begin{split}
	 \sum_{p=1}^{d_c} \mathbf{f}_p \sum_{\ell = 0}^{r_c-1} w_{(j-1)r+k}^\ell [\mathbf{\bar{y}}_j]_{\ell+1+(p-1)r_c} +\mathbf{f} \mathbf{n}_j
	= &  \sum_{\ell = 0}^{r_c-1}  \sum_{p=1}^{d_c} \mathbf{f}_p [\mathbf{\bar{y}}_j]_{\ell+1+(p-1)r_c}  w_{(j-1)r+k}^\ell +\mathbf{f} \mathbf{n}_j \\
	= & \sum_{\ell = 0}^{r_c-1} u_{\ell+1}  w_{(j-1)r+k}^\ell = P_j(w_{(j-1)r+k})+\mathbf{f} \mathbf{n}_j.
	\end{split}
	\end{equation*}
	Thus, $[\mathbf{r}_{j,c}]_k = P_j(w_{(j-1)r+k})+\mathbf{f} \mathbf{n}_j$. By Lemma \ref{Lemma:CycProb}, with probability equal to unity, $\mathbf{f} \mathbf{n}_j = \gamma_j = 0$ if and only if $\|\mathbf{n}_j\| = 0$.
	In other words, with probability equal to unity, $[\mathbf{r}_{j,c}]_k \not = P_j(w_{(j-1)r+k})$ if and only if $\|\mathbf{n}_j\| \not= 0$, which demonstrates the correctness of Lemma \ref{Thm:LBCDetection}.
	
\end{proof}

\subsection{Proof of Lemma \ref{thm:LBClocalcorrect}}
\begin{proof}
	The $j$th node needs to to compute and send to the PS
	\begin{equation*}
	\mathbf{z}_j = \mathbf{W}_j \mathbf{Y}_j\mathbf{1}_P.
	\end{equation*}
	
	Due to the assignment matrix $\mathbf{A}$, all nodes in group $j$ compute the same $\mathbf{Y}_j$, so $\mathbf{Y}_{(j-1)r+k} = \mathbf{Y}_{jr},$ for $k\in \{1,2,\ldots, r\}$.
	Let us partition $\mathbf{Y}_{jr} \mathbf{1}_P$ into $d/r$ vectors of size  $r\times 1$, i.e., 
	$$\mathbf{Y}_{jr}\mathbf{1}_P \triangleq \left[\begin{matrix}
	\mathbf{Y}_{jr,1}\\
	\mathbf{Y}_{jr,2}\\
	\ldots\\
	\mathbf{Y}_{jr,d/r}
	\end{matrix}
	\right].$$
	Then we have
	\begin{equation*}
	\begin{split}
	\mathbf{r}_{(j-1)r+k} 
	=& \mathbf{W}_{(j-1)r+k} \mathbf{Y}_{jr}\mathbf{1}_P 
	= \mathbf{I}_{d_c} \otimes [1, w_{(j-1)r+1}, w_{(j-1)r+2}^2, \ldots, w_{jr}^{r_c-1}] \left[\begin{matrix}
	\mathbf{Y}_{jr,1}\\
	\mathbf{Y}_{jr,2}\\
	\ldots\\
	\mathbf{Y}_{jr,d/r}
	\end{matrix}
	\right]\\
	=&\left[\begin{matrix}
	[1, w_{(j-1)r+1}, w_{(j-1)r+2}^2, \ldots, w_{jr}^{r_c-1}] \mathbf{Y}_{jr,1}\\
	[1, w_{(j-1)r+1}, w_{(j-1)r+2}^2, \ldots, w_{jr}^{r_c-1}] \mathbf{Y}_{jr,2}\\
	\ldots\\
	[1, w_{(j-1)r+1}, w_{(j-1)r+2}^2, \ldots, w_{jr}^{r_c-1}]	\mathbf{Y}_{jr,d/r}
	\end{matrix}
	\right].  
	\end{split}
	\end{equation*} 
	Therefore,
	\begin{equation*}
	\begin{split}
	&	[\mathbf{z}_{(j-1)r+1},\mathbf{z}_{(j-1)r+2},\ldots, \mathbf{z}_{jr}]
	=
	\left[\begin{matrix}
	\left[\hat{\mathbf{W}}_{j,r_c-1} \mathbf{Y}_{jr,1}
	\right]^T \\
	\left[\hat{\mathbf{W}}_{j,r_c-1} \mathbf{Y}_{jr,2}\right]^T\\
	\ldots\\
	\left[\hat{\mathbf{W}}_{j,r_c-1} \mathbf{Y}_{jr,d/r}
	\right]^T
	\end{matrix}\right].
	\end{split}
	\end{equation*}
	By definiton, we have \begin{equation*}
	\begin{split}
	\mathbf{R}_j & = [\mathbf{z}_{(j-1)r+1},\mathbf{z}_{(j-1)r+2},\ldots, \mathbf{z}_{jr}]
	+ [\mathbf{n}_{(j-1)r+1},\mathbf{n}_{(j-1)r+2},\ldots, \mathbf{n}_{jr}]\\
	&=  \left[\begin{matrix}
	\left[\hat{\mathbf{W}}_{j,r_c-1} \mathbf{Y}_{jr,1}
	\right]^T \\
	\left[\hat{\mathbf{W}}_{j,r_c-1} \mathbf{Y}_{jr,2}\right]^T\\
	\ldots\\
	\left[\hat{\mathbf{W}}_{j,r_c-1} \mathbf{Y}_{jr,d/r}
	\right]^T
	\end{matrix}\right] 
	+[\mathbf{n}_{(j-1)r+1},\mathbf{n}_{(j-1)r+2},\ldots, \mathbf{n}_{jr}].
	\end{split}
	\end{equation*}
	According to Lemma \ref{Thm:LBCDetection} and its assumptions, 
	with probability equal to unity, the index set $V$ contains all adversarial node indices and thus $U$ contains all the non-adversarial node indices.
	Thus, if $k\in U$, then $\mathbf{n}_{(j-1)r+k} = \mathbf{0}$, and the $k$th column of $\mathbf{R}_j$ only contains the first term in the above equation. 
	More precisely, we have
	\begin{equation*}
	\begin{split}
	\left[\mathbf{R}_j\right]_{\cdot,U} & = [\mathbf{z}_{(j-1)r+1},\mathbf{z}_{(j-1)r+2},\ldots, \mathbf{z}_{jr}]_{\cdot,U} 
	=  \left[\begin{matrix}
	\left[\hat{\mathbf{W}}_{j,r_c-1} \mathbf{Y}_{jr,1}
	\right]^T \\
	\left[\hat{\mathbf{W}}_{j,r_c-1} \mathbf{Y}_{jr,2}\right]^T\\
	\ldots\\
	\left[\hat{\mathbf{W}}_{j,r_c-1} \mathbf{Y}_{jr,d/r}
	\right]^T
	\end{matrix}\right]_{\cdot,U} \\
	&=  \left[\begin{matrix}
	\mathbf{Y}_{jr,1}^T
	\left[\hat{\mathbf{W}}_{j,r_c-1}\right]^T_{\cdot,U} \\
	\mathbf{Y}_{jr,2}^T
	\left[\hat{\mathbf{W}}_{j,r_c-1}\right]^T_{\cdot,U} \\
	\ldots\\
	\mathbf{Y}_{jr,d/r}^T
	\left[\hat{\mathbf{W}}_{j,r_c-1}\right]^T_{\cdot,U} \\
	\end{matrix}\right]	
	=  \left[\begin{matrix}
	\mathbf{Y}_{jr,1}^T\\
	\mathbf{Y}_{jr,2}^T \\
	\ldots\\
	\mathbf{Y}_{jr,d/r}^T \\
	\end{matrix}\right] \left[\hat{\mathbf{W}}_{j,r_c-1}\right]^T_{\cdot,U}.
	\end{split}
	\end{equation*}
	Note that and $r_c$ rows of  $\left[\hat{\mathbf{W}}_{j,r_c-1}\right]^T_{}$ is $\left[\hat{\mathbf{W}}_{j,r_c-1}\right]^T_{\cdot,U}$ must be invertible.
	We can then multiply each side by the inverse of $\left[\hat{\mathbf{W}}_{j,r_c-1}\right]^T_{\cdot,U}$, yielding
	\begin{equation*}
	\begin{split}
	\left[\mathbf{R}_j\right]_{\cdot,U} \left[\hat{\mathbf{W}}^{-T}_{j,r_c-1}\right]^{-1}_{\cdot,U} 
	&=  \left[\begin{matrix}
	\mathbf{Y}_{jr,1}^T\\
	\mathbf{Y}_{jr,2}^T \\
	\ldots\\
	\mathbf{Y}_{jr,d/r}^T \\
	\end{matrix}\right].
	\end{split}
	\end{equation*}
	Vectorizing it gives the desired gradients, we obtain
	\begin{equation*}
	\begin{split}
	\mathit{vec}\left(\left[\mathbf{R}_j\right]_{\cdot,U} \left[\hat{\mathbf{W}}_{j,r_c-1}^{T}\right]^{-1}_{\cdot,U} \right)
	&=  \left[\begin{matrix}
	\mathbf{Y}_{jr,1}\\
	\mathbf{Y}_{jr,2}\\
	\ldots\\
	\mathbf{Y}_{jr,d/r}
	\end{matrix}\right]	= \mathbf{Y}_{jr}\mathbf{1}_P.
	\end{split}
	\end{equation*}
\end{proof}

\subsection{Proof of Theorem \ref{Thm:LinearBlockCodeOptimality}}
\begin{proof}
Recall that Lemma \ref{Thm:LBCDetection} implies that with probability equal to unity,   $V = \phi(\mathbf{R}_j,j) =  \{i:\|\mathbf{n}_{j(r-1)+i}\|_0 \not=0\}$. 
Since $\left\vert{\{j:\|\mathbf{n}_{j}\|_0 \not=0\}}\right\vert$ $\leq s$, we must have $|V|\leq s$. Therefore, $|U| = |\{1,2,\ldots, r\}-V| \geq r-s$, and $U$ only contains the non-adversarial nodes with probability equal to unity. By Lemma \ref{thm:LBClocalcorrect}, we have $\mathbf{u}_j = \sum_{k=(j-1)r+1}^{jr}\mathbf{g}_{k}$ for $j=1,2,\ldots, P/r$.
Therefore, 
\begin{equation*}
\sum_{j=1}^{P/r}\mathbf{u}_j = \sum_{j=1}^{P/r}  \sum_{k=(j-1)r+1}^{jr} \mathbf{g}_{k} = \sum_{j=1}^{P} \mathbf{g}_j = \mathbf{G}\mathbf{1}_P.
\end{equation*}
This shows we correctly recover the sum of the gradient updates, while tolerating $s$ Byzantine nodes and ensuring communication compression ratio $r_c$.

Now let us consider the encoder and decoder complexity. 
For the encoder, first $\mathbf{Y}_j \mathbf{1}_P$ takes $\mathcal{O}(dr)$ flops, i.e., elementary addition and multiplication operations. Second, directly computing  $\mathbf{W}_j (\mathbf{Y}_j\mathbf{1}_P)$ takes  $\mathcal{O}(d_c d)$ flops. However, by definition  $\mathbf{W}_j = \mathbf{I}_{d_c} \otimes [1, w_j, w_j^2, \ldots, w_j^{r_c-1}]  $ is a sparse matrix with $r_c$ nonzeros per row. Therefore, using sparse matrix computation, it only takes $\mathcal{O}(d_c r_c) = \mathcal{O}(d) $ computations. 
Thus, the encoder function needs in total $\mathcal{O}(d r) + \mathcal{O}(d)= \mathcal{O}(dr )$ computations.
For the decoder, obtaining the adversarial node indices needs $\mathcal{O}(d_c r)$ for computing $\mathbf{r}_{j,c} = \mathbf{f} \mathbf{R}_j$. Solving the linear system takes $\mathcal{O}(r^3)$. Computing the polynomial needs $$\mathcal{O}\left(\frac{r_c+s-1}{s} (r_c+s-1+s)\right) = \mathcal{O}\left(r_c^2/s+r_c+s\right).$$ 
Evaluating the polynomial takes $\mathcal{O}(r (r_c-1)) = \mathcal{O}(r r_c)$. Computing $\left[\hat{W}^T_{j,r_c-1}\right]_{U,\cdot}^{-1}$ needs $\mathcal{O}(r_c^3)$, and computing $\left[\mathbf{R}_{j}\right]_{\cdot, U} \left[\hat{W}^T_{j,r_c-1}\right]_{U,\cdot}^{-1}$
needs $ \mathcal{O}(d_c r_c^2)$. Therefore, in total we need 
$$\mathcal{O}(d_c r) + \mathcal{O}(r^3) + \mathcal{O}(r_c^2/s+r_c+s)  + \mathcal{O}(r r_c)+ \mathcal{O}(r_c^3) + \mathcal{O}(d_c r_c^2) = \mathcal{O}(d_cr + r^3+d_c r_c^2).$$
There are in total $P/r$ iterations, so we have in total
$$P/r \cdot \mathcal{O}(d_c r + r^3+d_c r_c^2 ) = \mathcal{O}(Pr^2+ P d_c+ P d_c r_c^2/r)$$ 
flops. The final sum of all $\mathbf{u}_j$ takes $\mathcal{O}(P/r \cdot d) = \mathcal{O}(Pd/r).$
Thus, in total we have 

$$\mathcal{O}(Pd_c + Pr^2 + Pd_c r_c^2/r + P dr_c/r) = \mathcal{O}(Pd/r_c+ Pr^2+Pd r_c/r).$$ 

Suppose $d \gg P$, i.e., $d=\mathbf{\Omega}(P^2)$. 
Note that the redundancy ratio cannot be larger than $P$, we know $r\leq P$ and thus, $r^2 \leq P^2 = \mathbf{\Omega}(d)$. This implies that  
$$\mathcal{O}(Pd_c + Pr^2 + Pd_c r_c^2/r + P dr_c/r) = \mathcal{O}(Pd/r_c+ Pr^2+Pd r_c/r) = \mathcal{O}(Pd (1+\frac{1}{r_c}+\frac{r_c}{r})) $$ 
Thus, we have shown that the complexity of encoder and function is linear in the dimension of the gradient and the redundancy ratio, while that of the decoder is linear in the dimension of the gradient and the number of computing nodes $P$. 
\end{proof}

\eat{
\subsection{Proof of Theorem \ref{Thm:IndependenceCorrectness} }
\begin{proof}
Note that  $\mathbf{z}_j = E_j^{\mathit{Rep}}(\mathbf{Y}^{Rep}_j) = \mathbf{W}_j \mathbf{Y}^{Rep}_j\mathbf{1}_P$.
The received vector from the $j$th node is $\mathbf{z}_j+\mathbf{n}_j^*$, where $\mathbf{n}_j^*$ is some unknown value if the $j$th node is Byzantine (and is $\mathbf{0}$ if the $j$th node is normal).
By definition of $\mathbf{R}_j$, 

\begin{equation*}
\textrm{vec}\left(\mathbf{R}_j\right) = \begin{bmatrix}
\mathbf{z}_{j(r-1)+1} + \mathbf{n}^*_{j(r-1)+1}\\
\mathbf{z}_{j(r-1)+2} + \mathbf{n}^*_{j(r-1)+2}\\
\ldots\\
\mathbf{z}_{jr} + \mathbf{n}_{jr}^*
\end{bmatrix}
\end{equation*}
By definition of allocation matrix $\mathbf{A}$, $\mathbf{A}_{j(r-1)+\ell,\cdot} = \mathbf{A}_{jr,\cdot}$, and thus $ \mathbf{Y}_{j(r-1)+\ell}^{Rep} = \mathbf{Y}_{jr}^{Rep},   \ell=1,2,\ldots, r$.
Note that each node sends $\mathbf{z}_j = E_j^{\mathit{Rep}}(\mathbf{Y}^{Rep}_j)$ to the PS.
This implies that 
\begin{equation*}
\begin{split}
\textrm{vec}\left(\mathbf{R}_j\right) & =  \begin{bmatrix}
\mathbf{z}_{j(r-1)+1} + \mathbf{n}^*_{j(r-1)+1}\\
\mathbf{z}_{j(r-1)+2} + \mathbf{n}^*_{j(r-1)+2}\\
\ldots\\
\mathbf{z}_{jr} + \mathbf{n}_{jr}^*
\end{bmatrix}
=
\begin{bmatrix}
\mathbf{z}_{j(r-1)+1} \\
\mathbf{z}_{j(r-1)+2} \\
\ldots\\
\mathbf{z}_{jr} 
\end{bmatrix} + \begin{bmatrix}
 \mathbf{n}^*_{j(r-1)+1}\\
 \mathbf{n}^*_{j(r-1)+2}\\
\ldots\\
\mathbf{n}_{jr}^*
\end{bmatrix} \\
& = \begin{bmatrix}
\mathbf{W}_{j(r-1)+1} \mathbf{Y}^{Rep}_{j(r-1)+1}\mathbf{1}_P \\
\mathbf{W}_{j(r-1)+2} \mathbf{Y}^{Rep}_{j(r-1)+2}\mathbf{1}_P \\
\ldots\\
\mathbf{W}_{jr} \mathbf{Y}^{Rep}_{jr}\mathbf{1}_P \\
\end{bmatrix} + \begin{bmatrix}
 \mathbf{n}^*_{j(r-1)+1}\\
 \mathbf{n}^*_{j(r-1)+2}\\
\ldots\\
\mathbf{n}_{jr}^*
\end{bmatrix}\\
& = \begin{bmatrix}
\mathbf{W}_{j(r-1)+1} \mathbf{Y}^{Rep}_{jr}\mathbf{1}_P \\
\mathbf{W}_{j(r-1)+2} \mathbf{Y}^{Rep}_{jr}\mathbf{1}_P \\
\ldots\\
\mathbf{W}_{jr} \mathbf{Y}^{Rep}_{jr}\mathbf{1}_P \\
\end{bmatrix} + \begin{bmatrix}
 \mathbf{n}^*_{j(r-1)+1}\\
 \mathbf{n}^*_{j(r-1)+2}\\
\ldots\\
\mathbf{n}_{jr}^*
\end{bmatrix}\\
& = \begin{bmatrix}
\mathbf{W}_{j(r-1)+1}  \\
\mathbf{W}_{j(r-1)+2} \\
\ldots\\
\mathbf{W}_{jr} \\
\end{bmatrix} \mathbf{Y}^{Rep}_{jr}\mathbf{1}_P + \begin{bmatrix}
 \mathbf{n}^*_{j(r-1)+1}\\
 \mathbf{n}^*_{j(r-1)+2}\\
\ldots\\
\mathbf{n}_{jr}^*
\end{bmatrix}\\
& = \bar{\mathbf{W}}_j \mathbf{Y}^{Rep}_{jr}\mathbf{1}_P + \begin{bmatrix}
 \mathbf{n}^*_{j(r-1)+1}\\
 \mathbf{n}^*_{j(r-1)+2}\\
\ldots\\
\mathbf{n}_{jr}^*
\end{bmatrix}\\
\end{split}
\end{equation*}
Define $\mathbf{n}^*_j = \left[\mathbf{n}^{*T}_{j(r-1)+1}, \mathbf{n}^{*T}_{j(r-1)+2}, \ldots, \mathbf{n}^{*T}_{jr}\right]^T$.
Then the above equation can be written as $\bar{\mathbf{W}}_j \mathbf{Y}_{j r}^{Rep}\mathbf{1}_{P} + \mathbf{n}^*_j = \textit{vec}(\mathbf{R}_j)$.
Since $\bar{\mathbf{W}}_j \mathbf{u}_j + \mathbf{n}_j = \textit{vec}(\mathbf{R}_j)$, we have $\bar{\mathbf{W}}_j \mathbf{Y}_{j r}^{Rep}\mathbf{1}_{P} + \mathbf{n}^*_j = 
\bar{\mathbf{W}}_j \mathbf{u}_j + \mathbf{n}_j $ and thus  $\bar{\mathbf{W}}_j \left[\mathbf{Y}_{j r}^{Rep}\mathbf{1}_{P} - \mathbf{u}_j  \right]= 
 \mathbf{n}_j - \mathbf{n}^*_j$.
 Since there are at most $s$ Byzantine nodes, $\|\mathbf{n}^*_j\|_0 \leq s d_c$. By definition,  $\|\mathbf{n}_j\|_0 \leq s d_c$. Thus,
 $\|\mathbf{n}^*_j - \mathbf{n}_j\|_0 \leq 2 s r$. Thus, at least $(r-2s)d_c \geq r_c d_c = d$ elements in $\mathbf{n}^*_j - \mathbf{n}_j$ are 0. Let $\hat{U}$ be the set of indices of those $d$ zero elements. 
 Then we have $\bar{\mathbf{W}}_{j_{\hat{U},\cdot}} \left[\mathbf{Y}_{j r}^{Rep}\mathbf{1}_{P} - \mathbf{u}_j  \right]= 
 \left[\mathbf{n}_j - \mathbf{n}^*_j\right]_{\hat{U},\cdot}=\mathbf{0}$. By assumption, any $d$ rows of $\bar{\mathbf{W}}_{j}$ are linearly independent, and thus the $d\times d $ matrix $\bar{\mathbf{W}}_{j_{\hat{U},\cdot}}$ is invertible. This implies 
    $\mathbf{Y}_{j r}^{Rep}\mathbf{1}_{P} - \mathbf{u}_j = \mathbf{0}$, which finishes the proof.
\end{proof}

TODO: Clean the below.
\subsection{Proof of Theorem \ref{Thm:RedundancyRatioBound} }
For simplicity of proof, let us define a valid $s$-attack first.
\begin{definition}
	$\mathbf{N} =[\mathbf{n}_1,\mathbf{n}_2,\ldots, \mathbf{n}_P]$ is a valid $s$-attack if and only if $\left\vert{\{j: \|\mathbf{n}_{j}\|_0\not= 0 \}}\right\vert \leq s$. 
\end{definition}
Now we prove theorem \ref{Thm:RedundancyRatioBound}.
Suppose $(\mathbf{A}, E, D)$ can resist $s$ adversaries.
The goal is to prove 
$\|A\|_0 \geq P(2s+1)$.
    In fact we can prove a slightly stronger version: $\|\mathbf{A}_{\cdot,i}\|_0 \geq  \left(2s+1\right), i = 1,2,\ldots, B$.
	Suppose for some $i$,  $\|\mathbf{A}_{\cdot,i}\|_0 = \tau <  \left(2s+1\right)$.
	Without loss of generality, assume that $\mathbf{A}_{1, i}, \mathbf{A}_{2, i}, \mathbf{A}_{\tau,i} $ are nonzero. Let $\mathbf{G}_{-i} = [\mathbf{g}_1,\mathbf{g}_2,\ldots,\mathbf{g}_{i-1},\mathbf{g}_{i+1},\ldots, \mathbf{g}_P]$. 
	Since $(\mathbf{A},E,D)$ can protect against $s$ adversaries,  we have for any $\mathbf{G}$, 	\begin{align*}
		D(\mathbf{Z}^{\mathbf{A},E,\mathbf{G}}+\mathbf{N} ) = \mathbf{G} \mathbf{1}=  \mathbf{G}_{-i} \mathbf{1} + \mathbf{g}_i,
	\end{align*}
	for any valid $s$-attack $\mathbf{N}$.
	In particular, let $\mathbf{g}^1_i = \mathbf{1}_{d}$,  $\mathbf{g}^i_2 = -\mathbf{1}_{d}$,
	$\mathbf{G}^1 = [\mathbf{g}_1,\mathbf{g}_2,\ldots,\mathbf{g}_{i-1},\mathbf{g}_{i}^1,\mathbf{g}_{i+1},\ldots, \mathbf{g}_P]$, and $\mathbf{G}^2 = [\mathbf{g}_1,\mathbf{g}_2,\ldots,\mathbf{g}_{i-1},\mathbf{g}_{i}^2,\mathbf{g}_{i+1},\ldots, \mathbf{g}_P]$.  
	Then for any valid $s$-attack $\mathbf{N}^1, \mathbf{N}^2$,
	\begin{align*}
		D(\mathbf{Z}^{\mathbf{A},E,\mathbf{G}^1}+\mathbf{N}^1 ) =  \mathbf{G}_{-i} \mathbf{1}_{P-1} + \mathbf{1}_d.
	\end{align*}	
	and
	\begin{align*}
		D(\mathbf{Z}^{\mathbf{A},E,\mathbf{G}^2}+\mathbf{N}^2 ) =  \mathbf{G}_{-i} \mathbf{1}_{P-1} - \mathbf{1}_d.
	\end{align*}	
	Our goal is to find $\mathbf{N}^1, \mathbf{N}^2$ such that $D(\mathbf{Z}^{\mathbf{A},E,\mathbf{G}^1}+\mathbf{N}^1 ) = D(\mathbf{Z}^{\mathbf{A},E,\mathbf{G}^2}+\mathbf{N}^2 )$ which then will lead to a contradiction.
	Construct $\mathbf{N}^1$ and $\mathbf{N}^2$ by
	\begin{equation*}
	\mathbf{N}^1_{\ell,j}= 
	\begin{cases}
	\left[\mathbf{Z}^{\mathbf{A},E, \mathbf{G}^2}\right]_{\ell,j} - 	\left[\mathbf{Z}^{\mathbf{A},E, \mathbf{G}^1}\right]_{\ell,j} ,&j=1,2,\ldots, \ceil{\frac{\tau-1}{2}}\\
	0,              & \text{otherwise}
	\end{cases}
	\end{equation*}
	and
	\begin{equation*}
	\mathbf{N}^2_{\ell,j}= 
	\begin{cases}
		\left[\mathbf{Z}^{\mathbf{A},E, \mathbf{G}^1}\right]_{\ell,j} - 	\left[\mathbf{Z}^{\mathbf{A},E, \mathbf{G}^2}\right]_{\ell,j}, & j= \ceil{\frac{\tau-1}{2}}, \ceil{\frac{\tau-1}{2}}+1,\ldots, \tau\\
	0,              & \text{otherwise}
	\end{cases}
	\end{equation*}
	One can easily verify that $\mathbf{N}^1, \mathbf{N}^2$ are both valid $s$-attack. Meanwhile, we have
	\begin{align*}
		\left[\mathbf{Z}^{\mathbf{A},E,\mathbf{G}^1}\right]_{\ell,j}+\mathbf{N}^1_{{\ell,j}} = \left[\mathbf{Z}^{\mathbf{A},E,\mathbf{G}^2}\right]_{\ell,j}+\mathbf{N}^2_{{\ell,j}}, j=1,2,\ldots, \tau
	\end{align*}	 
	due to the above construction of $\mathbf{N}^1, \mathbf{N}^2$.
	Note that $\mathbf{A}_{j,i}=0$ for all $j>\tau$, which implies that for all compute nodes with index $j>\tau$, their encoder functions do not depend on the $i$th gradient. Since $\mathbf{G}^1$ and $\mathbf{G}^2$ only differ in the $i$th gradient, the encoder function of any compute node with index $j>\tau$ should have the same output.
	Thus, we have  
	\begin{align*}
		\left[\mathbf{Z}^{\mathbf{A},E,\mathbf{G}^1}\right]_{\ell,j}+\mathbf{N}^1_{{\ell,j}} = \left[\mathbf{Z}^{\mathbf{A},E,\mathbf{G}^1}\right]_{\ell,j} = \left[\mathbf{Z}^{\mathbf{A},E,\mathbf{G}^2}\right]_{\ell,j} = \left[\mathbf{Z}^{\mathbf{A},E,\mathbf{G}^2}\right]_{\ell,j}+\mathbf{N}^2_{{\ell,j}}, j> \tau
	\end{align*}
	Hence, we have 		
	\begin{align*}
		\left[\mathbf{Z}^{\mathbf{A},E,\mathbf{G}^1}\right]_{\ell,j}+\mathbf{N}^1_{{\ell,j}} =  \left[\mathbf{Z}^{\mathbf{A},E,\mathbf{G}^2}\right]_{\ell,j}+\mathbf{N}^2_{{\ell,j}}, \forall j
	\end{align*}
	which means
	\begin{align*}
		\mathbf{Z}^{\mathbf{A},E,\mathbf{G}^1}+\mathbf{N}^1 = \mathbf{Z}^{\mathbf{A},E,\mathbf{G}^2}+\mathbf{N}^2
	\end{align*}
Therefore, we have 
\begin{align*}
    D(\mathbf{Z}^{\mathbf{A},E,\mathbf{G}^1}+\mathbf{N}^1 ) = D(\mathbf{Z}^{\mathbf{A},E,\mathbf{G}^2}+\mathbf{N}^2 )
\end{align*}
	and thus
\begin{align*}
    \mathbf{G}_{-1} \mathbf{1}_{P-1} + \mathbf{1}_d = D(\mathbf{Z}^{\mathbf{A},E,\mathbf{G}^1}+\mathbf{N}^1 ) = D(\mathbf{Z}^{\mathbf{A},E,\mathbf{G}^2}+\mathbf{N}^2 ) = \mathbf{G}_{-1} \mathbf{1}_{P-1} - \mathbf{1}_d 
\end{align*}
This gives us a contradiction. Hence, the assumption is not correct and we must have $\|\mathbf{A}_{\cdot,i}\|_0 \geq  \left(2s+1\right), i = 1,2,\ldots, P$.	
Thus, we must have $\|A\|_0 \geq (2s+1)P$. \qed
	
A direct but interesting corollary of this theorem is a bound on the number of adversaries \solon{} can resist.

\begin{corollary}\label{Cor:AdversarialBound}
	$(\mathbf{A}, E, D)$ can resist at most $\frac{P-1}{2}$ adversarial nodes.
\end{corollary}
\begin{proof}
According to Theorem \ref{Thm:RedundancyRatioBound}, the redundancy ratio is at least $2s+1$, meaning that every data point must be replicated by at least $2s+1$. 
Since there are  $P$ compute node in total, we must have $2s+1\leq P$, which implies $s\leq \frac{P-1}{2}$. 
Thus, $(\mathbf{A},E,D)$ can resist at most $\frac{P-1}{2}$ adversaries.
\end{proof}
In other words, at least a majority of the compute nodes must be non-adversarial. \qed

\subsection{Proof of Theorem \ref{Thm:RepCodeOptimality} }
Since there are at most $s$ adversaries, there are at least $2s+1-s=s+1$ non-adversarial compute nodes in each group.
Thus, performing majority vote on each group returns the correct gradient, and thus the repetition code guarantees that the result is correct.
The complexity at each compute node is clearly $\mathcal{O}((2s+1)d)$ since each of them only computes the sum of $(2s+1)$ $d$-dimensional gradients.
For the decoder at the PS, within each group of $(2s+1)$ machine, it takes $\mathcal{O}((2s+1) d )$ computations to find the majority. Since there are $\frac{P}{(2s+1)}$ groups, it takes in total $\mathcal{O}((2s+1) d \frac{P}{(2s+1)} ) = \mathcal{O}(P d )$ computations. 
Thus, this achieves linear-time encoding and decoding. \qed

\subsection{Proof of Lemma \ref{Thm:CycCodeColumnSpan} }
We first prove that $\mathbf{A}_{j,k} = 0 \Rightarrow \mathbf{W}_{j,k} = 0$.

Suppose $\mathbf{A}_{j,k} = 0$ for some $j,k$.
Then by definition $k \in \alpha_j$.
By $\mathbf{0} = \begin{matrix} \begin{bmatrix}
\mathbf{q}_j & 1
\end{bmatrix}\end{matrix} \cdot  \left[\mathbf{C}_{L}\right]_{\cdot,\alpha_j} $ we have $ 0 =\begin{matrix} \begin{bmatrix}
\mathbf{q}_j & 1
\end{bmatrix}\end{matrix} \left[\mathbf{C}_L\right]_{\cdot,k} =\mathbf{W}_{j,k}$. 

Next we prove that for any index set $U$ such that $|U|\geq P-(2s+1)$, the column span of $\mathbf{W}_{\cdot,U}$ contains $\mathbf{1}$.
This is equivalent to that for any index set $U$ such that $|U|\geq P-(2s+1)$, there exists a vector $\mathbf{b}$ such that $\mathbf{W}_{\cdot,U} b = \mathbf{1}$. 
Now we show such $b$ exists.
Note that $\mathbf{C}_{L}$ is a $(P-2s)\times P$ full rank Vandermonde matrix and thus any $P-2s$ columns of 
$\mathbf{C}_{L}$ are linearly independent.
Let $\bar{U}$ be the first $P-2s$ elements in $U$.
Then all columns of $\left[\mathbf{C}_{L}\right]_{\cdot,\bar{U}}$ are linearly independent and thus $\left[\mathbf{C}_{L}\right]_{\cdot, \bar{U}}$ is invertible.
Let $\mathbf{b}_{\bar{U}} \triangleq \mathbf{\bar{b}} = \left(C^L_{\bar{U}}\right)^{-1} \begin{matrix} \begin{bmatrix}
0 & 0 & \ldots & 0 & 1
\end{bmatrix}\end{matrix}^T $.
For any $j \not\in \bar{U}$, let
 $\mathbf{b}_{j} = 0$.
Then we have 

\begin{equation*}
\begin{split}
\mathbf{W}_{U}  \mathbf{b} & = \begin{matrix} \begin{bmatrix}
		\mathbf{Q} & \mathbf{1}
	\end{bmatrix}\end{matrix} \times  \left[\mathbf{C}_L\right]_{\cdot,U} \mathbf{b} \\
	& =	\begin{matrix} \begin{bmatrix}
			\mathbf{Q} & \mathbf{1}
		\end{bmatrix}\end{matrix} \times  \left[\mathbf{C}_L\right]_{\cdot,\bar{U}} \mathbf{\bar{b}} \\	
	& = \begin{matrix} \begin{bmatrix}
			\mathbf{Q} & \mathbf{1}
		\end{bmatrix}\end{matrix} \times  \left[\mathbf{C}_L\right]_{\cdot,\bar{U}} \times \left[\mathbf{C}_L\right]_{\cdot,\bar{U}}^{-1} \begin{matrix} \begin{bmatrix}
			0 & 0 & \ldots & 0 & 1
		\end{bmatrix}\end{matrix}^T\\	
	& = \begin{matrix} \begin{bmatrix}
			\mathbf{Q} & \mathbf{1}
		\end{bmatrix}\end{matrix} \begin{matrix} \begin{bmatrix}
		0 & 0 & \ldots & 0 & 1
	\end{bmatrix}\end{matrix}^T\\
	& = \mathbf{1}.\\				
\end{split}
\end{equation*}
This completes the proof. \qed

\subsection{Proof of Lemma \ref{Thm:CycCodeDetection} }
We need a few lemmas first.
\begin{lemma}\label{Lemma:CycProb}
	Let a $P$-dimensional vector $\mathbf{\gamma} \triangleq \left[\gamma_1, \gamma_2,\ldots, \gamma_P \right]^T = \left(\mathbf{f} \mathbf{N}\right)^T$. Then we have 
	\begin{align*}
	\textit{Pr}( \{ j : \gamma_j \not= 0 \} = \{j: \| \mathbf{N}_{\cdot,j} \|_0 \not= 0 \}) = 1.
	\end{align*}
\end{lemma}
\begin{proof}
Let us prove that 
	\begin{align*}
		\textit{Pr}( \mathbf{N}_{\cdot,j} \not= 0 \} | \gamma_j \not = 0 ) = 1.
	\end{align*}
	and 
	\begin{align*}
		\textit{Pr}( \gamma_j \not = 0 | \mathbf{N}_{\cdot,j} \not= 0 \} ) = 1.
	\end{align*}
	for any $j$.	
	Combining those two equations we prove the lemma.
	
	The first equation is straightforward. Suppose $\mathbf{N}_{\cdot,j} = 0$.
	Then we immediately have $\gamma_j = \mathbf{f} \mathbf{N}_{\cdot,j} = 0$.
	For the second one, note that $\mathbf{f}$ has entries drawn independently from the standard normal distribution. Therefore we have that $\gamma_j = \mathbf{f} \mathbf{N}_{\cdot,j} \sim \mathcal{N}(\mathbf{1}^T \mathbf{N}_{\cdot,j}, \|\mathbf{N}_{\cdot,j}\|_2^2 )$.
	Since $\gamma_j$ is a random variable with normal distribution, the probability of it being any particular value is $0$.
	In particular, 
	\begin{align*}
		\textit{Pr}( \gamma_j = 0 | \mathbf{N}_{\cdot,j} \not= 0 \} ) = 0,
	\end{align*}	
	and thus 
	\begin{align*}
		\textit{Pr}( \gamma_j \not= 0 | \mathbf{N}_{\cdot,j} \not= 0 \} ) = 1
	\end{align*}	
	which proves the second equation and finishes the proof.
\end{proof}

\begin{lemma}\label{Lemma:Orthogonal}
	$\mathbf{R}^{Cyc} \mathbf{C}_{R}^\dag = \mathbf{N} \mathbf{C}_{R}^\dag$.
\end{lemma}
\begin{proof}
By definition, $\mathbf{R}^{Cyc} \mathbf{C}_{R}^\dag = \left( \mathbf{G} \mathbf{W} + \mathbf{N} \right) \mathbf{C}_{R}^\dag = \left( \mathbf{G} \begin{matrix} \begin{bmatrix}
\mathbf{Q} & \mathbf{1}
\end{bmatrix}\end{matrix}  \mathbf{C}_L + \mathbf{N} \right) \mathbf{C}_{R}^\dag = \mathbf{G} \begin{matrix} \begin{bmatrix}
\mathbf{Q} & \mathbf{1}
\end{bmatrix}\end{matrix}   \mathbf{C}_L \mathbf{C}_{R}^\dag + \mathbf{N} \mathbf{C}_{R}^\dag  = \mathbf{N} \mathbf{C}_{R}^\dag$.
In the last equation we use the fact that IDFT matrix is unitary and thus $\mathbf{C}_L \mathbf{C}_{R}^\dag = \mathbf{0}_{(P-2s)\times(2s)}$.
\end{proof}

\begin{lemma}\label{Lemma:DFT}
	Let a $P$-dimensional vector $\hat{\mathbf{h}} \triangleq [\hat{h}_0, \hat{h}_1,\ldots, \hat{h}_{P-1}]^T$ be the discrete Fourier transformation (DFT) of a $P$-dimensional vector $\hat{\mathbf{t}} \triangleq [\hat{t}_{1},\hat{t}_{2},\ldots,\hat{t}_{P-1}]^T$ which has at most $s$ nonzero elements, i.e., $\hat{\mathbf{h}} = \mathbf{C}^{\dag} \hat{\mathbf{t}}$ and $\| \mathbf{t} \|_0 \leq s$.
	Then there exists a $s$-dimensional vector $\hat{\beta} \triangleq  [\hat{\beta}_{0},\hat{\beta}_{1},\ldots, \hat{\beta}_{s-1}]^T$, such that
	\begin{align}\label{eq:RecoverLinearSystem}
	\begin{matrix}
		\begin{bmatrix}
			\hat{h}_{P-s-1}      & \hat{h}_{P-s} & \dots & \hat{h}_{P-2} \\
			\hat{h}_{P-s-2}       & \hat{h}_{P-s-1} & \dots & \hat{h}_{P-3} \\
			\hdots & \hdots & \ddots &\vdots \\
			\hat{h}_{P-2s}  & \hat{h}_{P-s+1} & \dots & \hat{h}_{P-s-1}
		\end{bmatrix}
		\hat{\beta}
		=
		\begin{bmatrix}
			\hat{h}_{P-1}\\
			\hat{h}_{P-2}\\
			\vdots\\
			\hat{h}_{P-s}
		\end{bmatrix}
	\end{matrix}.
\end{align}
Furthermore, for any $\hat{\beta}$ satisfying the above equations, 
\begin{align}\label{eq:InductionEquation}
	\hat{h}_{\ell} = \sum_{u=0}^{s-1} \hat{\beta}_{u} \hat{h}_{\ell+u-s}, 
\end{align}
always holds for all $\ell$, where $\hat{h}_{\ell} = \hat{h}_{P+\ell}$.
\end{lemma}
\begin{proof}
Let $i_1,i_2,\ldots, i_s$ be the index of the nonzero elements in $\hat{\mathbf{t}}$.
Let us define the location polynomial $p(\omega) = \prod_{k=1}^{s}(\omega-e^{-\frac{2\pi i}{P}i_k}  )\triangleq \sum_{k=0}^{s} \theta_k \omega^k$, where $\theta_s = 1$.
Let a $s$-dimensional vector $\hat{\beta}^* \triangleq -[\theta_0,\theta_1,\ldots, \theta_{s-1}]^T$.

Now we prove that  $\hat{\beta}=\hat{\beta}^*$ is a solution to the system of linear equations \eqref{eq:RecoverLinearSystem}.
To see this, note that by definition, for any $\lambda$, we have $0 = p(e^{-\frac{2\pi i}{P}i_\lambda}) = \sum_{k=0}^{s} \theta_k e^{-\frac{2\pi i}{P}i_\lambda k}$.
Multiply both  side by $\hat{t}_{i_\lambda} e^{-\frac{2\pi i}{P}i_\lambda \eta }$, we have
\begin{align*}
0 &= \hat{t}_{i_\lambda} e^{-\frac{2\pi i}{P}i_\lambda \eta} \sum_{k=0}^{s} \theta_k e^{-\frac{2\pi i}{P}i_\lambda k}\\
&=\hat{t}_{i_\lambda} \sum_{k=0}^{s} \theta_k e^{-\frac{2\pi i}{P}i_\lambda (k+\eta)}.
\end{align*}
Summing over $\lambda$, we have 
\begin{align*}
	0 &=\sum_{\lambda = 1}^{s}\hat{t}_{i_\lambda} \sum_{k=0}^{s} \theta_k e^{-\frac{2\pi i}{P}i_\lambda (k+\eta)}\\
	&=\sum_{k=0}^{s} \theta_k \sum_{\lambda = 1}^{s}\hat{t}_{i_\lambda}   e^{-\frac{2\pi i}{P}i_\lambda (k+\eta)}.
\end{align*}
By definition, $\hat{h}_j = \mathbf{C}_{j,\cdot} \hat{\mathbf{t}} = \frac{1}{\sqrt{P}}\sum_{k=0}^{P-1} e^{-\frac{2\pi i }{P} j k} \hat{t}_k  =  \frac{1}{\sqrt{P}}\sum_{\lambda=1}^{s} \hat{t}_{i_\lambda} e^{-\frac{2\pi i}{P}  i_\lambda j} $.
Hence, the above equation becomes
\begin{align*}
	0 &=\sum_{k=0}^{s} \theta_k \sqrt{P} \hat{h}_{k+\eta}
\end{align*}
which is equivalent to
\begin{align*}
	\hat{h}_{s+\eta} &=\sum_{k=0}^{s-1} -\theta_k \hat{h}_{k+\eta}
\end{align*}
due to the fact that $\theta_s = 1$. 
By setting $\eta=-s+P-1,-s+P-2,\ldots,-s+P-s$, one can easily see that the above equation becomes identical to the system of linear equations in \eqref{eq:RecoverLinearSystem} with $\hat{\beta} =\hat{\beta}^*=-[\theta_0,\theta_1,\ldots, \theta_{s-1}]^T$.

Now let us prove for any $\hat{\beta}$ that satisfies equation \eqref{eq:RecoverLinearSystem}, we have \eqref{eq:InductionEquation}. 
Note that an equivalent form of \eqref{eq:InductionEquation} is that the following system of linear equations

	\begin{align}\label{eq:CycInduction}
		\begin{matrix}
			\begin{bmatrix}
				\hat{h}_{P-s-1+\ell}      & \hat{h}_{P-s+\ell} & \dots & \hat{h}_{P-2+\ell} \\
				\hat{h}_{P-s-2+\ell}       & \hat{h}_{P-s-1+\ell} & \dots & \hat{h}_{P-3+\ell} \\
				\hdots & \hdots & \ddots &\vdots \\
				\hat{h}_{P-2s+\ell}  & \hat{h}_{P-s+1+\ell} & \dots & \hat{h}_{P-s-1+\ell}
			\end{bmatrix}
			\hat{\beta}
			=
			\begin{bmatrix}
				\hat{h}_{P-1+\ell}\\
				\hat{h}_{P-2+\ell}\\
				\vdots\\
				\hat{h}_{P-s+\ell}
			\end{bmatrix}
		\end{matrix}
	\end{align}
	holds for $\ell=0,1,2\ldots,P-1$.
	We prove this by induction.
	When $\ell=1$, this is true since $\hat{\beta}$ satisfies the system of linear equations in \eqref{eq:RecoverLinearSystem}.
	Assume it holds for $\ell = \mu$, i.e.,
	\begin{align*}
		\begin{matrix}
			\begin{bmatrix}
				\hat{h}_{P-s-1+\mu}      & \hat{h}_{P-s+\mu} & \dots & \hat{h}_{P-2+\mu} \\
				\hat{h}_{P-s-2+\mu}       & \hat{h}_{P-s-1+\mu} & \dots & \hat{h}_{P-3+\mu} \\
				\hdots & \hdots & \ddots &\vdots \\
				\hat{h}_{P-2s+\mu}  & \hat{h}_{P-s+1+\mu} & \dots & \hat{h}_{P-s-1+\mu}
			\end{bmatrix}
			\hat{\beta}
			=
			\begin{bmatrix}
				\hat{h}_{P-1+\mu}\\
				\hat{h}_{P-2+\mu}\\
				\vdots\\
				\hat{h}_{P-s+\mu}
			\end{bmatrix}
		\end{matrix}
	\end{align*}	
	Now we need to prove it also holds  when $\ell=\mu+1$, \ie 
	\begin{align*}
		\begin{matrix}
			\begin{bmatrix}
				\hat{h}_{P-s-1+\mu+1}      & \hat{h}_{P-s+\mu+1} & \dots & \hat{h}_{P-2+\mu+1} \\
				\hat{h}_{P-s-2+\mu+1}       & \hat{h}_{P-s-1+\mu+1} & \dots & \hat{h}_{P-3+\mu+1} \\
				\hdots & \hdots & \ddots &\vdots \\
				\hat{h}_{P-2s+\mu+1}  & \hat{h}_{P-s+1+\mu+1} & \dots & \hat{h}_{P-s-1+\mu+1}
			\end{bmatrix}
			\hat{\beta}
			=
			\begin{bmatrix}
				\hat{h}_{P-1+\mu+1}\\
				\hat{h}_{P-2+\mu+1}\\
				\vdots\\
				\hat{h}_{P-s+\mu+1}
			\end{bmatrix}
		\end{matrix}.
	\end{align*}
	First, since both $\hat{\beta}, \hat{\beta}^*$ satisfy the induction assumption,	we must have 
	\begin{align*}
		\begin{matrix}
			\begin{bmatrix}
				\hat{h}_{P-s-1+\mu}      & \hat{h}_{P-s+\mu} & \dots & \hat{h}_{P-2+\mu} \\
				\hat{h}_{P-s-2+\mu}       & \hat{h}_{P-s-1+\mu} & \dots & \hat{h}_{P-3+\mu} \\
				\hdots & \hdots & \ddots &\vdots \\
				\hat{h}_{P-2s+\mu}  & \hat{h}_{P-s+1+\mu} & \dots & \hat{h}_{P-s-1+\mu}
			\end{bmatrix}
			\end{matrix}
			(\hat{\beta} -\hat{\beta}^*)
			=\mathbf{0}_{s}.
	\end{align*}	
	Due to the induction assumption, one can verify that 
	\begin{align*}
		[\theta_{s-1},\theta_{s-2},\ldots, \theta_{0}]
		\begin{matrix}
			\begin{bmatrix}
				\hat{h}_{P-s-1+\mu}      & \hat{h}_{P-s+\mu} & \dots & \hat{h}_{P-2+\mu} \\
				\hat{h}_{P-s-2+\mu}       & \hat{h}_{P-s-1+\mu} & \dots & \hat{h}_{P-3+\mu} \\
				\hdots & \hdots & \ddots &\vdots \\
				\hat{h}_{P-2s+\mu}  & \hat{h}_{P-s+1+\mu} & \dots & \hat{h}_{P-s-1+\mu}
			\end{bmatrix}
		\end{matrix}
		= 
		\begin{matrix}
			\begin{bmatrix}
		\hat{h}_{P-s+\mu}  & \hat{h}_{P-s+\mu+1} & \ldots
		\hat{h}_{P-2+\mu+1} 
			\end{bmatrix}
		\end{matrix},
	\end{align*}	
	and thus we have 
	\begin{align*}
		&\begin{matrix}
			\begin{bmatrix}
				\hat{h}_{P-s+\mu}  & \hat{h}_{P-s+\mu+1} & \ldots
				\hat{h}_{P-2+\mu+1} 
			\end{bmatrix}
		\end{matrix}
		(\hat{\beta} -\hat{\beta}^*)
		\\ = 	&[\theta_{s-1},\theta_{s-2},\ldots, \theta_{0}]
		\begin{matrix}
			\begin{bmatrix}
				\hat{h}_{P-s-1+\mu}      & \hat{h}_{P-s+\mu} & \dots & \hat{h}_{P-2+\mu} \\
				\hat{h}_{P-s-2+\mu}       & \hat{h}_{P-s-1+\mu} & \dots & \hat{h}_{P-3+\mu} \\
				\hdots & \hdots & \ddots &\vdots \\
				\hat{h}_{P-2s+\mu}  & \hat{h}_{P-s+1+\mu} & \dots & \hat{h}_{P-s-1+\mu}
			\end{bmatrix}
		\end{matrix}
		(\hat{\beta} -\hat{\beta}^*)
		=0.
	\end{align*}		
Hence,
\begin{align*}
		& \begin{matrix}
			\begin{bmatrix}
				\hat{h}_{P-s+\mu}      & \hat{h}_{P-s-1+\mu} & \dots & \hat{h}_{P-1+\mu} \\
			\end{bmatrix}
		\end{matrix}
			\hat{\beta}\\
			= & 	\begin{matrix}
				\begin{bmatrix}
					\hat{h}_{P-s+\mu}      & \hat{h}_{P-s-1+\mu} & \dots & \hat{h}_{P-1+\mu} \\
				\end{bmatrix}
			\end{matrix}
			\hat{\beta}^* + \begin{matrix}
				\begin{bmatrix}
					\hat{h}_{P-s+\mu}      & \hat{h}_{P-s-1+\mu} & \dots & \hat{h}_{P-1+\mu} \\
				\end{bmatrix}
			\end{matrix}
			(\hat{\beta} - \hat{\beta}^*) 
			=\hat{h}_{P+\mu} = \hat{h}_{P-1+\mu+1}.
	\end{align*}	
	Furthermore, by induction assumption, we have
	\begin{align*}
		& \begin{matrix}
			\begin{bmatrix}
				\hat{h}_{P-s-2+\mu+1}      & \hat{h}_{P-s-1+\mu+1} & \dots & \hat{h}_{P-3+\mu+1} \\
				\hat{h}_{P-s-3+\mu+1}       & \hat{h}_{P-s-2+\mu+1} & \dots & \hat{h}_{P-4+\mu+1} \\
				\hdots & \hdots & \ddots &\vdots \\
				\hat{h}_{P-2s+\mu+1}  & \hat{h}_{P-s+1+\mu+1} & \dots & \hat{h}_{P-s+1+\mu+1}
			\end{bmatrix}
			\hat{\beta}
		\end{matrix}	
		= \begin{matrix}
			\begin{bmatrix}
				\hat{h}_{P-s-1+\mu}      & \hat{h}_{P-s-2+\mu} & \dots & \hat{h}_{P-2+\mu} \\
				\hat{h}_{P-s-2+\mu}       & \hat{h}_{P-s-1+\mu} & \dots & \hat{h}_{P-3+\mu} \\
				\hdots & \hdots & \ddots &\vdots \\
				\hat{h}_{P-(2s-1)+\mu}  & \hat{h}_{P-s+\mu} & \dots & \hat{h}_{P-s+\mu}
			\end{bmatrix}
		\hat{\beta}
		\end{matrix}\\
		=&
			\begin{matrix}
			\begin{bmatrix}
				\hat{h}_{P-1+\mu}\\
				\hat{h}_{P-2+\mu}\\
				\vdots\\
				\hat{h}_{P-(s-1)+\mu}
			\end{bmatrix}
		\end{matrix}
		=
		\begin{matrix}
			\begin{bmatrix}
				\hat{h}_{P-2+(\mu+1)}\\
				\hat{h}_{P-3+(\mu+1)}\\
				\vdots\\
				\hat{h}_{P-s+(\mu+1)}
			\end{bmatrix}
		\end{matrix}.		
	\end{align*}	
	Combing those two result we have proved 
	\begin{align*}
		\begin{matrix}
			\begin{bmatrix}
				\hat{h}_{P-s-1+\mu+1}      & \hat{h}_{P-s+\mu+1} & \dots & \hat{h}_{P-2+\mu+1} \\
				\hat{h}_{P-s-2+\mu+1}       & \hat{h}_{P-s-1+\mu+1} & \dots & \hat{h}_{P-3+\mu+1} \\
				\hdots & \hdots & \ddots &\vdots \\
				\hat{h}_{P-2s+\mu+1}  & \hat{h}_{P-s+1+\mu+1} & \dots & \hat{h}_{P-s-1+\mu+1}
			\end{bmatrix}
			\hat{\beta}
			=
			\begin{bmatrix}
				\hat{h}_{P-1+\mu+1}\\
				\hat{h}_{P-2+\mu+1}\\
				\vdots\\
				\hat{h}_{P-s+\mu+1}
			\end{bmatrix}
		\end{matrix}.
	\end{align*}	
	By induction, the equation \ref{eq:CycInduction} holds for all $\ell=0,1,\ldots,P-1$.
	Equation \ref{eq:CycInduction} immediately finishes the proof.	
\end{proof}

Now we are ready to prove Lemma \ref{Thm:CycCodeDetection}.
By Lemma \ref{Lemma:CycProb}, for the $P$-dimensional vector $\gamma = \left(\mathbf{f} \mathbf{N} \right)^T$, we have  
\begin{align*}
	\textit{Pr}( \{ j : \gamma_j \not= 0 \} = \{j: \| \mathbf{N}_{\cdot,j} \|_0 \not= 0 \}) = 1,
\end{align*}

Since there are at most $s$ adversaries, the number of nonzero columns in $\mathbf{N}$ is at most $s$ and hence there are at most $s$ nonzero elements in $\gamma$, i.e., $\|\gamma\|_0\leq s$, with probability 1.
Now consider the case when $\|\gamma\|_0\leq s$.
First note that $[h_{P-2s}, h_{P-2s+1},\ldots, h_{P-1}] = \mathbf{f} \mathbf{R}^{Cyc} \mathbf{C}_{R}^{\dag} = \mathbf{f} \mathbf{N} \mathbf{C}_{R}^{\dag} = \gamma^T \mathbf{C}_{R}^{\dag}$, where the second equation is due to Lemma \ref{Lemma:Orthogonal}.
Now let us construct $\hat{\mathbf{h}} = [\hat{h}_0,\hat{h}_1,\ldots, \hat{h}_{P-1}]^T$ by $\hat{\mathbf{h}} = \mathbf{C}^{\dag} \gamma$.
Note that $\mathbf{C}$ is symmetric and thus $\mathbf{C}^{\dag} = \left[\mathbf{C}^{\dag}\right]^T$. One can easily verify that $\hat{h}_{\ell} = {h}_{\ell},\ell = P-2s, P-2s+1,\ldots,P-1$.
Therefore, the equation 
\begin{align*}
	\begin{matrix}
		\begin{bmatrix}
			h_{P-s-1}      & h_{P-s} & \dots & h_{P-2} \\
			h_{P-s-2}       & h_{P-s-1} & \dots & h_{P-3} \\
			\hdots & \hdots & \ddots &\vdots \\
			h_{P-2s}  & h_{P-s+1} & \dots & h_{P-s+1}
		\end{bmatrix}
		\begin{bmatrix}
			\beta_{0}\\
			\beta_{1}\\
			\vdots\\
			\beta_{s-1}
		\end{bmatrix}
		=
		\begin{bmatrix}
			h_{P-1}\\
			h_{P-2}\\
			\vdots\\
			h_{P-s}
		\end{bmatrix}
	\end{matrix}
\end{align*}
becomes 
\begin{align*}
	\begin{matrix}
		\begin{bmatrix}
			\hat{h}_{P-s-1}      & h_{P-s} & \dots & \hat{h}_{P-2} \\
			\hat{h}_{P-s-2}       & \hat{h}_{P-s-1} & \dots & \hat{h}_{P-3} \\
			\hdots & \hdots & \ddots &\vdots \\
			\hat{h}_{P-2s}  & \hat{h}_{P-s+1} & \dots & \hat{h}_{P-s+1}
		\end{bmatrix}
		\begin{bmatrix}
			\beta_{0}\\
			\beta_{1}\\
			\vdots\\
			\beta_{s-1}
		\end{bmatrix}
		=
		\begin{bmatrix}
			\hat{h}_{P-1}\\
			\hat{h}_{P-2}\\
			\vdots\\
			\hat{h}_{P-s}
		\end{bmatrix}
	\end{matrix}
\end{align*}
which always has a solution.
Assume we find one solution $\bar{\beta} = [\bar{\beta}_0, \bar{\beta}_1,\ldots, \bar{\beta}_{P-1}]^T$.
By the second part of Lemma \ref{Lemma:DFT}, we have 
\begin{align*}
	\hat{h}_{\ell} = \sum_{u=0}^{s-1} \bar{\beta}_{u} \hat{h}_{\ell+u-s}, \forall \ell. 
\end{align*}
Now we prove by induction that $h_\ell = \hat{h}_\ell, \ell=0,1,\ldots, P-1$.

When $\ell=0$, we have 
\begin{align*}
	\hat{h}_{0} = \sum_{u=0}^{s-1} \bar{\beta}_{u} \hat{h}_{u-s} = \sum_{u=0}^{s-1} \bar{\beta}_{u} {h}_{u-s} = h_0
\end{align*}
where the second equation is due to the fact that  	$[h_{P-2s}, h_{P-2s-1},\ldots, h_{P-1}] = [\hat{h}_{P-2s}, \hat{h}_{P-2s-1},\ldots, \hat{h}_{P-1}]$ and $\hat{h}_{P+\ell} = \hat{h}_{\ell}, {h}_{P+\ell} = {h}_{\ell}$ (by definition).

Assume that for $\ell \leq\mu$, $\hat{h}_\ell = h_\ell$. 

When $\ell=\mu+1$, we have
\begin{align*}
	\hat{h}_{\mu+1} = \sum_{u=0}^{s-1} \bar{\beta}_{u} \hat{h}_{\mu+1+u-s} = \sum_{u=0}^{s-1} \bar{\beta}_{u} {h}_{\mu+1+u-s} = h_{\mu+1}
\end{align*}
where the second equation is because of the induction assumption for $\ell \leq\mu$, $\hat{h}_\ell = h_\ell$.

Thus, we have $h_\ell=\hat{h}_\ell$ for all $\ell$, which means $\mathbf{h} = \hat{\mathbf{h}} = \mathbf{C}^{\dag} \gamma$.
Thus $\mathbf{t}$,  the IDFT of $\mathbf{h}$, becomes $\mathbf{t} = \mathbf{C} \mathbf{h} = \mathbf{C} \mathbf{C}^{\dag} \gamma = \gamma$. 
Then the returned Index Set $V=\{j:e_{j+1}\not=0\} = \{j:\gamma_{j}\not=0\}$.
By Lemma \ref{Lemma:CycProb}, with probability 1, 
$\{j:\gamma_{j}\not=0\} = \{j:\|\mathbf{n}_{j}\|_0\not=0\}$.
Therefore, we have with probability 1, 
$V = \{j:\|\mathbf{n}_{j}\|_0\not=0\}$, which finishes the proof. \qed

\subsection{Proof of Theorem \ref{Thm:CycCodeOptimality} }
We first prove the correctness of the cyclic code.
By Lemma \ref{Thm:CycCodeDetection}, the set $U$ contains the index of all non-adversarial compute nodes with probability 1.
By Lemma \ref{Thm:CycCodeColumnSpan}, there exists $\mathbf{b}$ such that $\mathbf{W}_{\cdot,U} b = \mathbf{1}$. 
Therefore, $\mathbf{u}^{Cyc} = \mathbf{R}_{\cdot, U}^{Cyc} \mathbf{b} =  (\mathbf{G} \mathbf{W}+\mathbf{N})_{\cdot,U} \mathbf{b} = \mathbf{G}\mathbf{W}_{\cdot,U} \mathbf{b} = \mathbf{G} \mathbf{1}_{P}$. 
Thus,  The cyclic code  $(\mathbf{A}^{\mathit{Cyc}}, E^{\mathit{Cyc}}, D^{Cyc})$ can recover the desired gradient and hence resist any $\leq s$ adversaries with probability 1.

Next we show the efficiency of the cyclic code.
By the construction of $\mathbf{A}^{Cyc}$ and $\mathbf{W}$, the redundancy ratio is $2s+1$ which reaches the lower bound.
Each compute node needs to compute a linear combination of the gradients of the data it holds, which needs $\mathcal{O}((2s+1)d)$ computations.
For the PS, the detection function $\phi(\cdot)$ takes $\mathcal{O}(d)$ (generating the random vector $\mathbf{f}$) + $\mathcal{O}(dP + 2 P s )$ (computing $\mathbf{f} \mathbf{R} \mathbf{C}_{R}^{\dag}$) + $\mathcal{O}(s^2)$ (solving the Toeplitz system of linear equations in \eqref{eq:RecoverLinearSystem} ) + $\mathcal{O}((P-2s)s)$ (computing $h_{\ell}, \ell=0,1,2,\ldots, P-2s-1$ ) +  $\mathcal{O}(P\log P)$ (computing the DFT of $\mathbf{h}$ ) + $\mathcal{O}(P)$ (examining the nonzero elements of $\mathbf{t}$ ) = $\mathcal{O}( d + dP + 2 P s +s^2 +(P-2s)s+ P\log P+P) = \mathcal{O}( dP + P s + P\log P) $.
Finding the vector $\mathbf{b}$ takes $\mathcal{O}(P^3)$ (by simply constructing $\mathbf{b}$ via $\left[\mathbf{C}_{L}\right]_{\cdot,\bar{U}}$, though better algorithms may exist).
The recovering equation $\mathbf{R}_{\cdot,U} \mathbf{b}$ takes $O(dP)$.
Thus, in total, the decoder at the PS takes $\mathcal{O}(dP + P^3 + P\log P)$.
When $d\gg P$, \ie $d = \Omega(P^2)$, this becomes $\mathcal{O}(dP)$.
Therefore, $(\mathbf{A}^{\mathit{Cyc}}, E^{\mathit{Cyc}}, D^{Cyc})$ also achieves linear-time encoding and decoding. \qed

\section{Streaming Majority Vote Algorithm}
In this section we present the Boyer–-Moore majority vote algorithm \cite{MajorityVote}, which is an algorithm that only needs computation linear in the size of the sequence.

\begin{algorithm}[H]
	\SetKwInOut{Input}{Input}
	\SetKwInOut{Output}{Output}
	\Input{$n$ items $I_1, I_2,\ldots, I_n$}
	\Output{The majority of the $n$ items}
	Initialize an element $\textit{Ma} = I_1$ and a counter $\textit{Counter} = 0$.\\
    \For{$i=1$ \KwTo $n$ }
    {
        \uIf{$\textit{Counter}==0$}
        {
             $Ma = I_i$.\\
             $\textit{Counter} = 1$.
        }
        \uElseIf{$Ma == I_i$}
        {
            $\textit{Counter} = \textit{Counter} + 1$.
        }
        \Else
        {
            $\textit{Counter} = \textit{Counter} - 1$.
        }
  }
  Return $Ma$.
	\caption{Streaming Majority Vote.}
	\label{Alg:MajorityVote.}
\end{algorithm}
Clearly this algorithm runs in linear time and it is known that if there is a majority item then the algorithm finally will return it \cite{MajorityVote}.}

\section{Equivalence of the Byzantine recovery to sparse recovery in Section \ref{ocs}}
\label{sp}
Here we explain how the problem of deigning a regular mechanism is equivalent to a sparse recover problem. We consider the setting, where we want to recover a $d$-dimensional vector $g$ from $m:=rd_c$-dimensional observations $R = Zg+n$, where $n$ is $k:=sd_c$-sparse. First we argue that there are many design matrices $Z$ that allow this. As we saw, the requirement on $Z$ is that when $Zv\in B_{2s,r}$, then $v=0$. Suppose $w=Zv\in B_{2s,r}$ and let $S$ be the set of size at most $k$ of the nonzero coordinates of $w$. Letting $Z' = Z_{S^c,\cdot}$ be the $(m-|S|)\times d$ matrix formed by the rows of $Z$ outside of $S$, we have that $Z'v=0$. If $m-|S|\ge d$ and $Z'$ has full rank $d$, then this implies $v=0$ and we are done. As $|S|\le 2k$, it is enough to ensure that each $(m-2k)\times d$ submatrix formed by the rows of $Z$ has rank $r$. When $m-2k\ge r$, this holds with probability one  when---for instance---the entries of $Z$ are sampled iid from a distribution with a density that is absolutely continuous with respect to the Lebesgue measure. The reason is that the set of rank-deficient matrices is a set of zero Lebesgue measure in the space of all matrices \cite{lax2007linear}.

Thus, we have such a matrix $Z$ and the observation $R = Zg+n$, where $n$ is $k$-sparse, and the goal is to find the unique vector $g$ such that $R = Zg+n$. It is possible to solve this problem by enumerating all subsets $S$ of size at most $k$ to find the unique subset for which  $R_{S^c}  \in span(Z_{S^c,\cdot})$, or equivalently  $ P_{S} R_{S^c} = 0$, where $ P_{S}$ is the projection into the orthocomplement of the column span of the matrix $Z_{S^c,\cdot}$. This is a combinatorial algorithm, which requires a search over $\binom{m}{k}$ subsets, and thus illustrates the challenges of this problem.

In fact, multiplying $R = Zg+n$ with an orthogonal complement $Z^\perp$ such that $Z^\perp Z = 0_{m-d,d}$, we see that it is sufficient to be able to find the unique $k$-sparse $n$ such that $y = Z^\perp n$, where $y := Z^\perp R$. Moreover, this is also necessary, because uniqueness means that $y = Z^\perp n = Z^\perp n'$, or also $Z^\perp(n'-n'')=0$ implies $n-n'=0$, or equivalently that $Ker(Z^\perp) \cap B_{2s,r} = \{0\}$. Since $Ker(Z^\perp)=Im(Z)$, this is equivalent to the previous claim. In conclusion, our question is exactly equivalent to a sparse recovery problem, finishing the argument. 
\section{Experimental Details}
\label{Sec:SOLON:Appendix:setup}
Additional experimental details are discussed here.

\paragraph{Experimental setups.}
All experiments were conducted on a cluster of 50 machines. 
Each machine is equipped with 20 Intel Xeon E5-2660 2.6 GHz cores,
160 GB RAM, and 200GB disk with Ubuntu 18.04 LTS as the OS.
All code was implemented in Python 3.8.
The entire experiments took several months, including debugging and evaluation time. 
Note that this is mostly because the defense approach BULYAN is slow.
Evaluating \solon alone can be much faster.
In addition, \solon was built in Python for demonstration purposes.
Code optimization with addition tools (such as Cython) can give extra speedups.

\paragraph{Hyper-parameters.} We evaluate \solon{}, \bulyan{}, \draco{}, and \signum{} using the datasets and models in Table \ref{Tab:solon:DataStat}. We add  vanilla SGD without any Byzantine adversary as the gold standard for accuracy. 
Our cluster consists of one parameter server (PS) and 100 compute nodes, hosted on 50 real c220g5 machines. \solon{} and \draco{} partition the compute nodes evenly into 5 groups.  
The training batch size  for \solon{} and \draco{} is 120 per group, and equivalently six per compute node for other baselines. 
We fix the compression ratio $r_c$ to 10, except in the experiment where we evaluate the effect of compression ratio. For \draco{}, we only test its repetition code scheme, since the cyclic code scheme has a slightly slower performance \cite{AggregathorDamaskinosEGGR19}. We calculate the redundancy ratio by $r=2s+1$ for \draco{}. To compare with the best possible performance of \draco{}, we set its $r=11$ such that it only uses 55 compute nodes to reduce communication overhead in the end-to-end performance test.  These methods are trained for 3,000  iterations and evaluated on the test set every 25 steps. The learning rate is set to 0.1. 

Under the constant attack, however, we found that \signum{} may diverge. We therefore  decay the learning rate as $5 \cdot 10 ^{-5} \cdot 0.95^{\floor{t/10}}$ for \signum{} to get a more stable accuracy curve in Figure \ref{fig:against-defense}(b), where $t$ is the  number of iterations. 
We also observe that \bulyan{} does not converge under the ALIE attack, matching the observation in~\cite{detox2019}. We therefore lower the learning rate to $0.002 \cdot 0.95^{\floor{t/10}}$ in order to enable \bulyan{} to converge to a lower accuracy in Figure \ref{fig:against-defense}(c). 

For language model task using LSTM over Wikitext-2 dataset, we observe that \bulyan{} and \signum{} do not converge very well under small batches. We then decide not to keep the linearity of batch size (batch size linearity is mentioned in \cite{vogels2019powersgd}) among \solon{} and these two baselines.  As a result, the batch size setting is $b=3$ for Vanilla SGD with LSTM and $b=60$ for others. The learning rate is set as $lr=40$ during \solon{}, \draco{}, Vanilla SGD, and $lr=20$ for \bulyan{} and \signum{}. We use learning rate warm-up for the first 800 iterations of \solon{}, \draco{} and Vanilla SGD, and during the first 400 iterations for the rest of two. Since we observe in previous experiments that \bulyan{} and \signum{} fail in ALIE and constant attacks respectively, we only show their performance under reverse-gradient attack. 

\begin{figure*}[t] 
\centering
\includegraphics[width=0.92\textwidth]{Figure/accu_iter_hour_legend_node50.pdf}\\
\subfigure[Accuracy vs steps, rev-grad]{\includegraphics[width=0.33\textwidth]{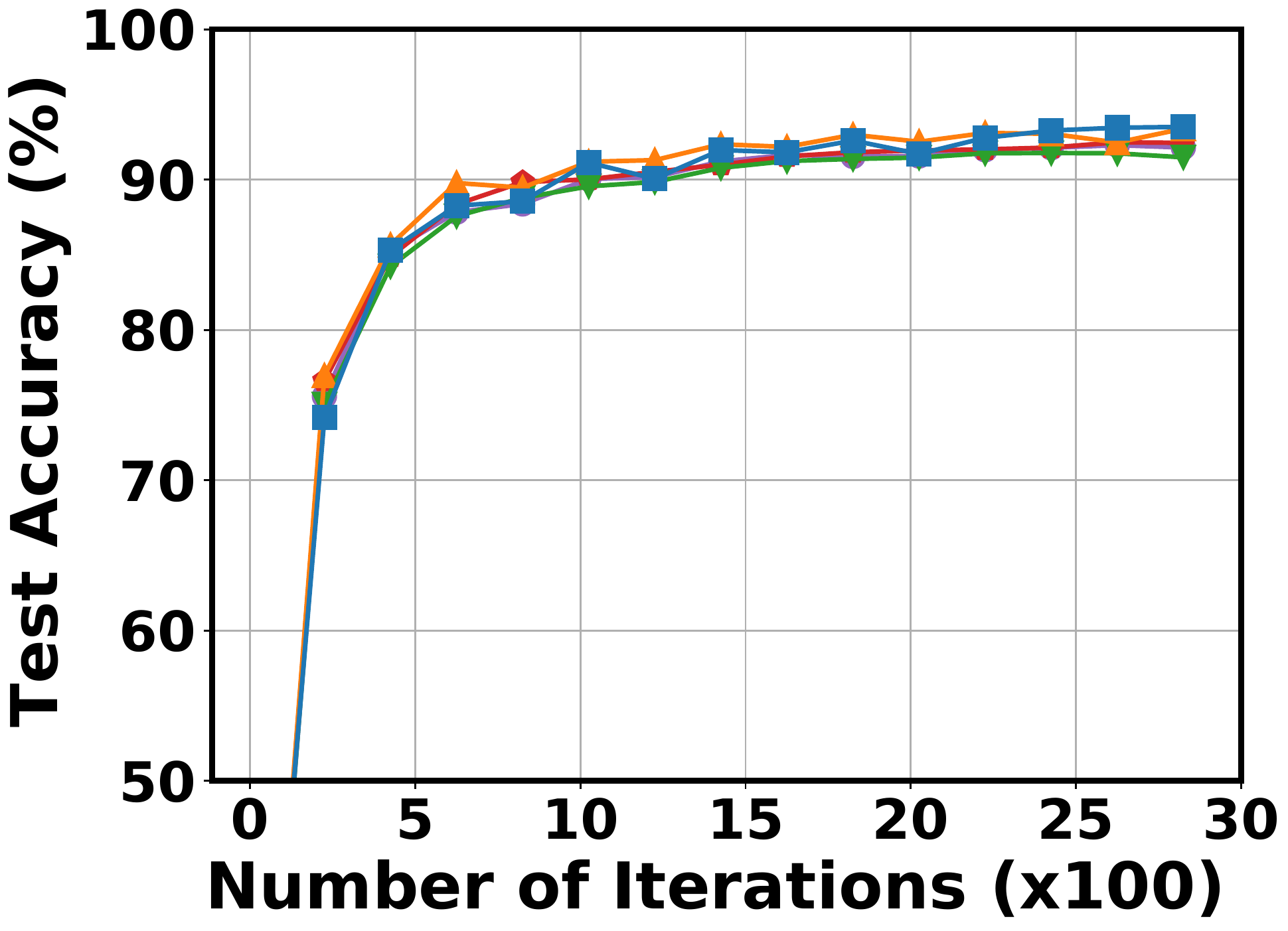}}
\subfigure[Accuracy vs steps,  constant]{\includegraphics[width=0.31\textwidth]{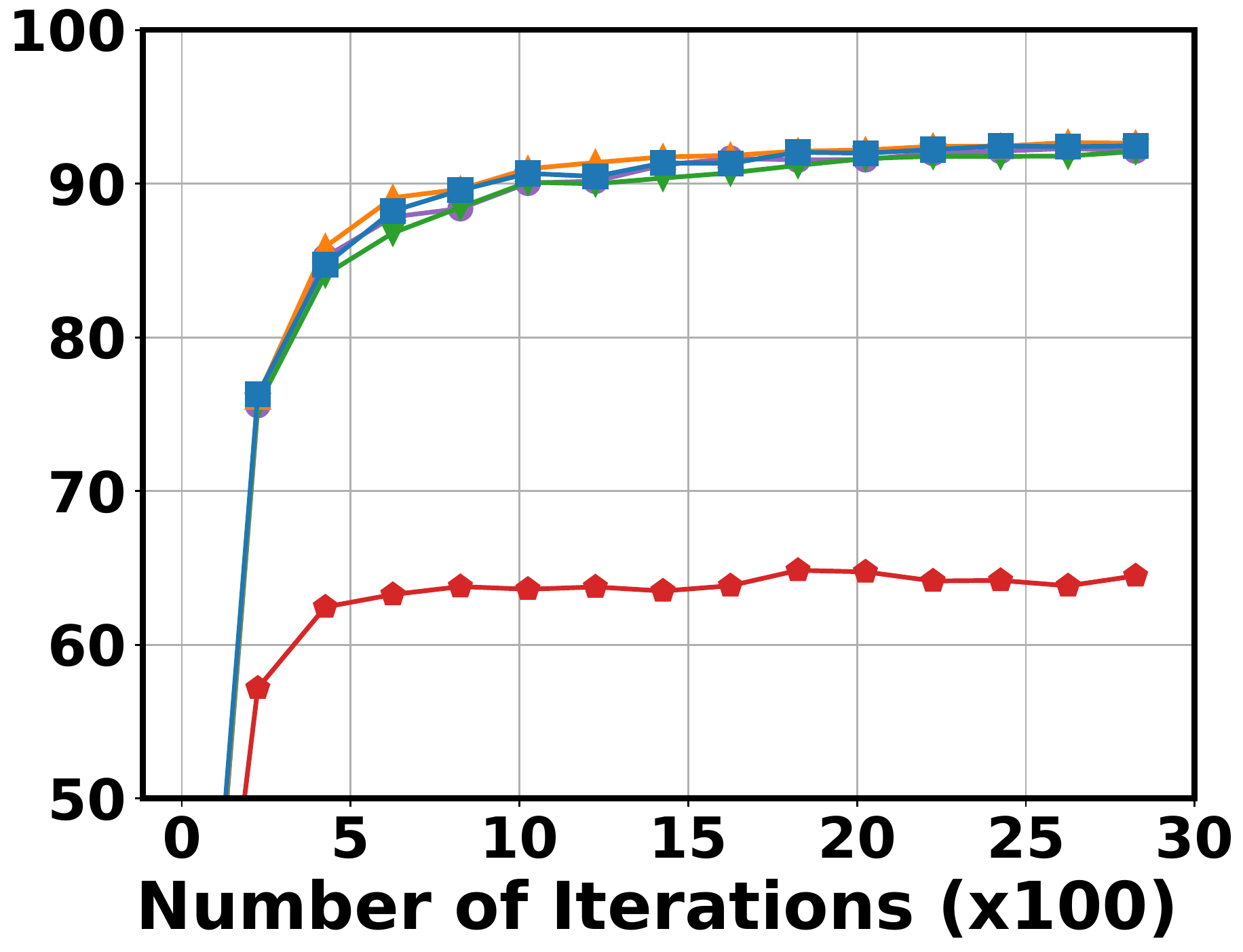}}
\subfigure[Accuracy vs steps, ALIE]{\includegraphics[width=0.31\textwidth]{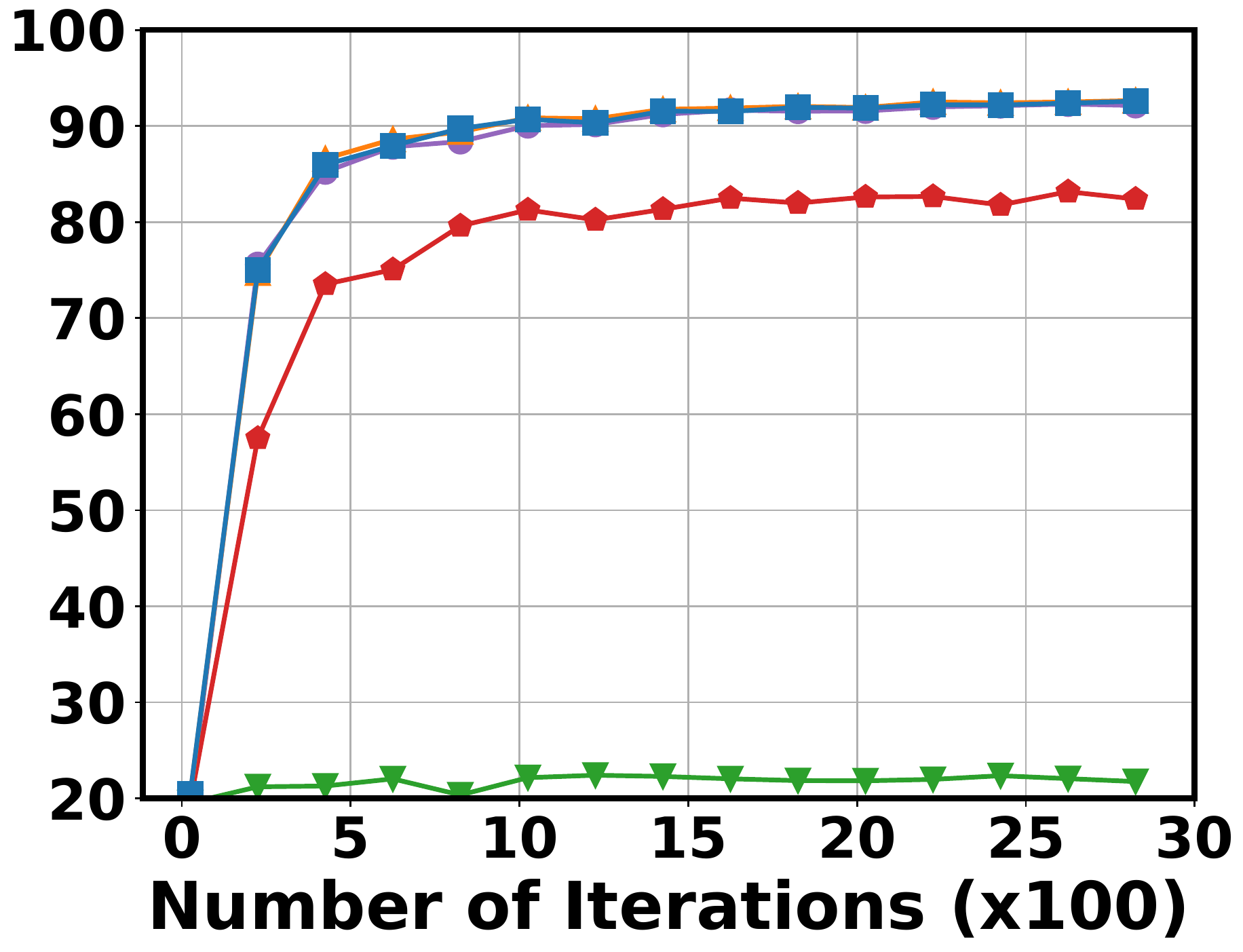}}
\subfigure[Accuracy vs time, rev-grad]{\includegraphics[width=0.33\textwidth]{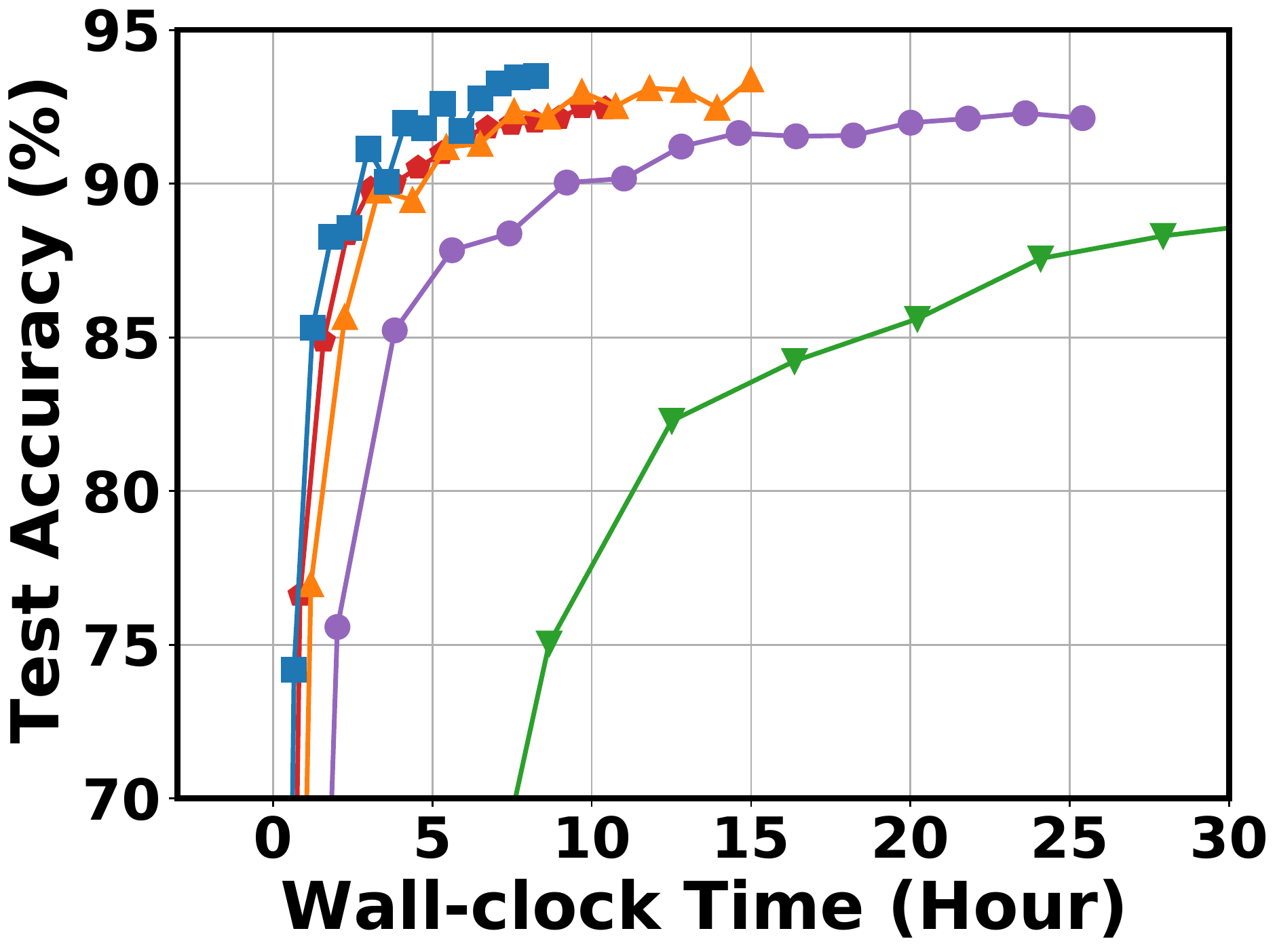}}
\subfigure[Accuracy vs time, constant]{\includegraphics[width=0.31\textwidth]{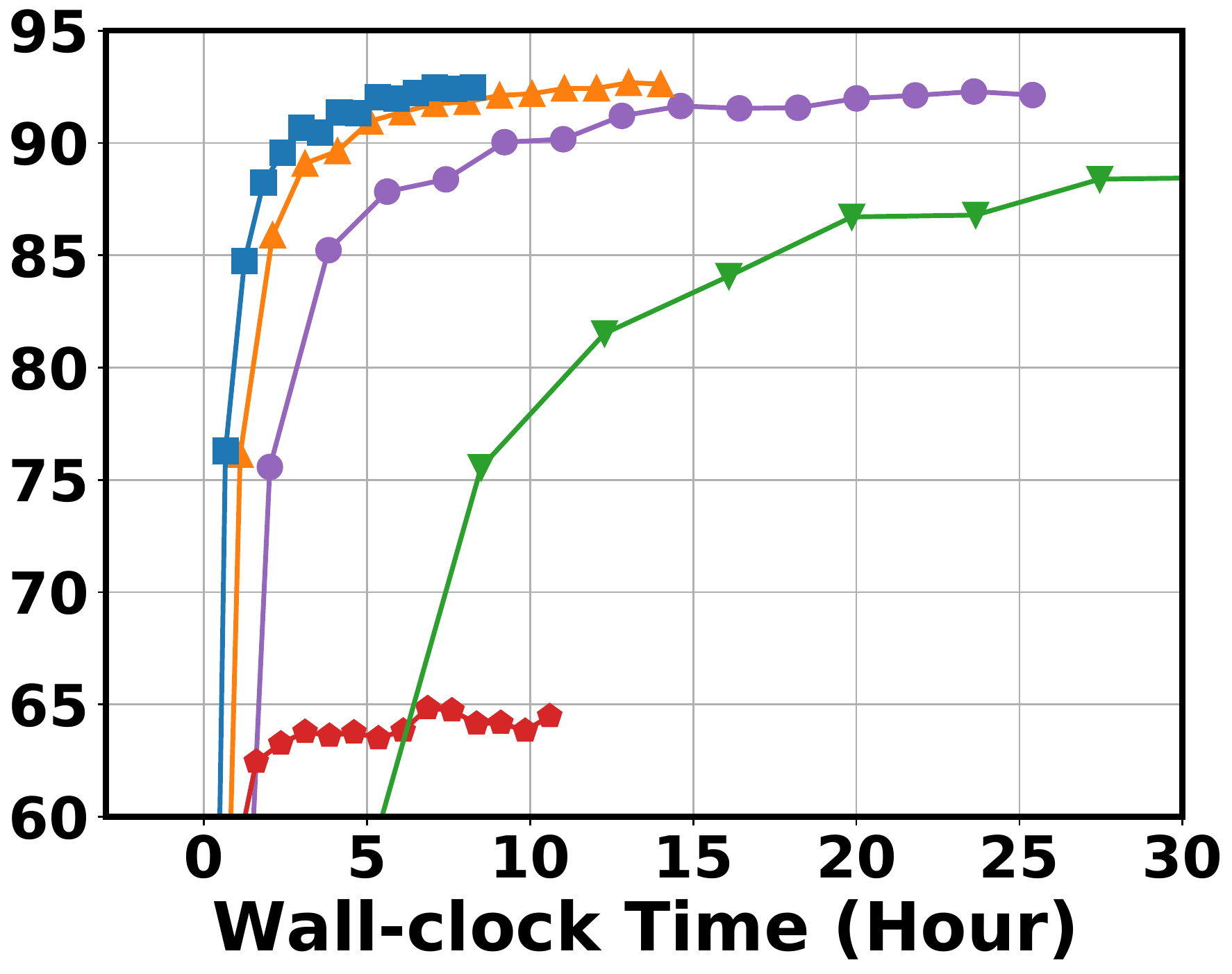}}
\subfigure[Accuracy vs time, ALIE]{\includegraphics[width=0.315\textwidth]{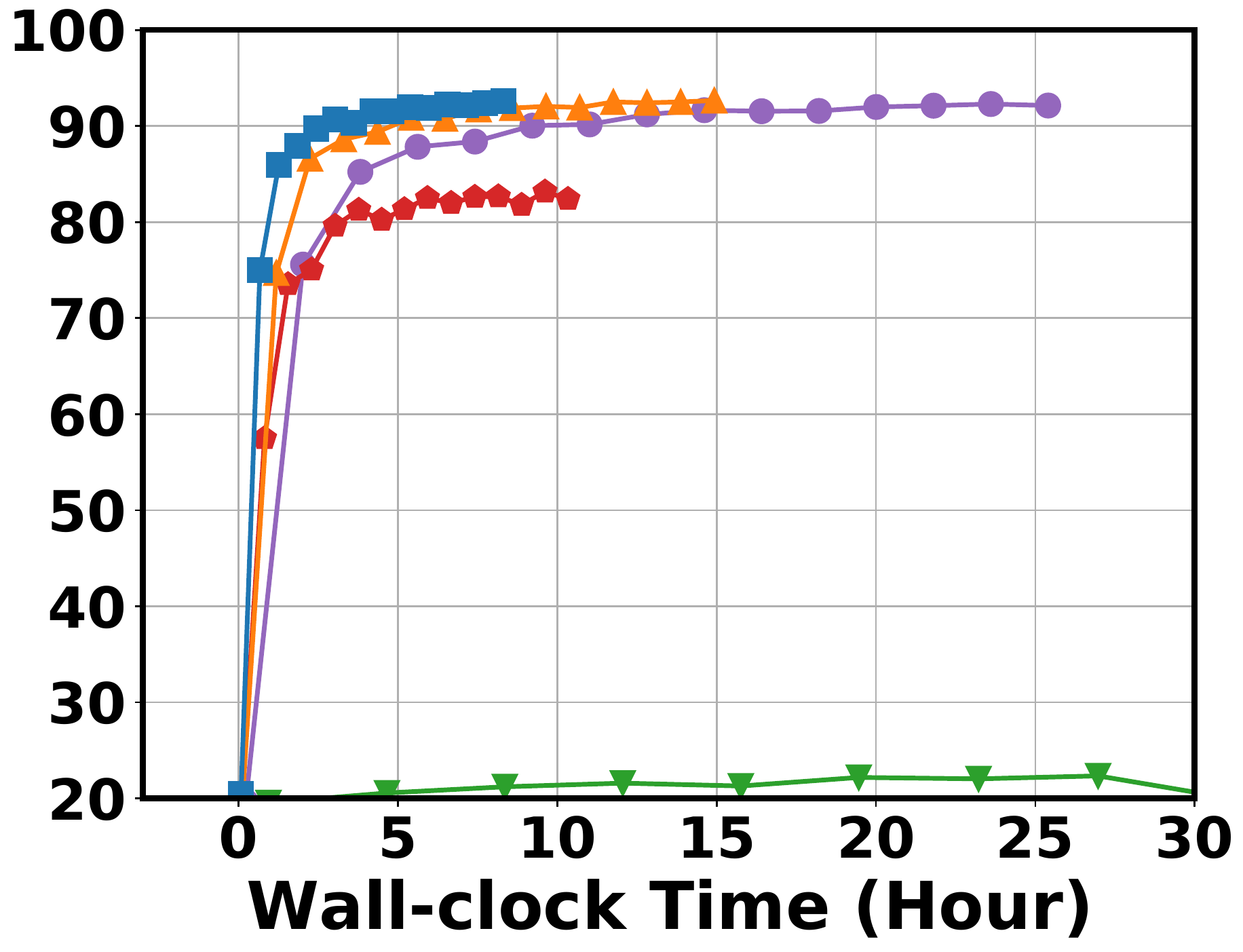}}
\vspace{-2mm}
\caption{End to end convergence performance of \solon{} and other baselines on VGG13-BN and SVHN. }
\label{fig:against-defense:SVHN}
\end{figure*}

\section{Additional Empirical Results}
\label{Sec:SOLON:Appendix:experiments}
Now we give additional empirical findings. 
The goal is to (i) verify if \solon's performance gain is valid in wider applications, (ii)  study how the number of Byzantine attacks  $s$ as well as the compression ratio $r_c$ may affect the performance of \solon, and (iii) understand the computational cost of \solon{}'s  decoder. 

\paragraph{End to end convergence performance, ctd. } 
To verify if  \solon{} is substantially more robust and efficient in wider applications,   we conducted evaluations for training an additional model, VGG13 on SVHN. Figure \ref{fig:against-defense:SVHN} shows the end to end performance of \solon and other baselines under a few different attacks. 
Overall, we observe similar trends seen for other models in the main paper. 
Generally, \solon{} converges to the same accuracy level of vanilla SGD under no Byzantine attack, while providing much faster runtime performance than all baselines. \bulyan{} and \signum{} fail by losing around 60\% and 30\% accuracy under ALIE and constant attack in the experiment of VGG13+SVHN. However, we notice two slight differences. First, the advantage in runtime performance of \solon{} in Figure \ref{fig:against-defense:SVHN} seems to be not as remarkable as it is in RN18 (Figure \ref{fig:against-defense}). This is because SVHN dataset is an easier task and the model accuracy rises faster than ResNet18+CIFAR10. It then makes the baseline performances look better. However, if there are harder tasks where the accuracy climbs slower, the advantage of \solon{} will be clearer. Second, \bulyan{} and \signum{} perform worse than \solon{} both in final accuracy and runtime in the language model task of LSTM+Wikitext-2. We think it's because this task is more sensitive to gradient changes, such that the exact gradient recovery schemes like \solon{} have better performance than approximated ones. 

\begin{figure*}[t] 
\centering
\includegraphics[width=0.9\textwidth]{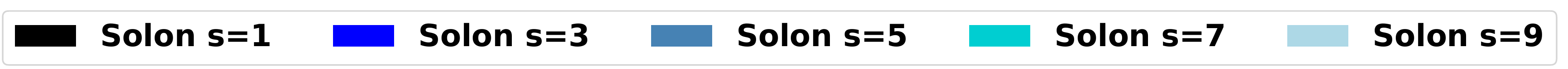}\\
\subfigure[ResNet18, CIFAR10]{\includegraphics[width=0.32\textwidth]{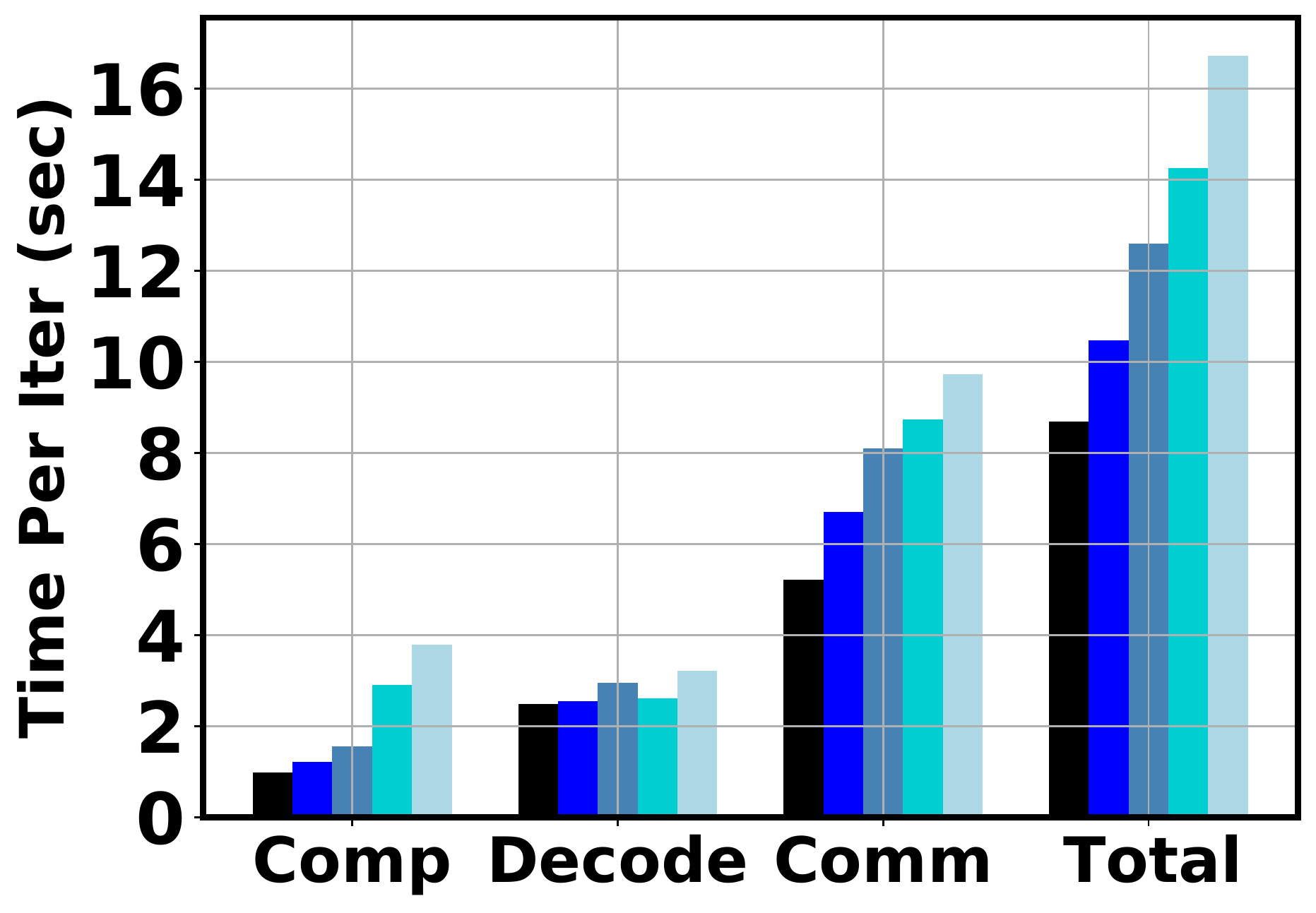}}
\subfigure[VGG13-BN, SVHN ]{\includegraphics[width=0.32\textwidth]{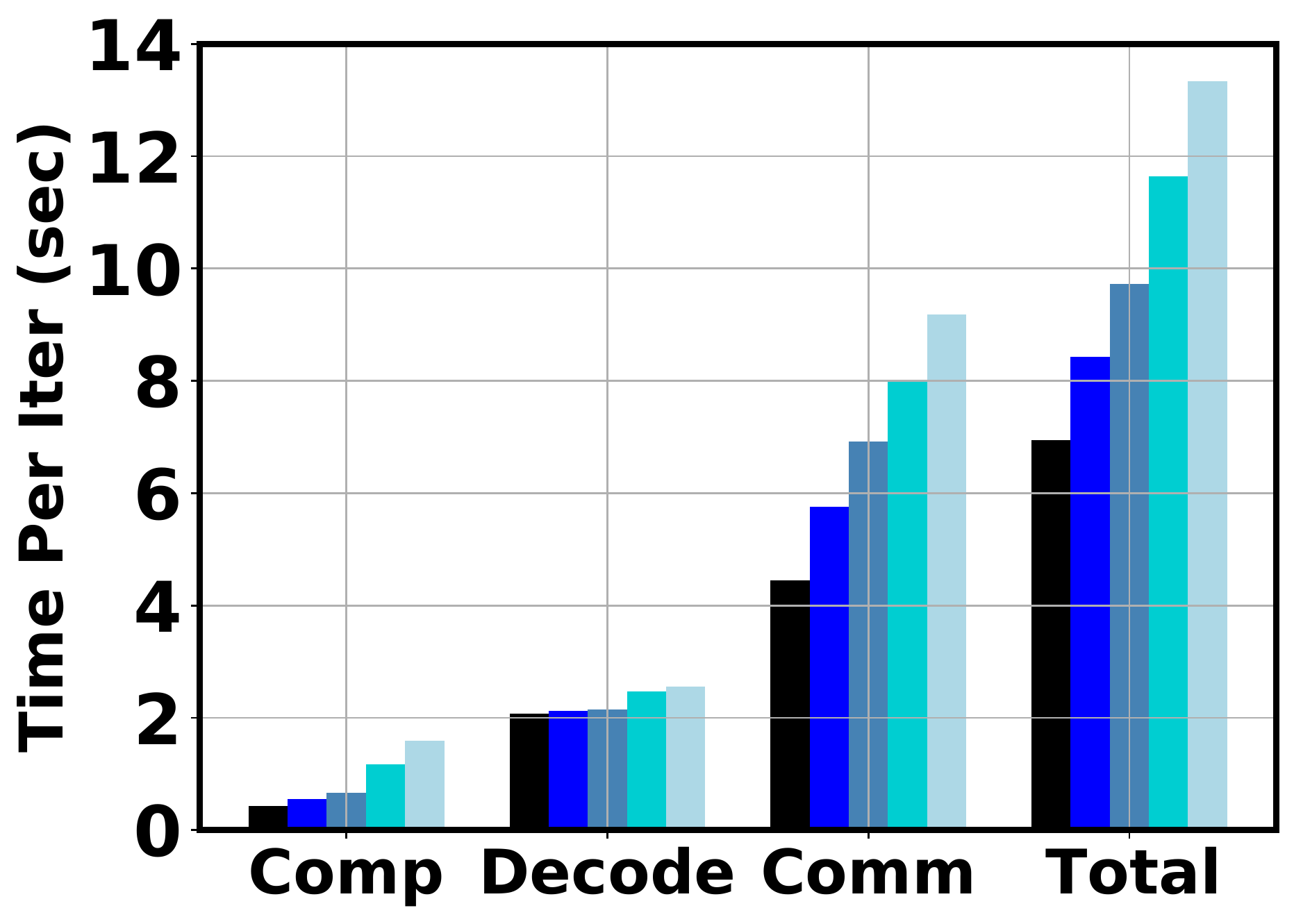}}
\subfigure[LSTM, Wikitext-2]{\includegraphics[width=0.32\textwidth]{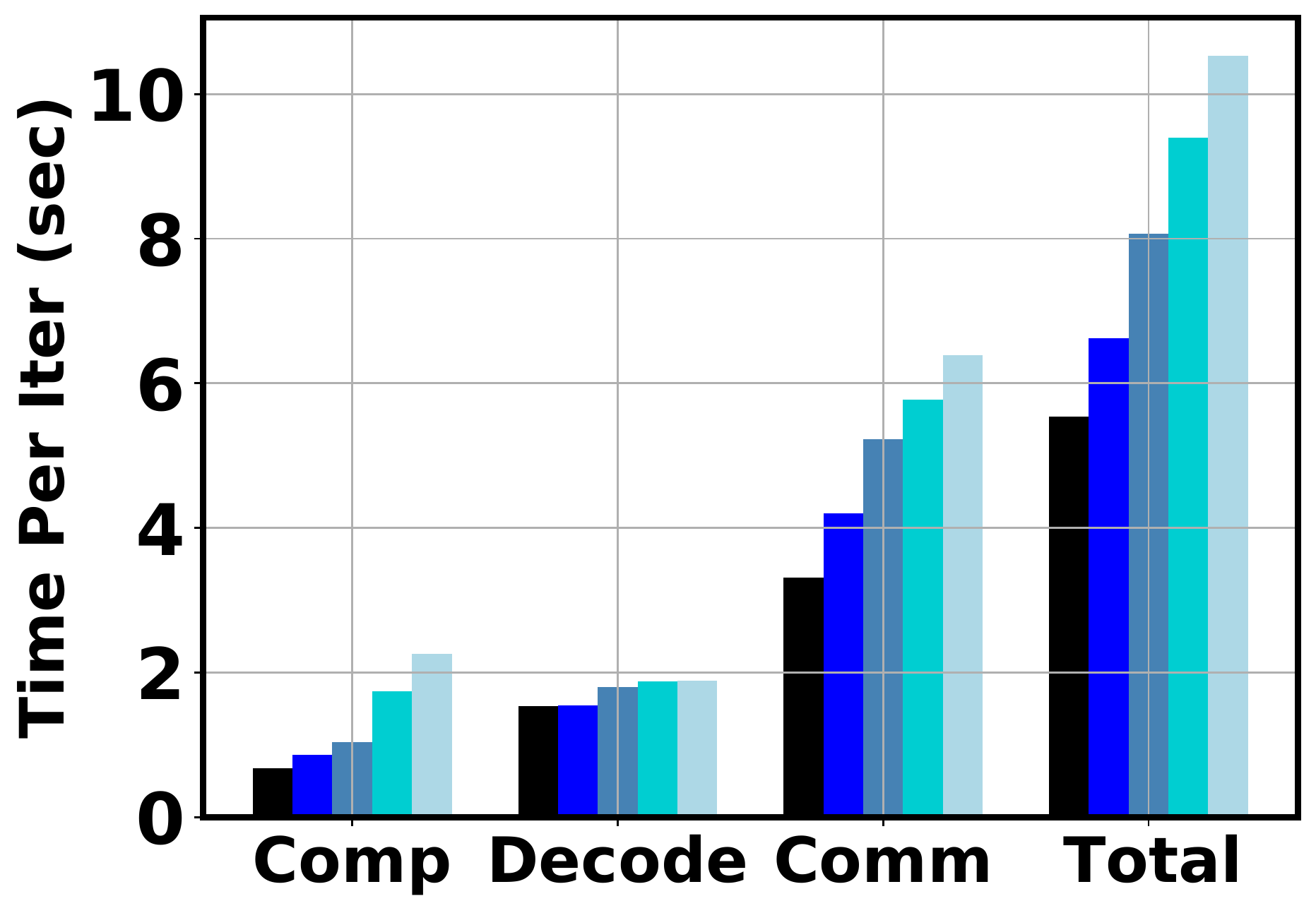}}
\caption{Time breakdown per iteration of \solon{}, varying $s$ with fixed $r_c=10$ over three models+datasets. The redundancy ratio is set as $r=2s+r_c$, and thus increases as $s$ increases. It is easy to see that all costs increase linearly as the number of adversaries $s$ increases. } 
\label{fig:timebreakdown:solonvarys}
\end{figure*}

\paragraph{The influence of the number of adversaries.} 
We now evaluate the effects of the number  $s$ of adversaries on the runtime of \solon{}, as shown in Figure \ref{fig:timebreakdown:solonvarys}. We observe that the computation time,  communication time and decoding time of \solon{} increase roughly linearly as the number of adversaries $s$ increases. This is reasonable, because the minimum  redundancy ratio $r$ increases with the number of adversaries $s$, when $r_c$ is fixed. This leads to an increase in the total number of workers and batch sizes, \ie the communication and computation overhead. This result shows that the cost of \solon{} increases linearly as $s$ increases, and verifies that \solon{} is scalable to distributed training. 

\begin{figure*}[b] 
\centering
\subfigure[ResNet18, Cifar10]{\includegraphics[width=0.34\textwidth]{Figure/time_breakdown_solonrc_node50_RN18_fixn.pdf}}
\subfigure[VGG13-BN, SVHN ]{\includegraphics[width=0.32\textwidth]{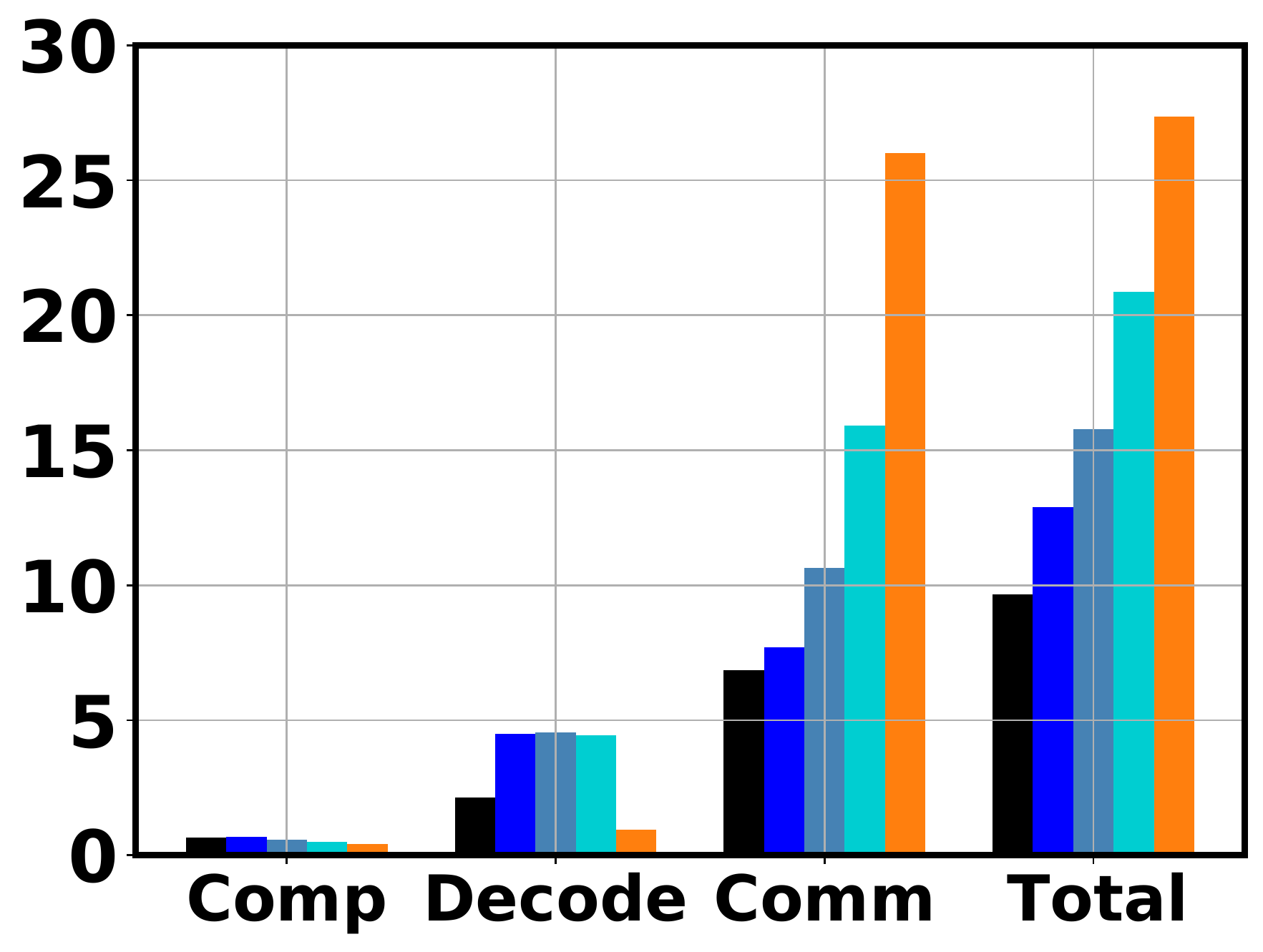}}
\subfigure[LSTM, Wikitext-2]{\includegraphics[width=0.32\textwidth]{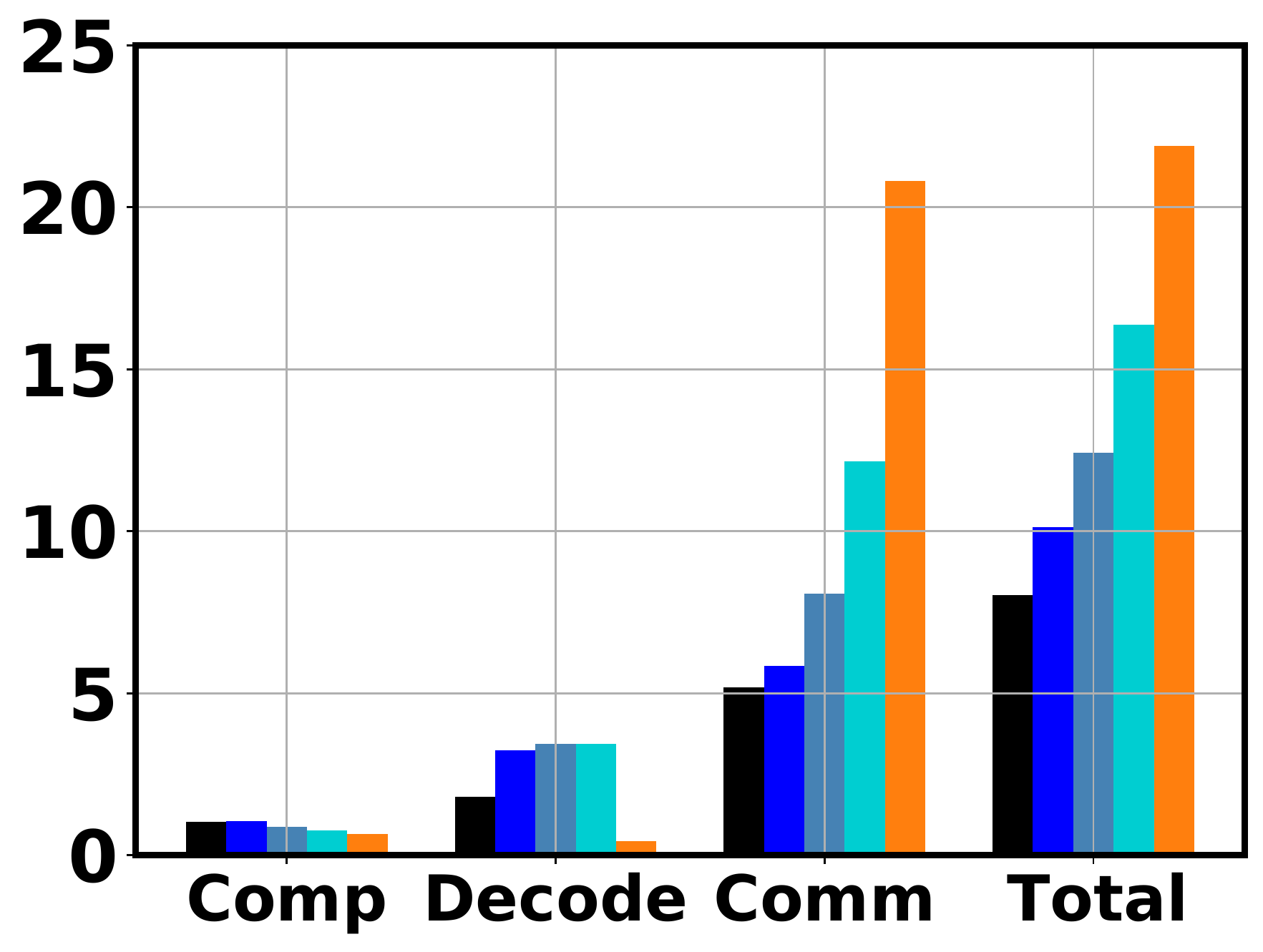}}
\caption{Time breakdown per iteration of \solon{}  over three models+datasets, varying $r_c$ with fixed $s=5$. The redundancy ratio is calculated as $r=2s+r_c$, and thus increases when $r_c$ increases from $2$ to $10$. } 
\label{fig:timebreakdown:solonvaryrc}
\end{figure*}

\paragraph{The effects of the compression ratio, ctd. } 
In the main text, we showed the effects of the compression ratio $r_c$ when $s$ and number of nodes $P$ are fixed for ResNet-18. The remaining results for VGG13 and LSTM are shown in Figure \ref{fig:timebreakdown:solonvaryrc}. Overall, the total time decreases as $r_c$ increases, showing that \solon{} is efficient in reducing communication time and can be applied to large clusters. But there are some subtleties. First, the communication time decreases as $r_c$ increases, because the size of compressed gradients decreases. However, when $r_c$ becomes larger, the additional benefit of communication cost becomes smaller. This is because the increase of the redundancy ratio $r$ adds some overhead. Also, the overhead in communication also limits the best possible communication results achieved.  
Next, the computation time increases linearly when $r_c$ increases from two to $10$. This is also because the number of groups gradually decreases as $r$ increases, which introduces larger batch sizes for each group in order to maintain total equivalent batch sizes. This overhead is common across all three tasks. 

\begin{figure*}[t] 
\centering
\includegraphics[width=\textwidth]{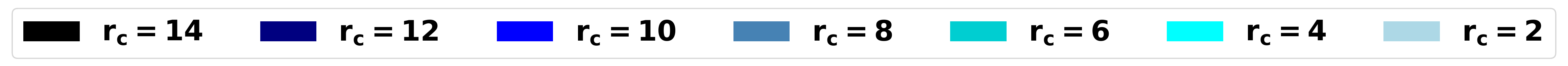}\\
\subfigure[Using scipy.linalg.lstsq]{\includegraphics[width=0.48\textwidth]{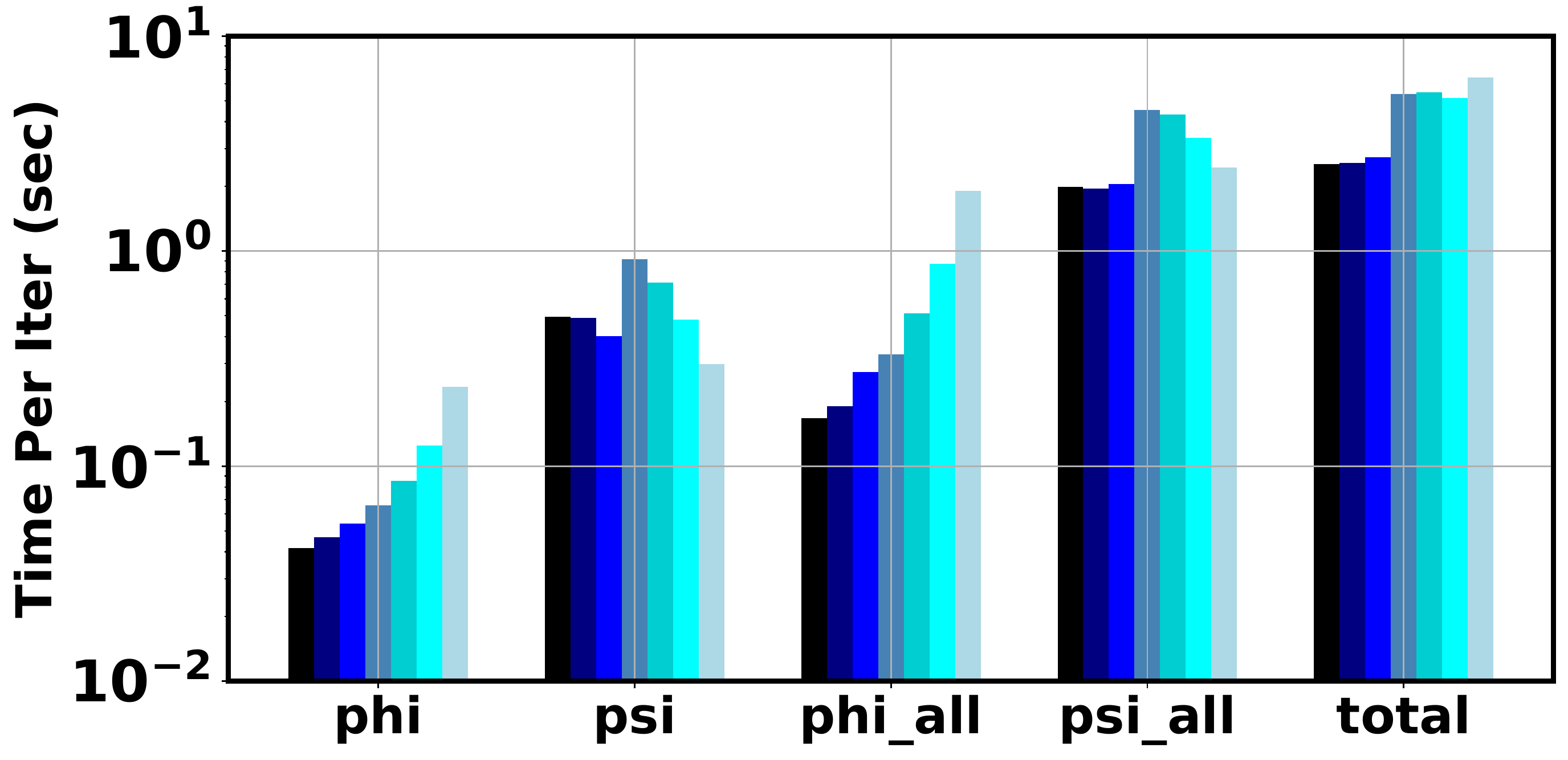}}
\hskip 5pt
\subfigure[Using scipy.linalg.pinv ]{\includegraphics[width=0.48\textwidth]{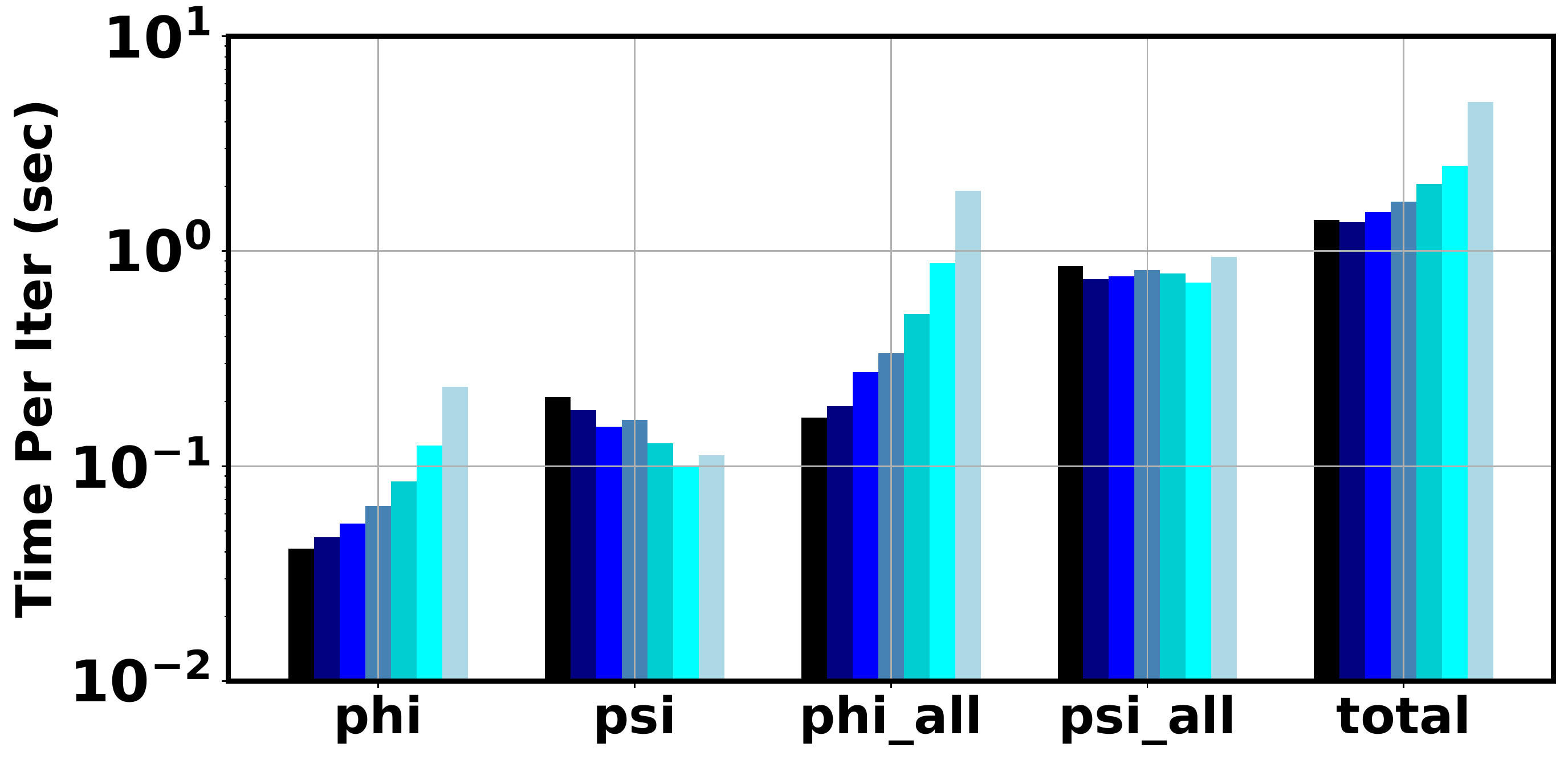}}
\caption{Decoder time breakdown per iteration of \solon{} over ResNet-18+CIFAR-10, varying $r_c$ with fixed $s=5$, in log scale. (a) Using linear equation solver to build  $\psi$. (b) Using pseudo-inverse to build $\psi$. ``phi'' and ``psi'' refer to the single execution time for function $\phi$ and $\psi$. ``phi\_all'' and ``psi\_all'' refer to the execution time of the designated functions for all groups in the whole cluster. ``total'' means the complete decoding time per iteration. } 
\label{fig:timebreakdown:decoder}
\end{figure*}

In addition, we push the analysis on the decoding time one step further, where the breakdown results are shown in Figure \ref{fig:timebreakdown:decoder}(a). To test the decoder time breakdown, we fix the maximum number of threads used by pytorch by setting $\text{MKL\_NUM\_THREADS}=10$ instead of 20 in the previous end to end tests to avoid all possible computing resource conflict on a single machine.  On the first glance, the total decoding time does not seem to have a clear pattern when $r_c$ increases. However, we notice that the time for $\phi (\cdot)$ increases almost linearly as $d_c$ increases (by decreasing $r_c$). This is reasonable because the major overhead for $\phi$ is computing $\mathbf{r}_{j,c} = \mathbf{f} \mathbf{R}_j$, of which the computation complexity is $O(P/r\cdot r\cdot d_c)=O(Pd_c)=O(Pd/r_c)$ as discussed in Section \ref{sec:solon}, Algorithm 1 and Appendix~Section~C. 
On the other hand, the time for $\psi$ seems to be non-monotonic, but the major overhead is to solve the linear system $    \begin{bmatrix}\hat{\mathbf{W}}_{j,r_c+s-1},  & -\hat{\mathbf{W}}_{j,s-1} \odot \left( \mathbf{r}_{j,c}^T \mathbf{1}_{s}^T\right)\end{bmatrix}\mathbf{a}= \mathbf{r}_{j,c} \odot\left[\hat{\mathbf{W}}_{j,s}\right]_{\cdot,s}$, whose computation complexity is roughly $\mathcal{O}(Pr^2+Pd r_c/r)$. We later realize that the non-monotonicity is caused by the linear equation solver ``scipy.linalg.lstsq''. When we altered the solver using ``scipy.linalg.pinv'' presented in Figure \ref{fig:timebreakdown:decoder}(b), we observe that the time for $\psi$ is decreasing monotonically as $r_c$ decreases, which matches the theoretical analysis. We then argue that the implementation of the linear solver would determine the actual runtime performance of the decoder. 
Thus, we end the discussion here at the breakdown level of python functions, since the efficient design, implementation and analysis of the linear solvers are beyond the scope of this work. 

\section{Potential Limitation of \solon{}. }
\label{Sec:SOLON:socialimpact}

\paragraph{Discussion.}
To conclude our evaluations, \solon{} greatly reduces communication cost and is scalable to large clusters, but its parameters should be chosen carefully for different cluster resources and tasks.  An interesting question is when to use \solon{} instead of other methods. The experiments and discussions already reveal a partial answer. If there are strong attacks in term of the attack ratio and attack type and limited network bandwidth, but higher accuracy is demanded, \solon{} is preferred.  On the contrary, if the attacks are expected to be weak, and if quick results are needed with less precision, algorithms which use more approximations--such as \signum{} and its further variations--can be chosen. In addition, \solon{} requires $r\geq 2s+r_c$ and $r$ dividing $P$, which may require some tuning on the number of machines used. Furthermore, $r_c$ cannot be too large. Extremely large $r_c$ wastes computing resources, and may cause numerical issues in decoding (\eg as we observed when $r_c>14$ and precision is limited). As future work, we will consider developing a set of rules to pick the best Byzantine-resilient algorithm given a scenario, and developing \solon{} variations for asynchronous and decentralized training. 

\eat{\paragraph{Societal Impact.}
Distributed training has been widely adopted for training ML models on large scale datasets.    
Multiple concerns exist in distributed learning systems. 
For example, security is a common concern, i.e., whether certain components of the systems are occupied by third parties and thus deliberately hurt the downstream learning tasks. 
Communication cost is another important issue.
While existing work usually treats each concern separately, the proposed \solon aims at resisting attacks and reducing communication cost simultaneously. 
This shows the possibility and benefits to optimize multiple objectives in distributed learning design.
To stimulate more research in this area, we plan to open source our code in the camera ready version.
}

\end{document}